\documentclass[11pt]{article}

\usepackage[preprint]{acl}
\usepackage{times}
\usepackage{latexsym}

\usepackage[T1]{fontenc}
\usepackage[utf8]{inputenc}

\usepackage{microtype}

\usepackage{inconsolata}

\usepackage{graphicx}

\usepackage{subcaption}
\usepackage{booktabs}
\usepackage{tabularx}
\usepackage{array}
\usepackage{ragged2e}
\newcolumntype{Y}{>{\RaggedRight\arraybackslash}p{0.23\textwidth}}
\usepackage[table]{xcolor}   % 표 색상
\definecolor{examplegray}{gray}{0.95} % 아주 옅은 회색
\definecolor{groupgray}{gray}{0.93} % 그룹 헤더용 옅은 회색
\usepackage{amsmath}

\usepackage{graphicx} % 이미지 처리 기본 패키지
\usepackage{float}    % 그림 위치 고정([H])을 위해 (선택사항)
\usepackage{booktabs}
\usepackage{tcolorbox}
\usepackage{tikz}
\usetikzlibrary{positioning, arrows.meta, shapes.geometric}
\usepackage{tabularx}
\usepackage{array}
\usepackage{booktabs}
\usepackage{listings}
\usepackage{multirow}
\tcbuselibrary{breakable,skins}
\usepackage{placeins,needspace}
\title{EconCausal: A Context-Aware Economic Reasoning Benchmark \\for Large Language Models}
\definecolor{grayrow}{gray}{0.9}
\newcommand{\mypara}[1]{\vspace{2.5pt}\noindent\textbf{#1}}
\newcommand{\econcausal}{EconCausal}

\tikzset{
    node_style/.style={
        circle, 
        draw=black, 
        thick, 
        minimum size=1.2cm, 
        font=\bfseries
    },
    arrow_style/.style={
        -{Latex[length=3mm]}, 
        thick, 
        draw=gray!80!black
    },
    causal_arrow/.style={
        -{Latex[length=3mm]}, 
        very thick, 
        draw=blue!80!black
    }
}
\author{
\textbf{Donggyu Lee\textsuperscript{1}},
\textbf{Hyeok Yun\textsuperscript{2}},
\textbf{Meeyoung Cha\textsuperscript{3}},
\textbf{Sungwon Park\textsuperscript{4,*}},
\textbf{Sangyoon Park\textsuperscript{5,*}},
\textbf{Jihee Kim\textsuperscript{2,*}}
\\
\textsuperscript{1}Graduate School of Data Science, KAIST \\
\textsuperscript{2}College of Business, KAIST \\
\textsuperscript{3}Data Science for Humanity Group, MPI-SP \\
\textsuperscript{4}School of Computing, KAIST \\
\textsuperscript{5}Division of Social Science, HKUST \\[-0.5ex]
\texttt{\{donggyu.lee, ed\_yun98, psw0416, jiheekim\}@kaist.ac.kr} \\
\texttt{mia.cha@mpi-sp.org},
\texttt{sangyoon@ust.hk}
\\[-0.2ex]
}

\begin{document}
\maketitle

\begin{abstract}
Socio-economic causal effects depend heavily on their institutional and environmental contexts. The same intervention can produce different, even opposite, effects across regulatory regimes, market conditions, time periods, or populations. This poses a challenge for large language models (LLMs) in decision-support roles: can they infer the direction of a causal effect under a specified context, and revise that judgment when the context changes?

To address this, we introduce \textsc{\econcausal{}}, a large-scale benchmark of 10,490 context-annotated causal triplets extracted from 2,595 high-quality empirical studies in top-tier economics and finance journals, constructed through a rigorous four-stage pipeline with multi-run consensus, context refinement, and multi-critic filtering.

Across 19 models, we observe source-sign
over-transfer in cross-context prediction.
While top models reach $\sim$84\% accuracy in single-context
prediction, mean accuracy is 74.75\% overall
but drops to 44.27\% when source and target signs disagree.
On these cases, accuracy is 11.54~pp lower with the signed
source example than with the target input alone.
Mean accuracy is below 60\% under corrupted source signs.
Models also struggle to predict study-reported
nonsignificant effects (19.16\% accuracy). The dataset and benchmark are publicly available at \url{https://anonymous.4open.science/r/econcausal-benchmark-6F12}.

\end{abstract}

\section{Introduction}

\begin{figure*}[!t]
\vspace{-5mm}
\centering
\includegraphics[width=\textwidth]{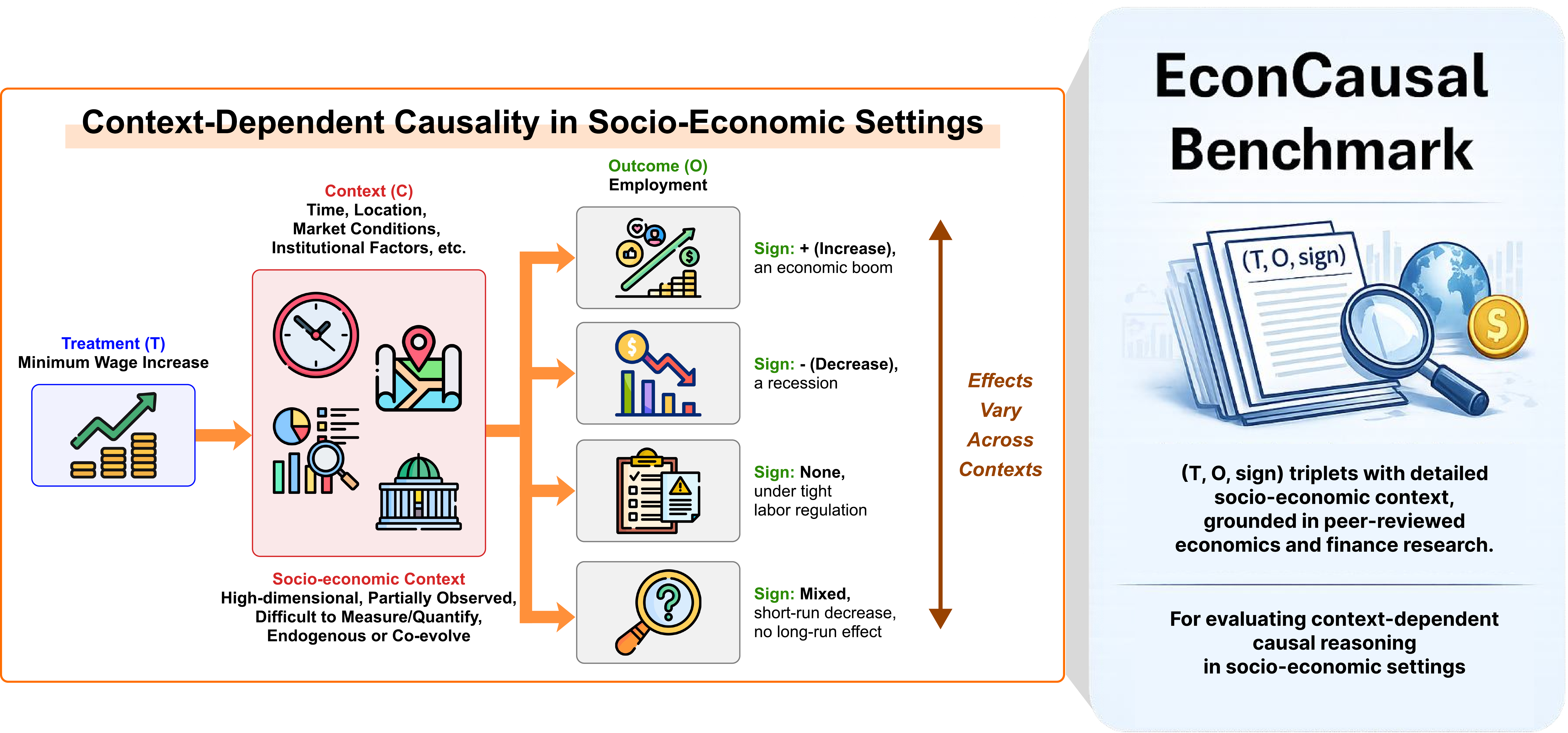}
% \vspace{-3mm}
\caption{Context-dependent causality in socio-economic settings.
Causal effects can vary across high-dimensional and partially observed contexts, posing a challenge for LLM-based reasoning in socio-economic settings.
}
\label{fig:why}
\vspace*{-3mm}
\end{figure*}

Understanding causal relationships in real-world socio-economic settings underpins decisions ranging from government policy and corporate strategy to everyday household choices. As large language models (LLMs) are increasingly integrated into decision-support tools---not only summarizing information but also recommending actions and, in some settings, triggering downstream decisions in semi-autonomous or agentic workflows---a key question is whether they can make reliable judgments in complex socio-economic settings.

Crucially, socio-economic causality is often context-dependent. Socio-economic ``context'' is often high-dimensional, only partially observed, difficult to quantify (e.g., enforcement intensity or informal norms), rarely subject to controlled manipulation, and sometimes endogenous to outcomes, since institutions and behavior can co-evolve with the system being studied. An illustrative example is the employment response to a minimum-wage increase: the sign of the estimated effect can vary across environments shaped by regulation, enforcement, and market conditions. A finding that is valid in one empirical setting may therefore not apply to another, even when the treatment and outcome are similar (Figure~\ref{fig:why}).

This context sensitivity poses a practical challenge for LLM-based decision
support. While LLMs show promise in general reasoning
\cite{kojima2022large,srivastava2023beyond,wang2024mmlu,wei2022chain},
applying them to economic evidence requires predictions that account for
the empirical setting.
% We operationalize context-dependent economic reasoning as prediction of
% study-reported empirical findings.
% Given a socio-economic context ($C$) and a treatment--outcome pair
% ($T$,$O$), a model predicts the effect category reported by the
% corresponding study under its preferred empirical specification.
% We represent each finding as a causal triplet: a treatment--outcome pair
% together with this reported category.
% The task evaluates agreement with documented findings rather than
% identification of causal effects from context alone; it does not assume
% that every reported category is uniquely recoverable from the supplied
% description.
We examine a focused aspect of context-dependent economic
reasoning: predicting empirical effect categories across
settings and using findings from related studies.
Given a socio-economic context $C$ and a treatment--outcome
pair $(T,O)$, a model predicts an effect category.
We score this prediction against the corresponding study's
reported result under its preferred empirical specification,
represented as a context-annotated treatment--outcome--sign
triplet.
This provides a behavioral evaluation of contextual prediction
and evidence use, rather than causal identification.

Existing causal reasoning benchmarks are insufficient
for this setting for three reasons. First, they
provide little coverage of empirically grounded
causal claims in socio-economic contexts---most
focus on generic, context-light relations or
synthetic causal chains rather than evidence-based
findings from the economics and finance literature
\citep{jin2023cladder,wang2024causalbench,
guo2024econnli,quan2024econlogicqa}.
Second, they typically omit the institutional and
empirical context needed to interpret a claim,
treating causal relations as isolated facts
\citep{jin2023cladder,wang2024causalbench}.
Third, they rarely test robustness under context
shifts, particularly whether models can determine
when a valid empirical finding from one setting
should not transfer to another, even though
socio-economic causal effects depend on
high-dimensional, partially observed, and
potentially endogenous contexts
\citep{vivalt2020how,pritchett2015learning}
(Figure~\ref{fig:why}).

To address these limitations, we introduce \textbf{\econcausal{}}, a dataset
and benchmark for evaluating context-dependent economic reasoning through
prediction of study-reported empirical findings.
\econcausal{} represents these findings as context-annotated
treatment--outcome--sign triplets, enabling evaluation of study-result
prediction under explicit socio-economic contexts and the influence of
findings from related settings.

\mypara{Contributions.}
This paper makes three main contributions.
First, we introduce \textbf{\econcausal{}}, a large-scale
\textbf{context-dependent economic reasoning benchmark} built from
high-quality empirical economics and finance research, comprising 10,490
context-annotated treatment--outcome--sign triplets from 2,595 papers.
These triplets are constructed through a rigorous LLM-based pipeline
with multi-run agreement, context refinement, multi-critic filtering,
and expert validation.
Second, we design \textbf{context-shift evaluation tasks} that assess
prediction of study-reported effect categories, the use of signed
findings from related studies, and sensitivity to deliberately corrupted
source labels.
Lastly, experiments across 19 LLMs reveal
\textbf{source-sign influence in cross-context prediction}.
On Task~2 sign-mismatch cases, descriptive comparisons of stored runs show
mean target accuracy 11.54 percentage points lower and source-sign error
rates 15.92 percentage points higher with a signed source example than
with the target input alone.

\section{Related Work}
% \subsection{Causal Reasoning Benchmarks for LLMs}
% Recent literature has increasingly examined the causal reasoning capacities of LLMs, raising concerns that these systems may rely on surface-level linguistic heuristics rather than an internalized grasp of causal structures~\cite{zevceviccausal, niven2020causal, tandon2023barriers}. To address this, benchmarks such as \textbf{CLADDER}~\cite{jin2023cladder} have been developed to probe causal understanding through formal deductive tasks, while \textbf{CausalBench}~\cite{zhou2024causalbench} utilizes everyday 'common-sense' scenarios—such as determining whether adding sugar causes water to become sweet. While these provide well-controlled environments for evaluating counterfactual and interventional reasoning~\cite{huang2024clomo, trichelair2019causal}, they largely operate in decontextualized or synthetic settings. This limitation is particularly critical in socio-economic environments, where causal claims are inextricably linked to institutional frameworks and shifting policy environments. In such domains, the magnitude and directionality of an effect can vary across implicit contexts that are often only partially observed. Consequently, existing benchmarks offer insufficient evidence as to whether LLMs can render robust causal judgments under complex contexts that characterize real-world economic decision-making. \looseness=-1
\subsection{Context-Dependent Reasoning in NLP}

Recent work has studied context-dependent reasoning, where the correct answer or judgment depends on explicitly specified conditions rather than being fixed for an input. Situated and temporal QA benchmarks test whether models can answer questions whose answers vary with time, geography, or evolving facts~\cite{zhang2021situatedqa,chu2024timebench}, while ambiguous QA studies underspecified questions that require additional context to resolve multiple valid interpretations~\cite{min2020ambigqa}. Beyond QA, situated reasoning benchmarks examine whether models adjust their judgments when scenario conditions or normative constraints change~\cite{ashida2022possible,emelin2021moral}.

However, these benchmarks typically use compact contexts centered on a single salient factor, such as time, location, ambiguity, or scenario condition. Multi-dimensional socio-economic contexts---where institutions, markets, policies, regions, populations, and baseline conditions jointly shape how a relationship should be interpreted---remain comparatively underexplored.

\subsection{Economic and Causal Reasoning Benchmarks for LLMs}

Recent work has evaluated economic reasoning in LLMs through domain-specific benchmarks and extraction tools. STEER and STEER-ME~\cite{raman2024steer,raman2026steer} assess microeconomic rationality and game-theoretic optimality, while EconNLI~\cite{guo2024econnli} and EconLogicQA~\cite{quan2024econlogicqa} evaluate inference and sequential reasoning in economic corpora. Large-scale extraction of causal claims has also been used to analyze empirical economics research~\cite{garg2025causal}. Related causal-reasoning benchmarks, such as CLADDER~\cite{jin2023cladder} and CausalBench~\cite{wang2024causalbench}, probe formal or commonsense causal understanding, with complementary work evaluating counterfactual and interventional reasoning in controlled settings~\cite{huang2024clomo}.

However, these benchmarks mainly target rationality, inference, claim extraction, or controlled causal reasoning, rather than context-dependent sign prediction across changing socio-economic settings. 
% EconCausal is complementary but distinct: it does not probe formal causal structure (e.g., graphical models or counterfactual identification), but instead tests whether models apply empirically established signs to the right contexts. It provides context-annotated treatment–outcome–sign triplets grounded in peer-reviewed economics and finance research, and tests whether LLMs can revise causal judgments across institutional, temporal, market, and policy contexts.
\econcausal{} is complementary but distinct: it does not
probe formal causal structure (e.g., graphical models
or counterfactual identification), but instead evaluates
context-sensitive prediction of empirical economic findings.
Its context-annotated treatment--outcome--sign triplets
support comparisons across institutional, temporal,
market, and policy settings, including tests of how
related findings influence target-context predictions.

% \subsection{Economic Reasoning Benchmarks for LLMs}

% Recent work has evaluated economic reasoning in LLMs through domain-specific benchmarks and extraction tools. 
% \textbf{STEER} and \textbf{STEER-ME}~\cite{raman2024steer, raman2025steer} assess microeconomic rationality and game-theoretic optimality, while \textbf{EconNLI}~\cite{guo2024econnli} and \textbf{EconLogicQA}~\cite{quan2024econlogicqa} evaluate inference and sequential reasoning in economic corpora. 
% Large-scale extraction of causal claims has also been used to analyze empirical economics research~\cite{garg2025causal}. 

% However, these benchmarks mainly target rationality, inference, or claim extraction, rather than context-dependent causal sign prediction across changing socio-economic settings. 
% In contrast, \textbf{EconCausal} provides context-annotated treatment--outcome--sign triplets grounded in peer-reviewed economics and finance research, and tests whether LLMs can revise causal judgments across institutional, temporal, market, and policy contexts.

\begin{figure*}[t]
\vspace{-5mm}

\centering
\includegraphics[width=0.9\textwidth]{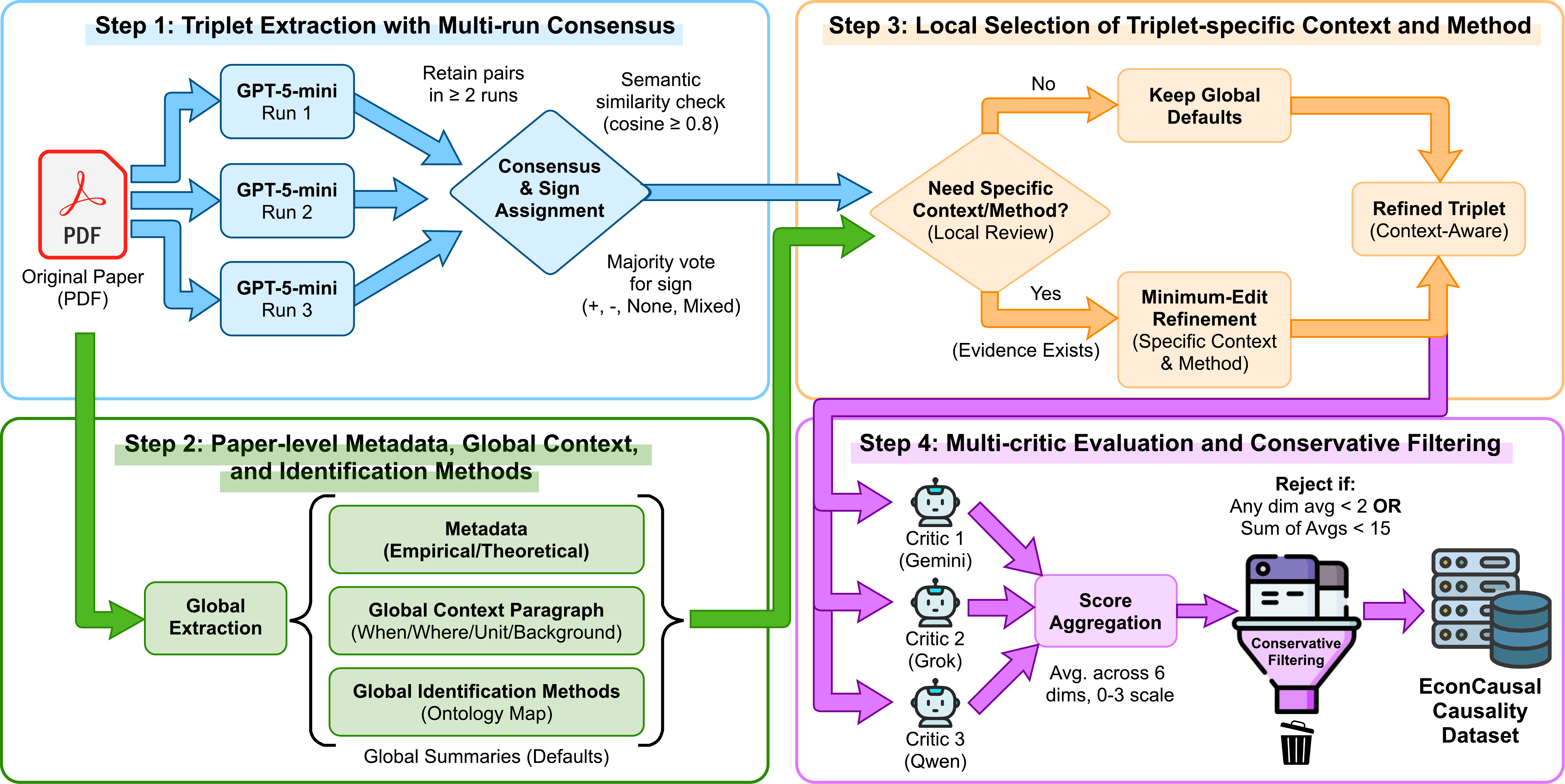}
\caption{Overview of the \econcausal{} dataset extraction pipeline. }
\label{fig:pipeline}
\vspace*{-3mm}
\end{figure*}

\section{\econcausal{}}

\subsection{Causality Dataset Construction}
The dataset is constructed from the \textbf{NBER Working Paper series}~\cite{nber_wp_series} (1991--2025), which serves as the primary preprint platform for influential, policy-relevant research. To prioritize empirical rigor and external validity, the corpus is restricted to working papers subsequently published in the ``top-five'' economics journals---\emph{American Economic Review}, \emph{Econometrica}, \emph{Journal of Political Economy}, \emph{Quarterly Journal of Economics}, and \emph{Review of Economic Studies}---and the ``top-three'' finance journals---\emph{Journal of Finance}, \emph{Journal of Financial Economics}, and \emph{Review of Financial Studies}. These venues are globally recognized for their stringent review standards and rigorous causal identification strategies. This focus aligns with the \textbf{``credibility revolution''} in empirical economics~\cite{angrist2010credibility}, which marked a methodological shift toward explicit identification and the rigorous treatment of endogeneity. We rely on the empirical identification strategies employed in the original studies and use effect directions reported under the authors' preferred specifications as reference labels. Working papers were matched to their final published versions using official NBER metadata, resulting in a curated corpus of \textbf{5,006 papers} spanning more than three decades of research.

\subsection{Dataset Extraction Pipeline}
\label{sec:extraction_pipeline}

We build our causal-relation dataset using a four-stage LLM-based extraction pipeline. Figure~\ref{fig:pipeline} depicts the overall workflow of the data construction process.

\mypara{Step 1: Triplet extraction with multi-run consensus.}
To ensure consistent treatment/outcome extraction, we run \texttt{GPT-5 mini} three independent times per paper and retain treatment--outcome pairs that appear in at least two of the three runs.
Pairs with minor wording differences are merged using cosine similarity over \texttt{text-embedding-3-small} vectors; a pair is considered identical when both the treatment similarity and the outcome similarity reach or exceed $0.8$ (see Appendix~\ref{app:extraction} for details).
For each retained pair, the sign label $\in \{+, \text{ }-, \texttt{ None}, \texttt{ Mixed}\}$ is assigned by majority vote, and up to three supporting evidence paragraphs are stored verbatim for traceability.

\mypara{Step 2: Paper-level metadata, global context, and identification methods.}
We extract paper-level metadata (empirical vs.\ theoretical), a concise global context paragraph summarizing core elements such as setting and unit of analysis, and a deduplicated set of identification methods mapped to a fixed ontology (e.g., DiD, IV, RDD, RCT).
These global summaries serve as defaults for subsequent triplet-level refinement.

\mypara{Step 3: Triplet-specific context and identification method.}
From this step onward, we consider only triplets from papers classified as \emph{empirical} in Step~2, restricting the dataset to causal claims validated by established identification methods.
For each triplet, we verify whether the global context and identification-method summaries remain valid at the claim level.
If the paper explicitly indicates that a given triplet is associated with a different or more specific context or identification method, we apply a minimal edit to produce a triplet-specific context; otherwise, the global defaults are retained unchanged.

\mypara{Step 4: Multi-critic evaluation and conservative filtering.}
We apply an LLM-as-a-critic stage using three independent critic models, each scoring every triplet on six quality dimensions using a 0--3 rubric (see Appendix~\ref{app:critics} for model and rubric details).
Scores are averaged across critics, and triplets failing conservative quality thresholds are removed, filtering out 27.3\% of candidates. \\

In total, the pipeline produces 10,490 context-annotated triplets drawn from 2,595 papers. The source-journal distribution is reported in Appendix~\ref{app:source_distribution}. The full extraction prompts for the four construction stages are provided in Appendix~\ref{app:extraction_prompts}.

% \vspace{-1em}
\subsection{Validation by Economists}

We validate the critic-based filtering by comparing LLM critic scores with evaluations from three economics professors, who assessed 206 triplets within their areas of expertise using the same six criteria. The expert validation suggests that the filtering rule is conservative: while economists approved 83.5\% of the evaluated Step~3 triplets, Step~4 retained 76.2\%, and 86.6\% of the retained triplets were also approved by economists. Criterion-level scores are also high, and further details are provided in Appendix~\ref{app:validation}.

\begin{table*}[t]
\centering
\vspace{-5mm}
\small
\setlength{\tabcolsep}{5pt}
\renewcommand{\arraystretch}{1.22}
\begin{tabular}{@{}p{0.15\textwidth}p{0.41\textwidth}p{0.39\textwidth}@{}}
\toprule
\textbf{Task \#} & \textbf{Question} & \textbf{Evaluation Goal} \\
\midrule

\textbf{Task~1}\newline
\textit{Causal Sign}\newline
\textit{Identification}
& Given context $c$ and a treatment--outcome pair $(T,O)$,
what is the most likely effect category of $T \rightarrow O$?

& Can the model predict the study-reported effect category
under its documented context? (\textbf{RQ1}) \\

\rowcolor{examplegray}
\multicolumn{3}{@{}p{\dimexpr0.95\textwidth+4\tabcolsep\relax}@{}}{%
\footnotesize
\textbf{Example}\quad
Given context $c$: ``minimum-wage increase (\$4.25 to \$5.05) on fast-food establishments in New Jersey in April 1992, \ldots, amid a concurrent regional recession'', \textit{$T$}: minimum-wage increase, \textit{$O$}: employment, what is the causal sign of $T$ to $O$?
} \\
\hline
\addlinespace[0.5em]

\textbf{Task~2}\newline
\textit{Context-Dependent}\newline
\textit{Sign Prediction}
& Given a signed finding under context $c_1$, predict the
effect category for the same or a comparable relation
under context $c_2$.
& Can the model distinguish target-study findings from
related source findings when their reported signs differ? (\textbf{RQ2}) \\

\rowcolor{examplegray}
\multicolumn{3}{@{}p{\dimexpr0.95\textwidth+4\tabcolsep\relax}@{}}{%
\footnotesize
\textbf{Example}\quad
Given the triplet ($T$: higher subsidy, $O$: insurance take-up rate, $+$) under context $c_1$: ``a two-year (2010--2011) randomized field experiment in Jiangxi province, China, \ldots'', what is the sign of the pair ($T$, $O$) under a different context $c_2$: ``Massachusetts' subsidized insurance exchange \textit{CommCare} from 2009--2013, \ldots''?
} \\
\hline
\addlinespace[0.5em]

\textbf{Task~3}\newline
\textit{Misinformation-Robust}\newline
\textit{Sign Prediction}
& Given the triplet ($T$,$O$, sign) under context $c_1$ with its sign deliberately corrupted (noise), what is the causal sign of the target relation under a different context $c_2$?
& Do target-study predictions remain accurate when the
source sign is corrupted? (\textbf{RQ3}) \\

\rowcolor{examplegray}
\multicolumn{3}{@{}p{\dimexpr0.95\textwidth+4\tabcolsep\relax}@{}}{%
\footnotesize
\textbf{Example}\quad
Same as in Task \#2 example, except the given triplet is ($T$: higher subsidy, $O$: insurance take-up rate, $-$), where the sign is incorrect.
} \\

\bottomrule
\end{tabular}
\caption{Overview of the three \econcausal{} benchmark tasks. A complete instantiated example, 
including input contexts and model responses,
is provided in Appendix~\ref{app:task_examples}.}

\label{tab:task-overview}
\vspace*{-3mm}

\end{table*}

\subsection{\econcausal{}'s LLM Benchmark Tasks}

\mypara{Research Questions.}
Our benchmark examines three aspects of context-conditioned prediction of empirical causal-effect signs. First, we ask whether LLMs can predict the effect signs reported in peer-reviewed economics and finance studies, given a treatment--outcome pair and its documented empirical context (\textbf{RQ1}). Second, we examine how a finding from a related source study influences predictions for a target study, particularly when their reported effect signs differ (\textbf{RQ2}). Third, we assess the robustness of target-study predictions when the supplied source-study sign is deliberately corrupted (\textbf{RQ3}).

\mypara{Study-reported causal effect categories.}
Each triplet is labeled using the authors' preferred
empirical specification. Throughout the paper, ``sign''
refers to one of four study-reported categories:
\begin{itemize}
    \item $+$: A statistically significant positive effect
    is reported.
    \item $-$: A statistically significant negative effect
    is reported.
    \item \texttt{None}: No statistically significant effect
    is reported.
    \item \texttt{Mixed}: Heterogeneous effects across subgroups
    or specifications, or multiple equally central results with
    opposite signs, are reported.
\end{itemize}

\mypara{Task descriptions.}
Table~\ref{tab:task-overview} summarizes the three
benchmark tasks; full task prompts are provided in
Appendix~\ref{app:task_prompts}.

\begin{enumerate}
  \item \textbf{Task 1 (Causal Sign Identification).}
  Given a context $c$ and a treatment--outcome pair $(T,O)$, predict the
  effect category reported by the corresponding study for
  $T \rightarrow O$.
  We sample up to 30 triplets per publication year from each domain,
  yielding 947 Economics and 860 Finance questions (1,807 total).

  \item \textbf{Task 2 (Context-Dependent Sign Prediction).}
  Given a source-study finding for a treatment--outcome pair $(T,O)$
  under context $c_1$, predict the effect category reported by a target
  study for the same or an economically comparable pair under
  context $c_2$.
  To construct cross-study pairs, we identify citation-linked papers
  sharing a three-character JEL code, rank candidate relations by the minimum
  treatment/outcome embedding similarity, and apply LLM-based
  comparability filtering.
  This yields 415 source--target pairs.
  Further details on pair construction and economist validation are
  provided in Appendix~E.

  \item \textbf{Task 3 (Misinformation-Robust Sign Prediction).}
  Using the Task~2 input format, replace the source-study sign with an
  incorrect category and predict the target study's reported category.
  We replace the original source sign with each of the remaining three
  labels among $\{+, -, \text{None}, \text{Mixed}\}$, resulting in three
  noisy variants per instance and a total of 1,245 questions.
  Models are not informed of the corruption.
  This task evaluates robustness to corrupted source labels,
  rather than independent verification of the source finding.
\end{enumerate}

\section{Experiments}

\begin{table*}[t]
\centering
\vspace{-7mm}
{\footnotesize
\setlength{\tabcolsep}{3pt}
\renewcommand{\arraystretch}{1.05}
\begin{tabular*}{\textwidth}{@{\extracolsep{\fill}} l c c c c c @{}}
\toprule
 & & & \multicolumn{2}{c}{Task 2} & \\
\cmidrule(lr){4-5}
Model & Task 1 (Econ) & Task 1 (Finance) & Overall & Sign-Mismatch & Task 3 \\
\midrule
\rowcolor{examplegray}
\multicolumn{6}{@{}l}{\textbf{Closed-Source LLMs}} \\
\midrule
Gemini 3.1 Pro & 0.838 (0.609) & \textbf{0.831 (0.606)} & \textbf{0.906 (0.840)} & \textbf{0.806 (0.805)} & \textbf{0.877 (0.733)} \\
Gemini 3 Flash & \textbf{0.839 (0.592)} & 0.821 (0.592) & 0.863 (0.674) & 0.621 (0.566) & 0.736 (0.539) \\
Gemini 2.5 Pro & 0.774 (0.500) & 0.757 (0.480) & 0.776 (0.547) & 0.581 (0.501) & 0.685 (0.455) \\
Gemini 2.5 Flash & 0.752 (0.460) & 0.748 (0.475) & 0.764 (0.494) & 0.476 (0.384) & 0.453 (0.315) \\
GPT-5.2 & 0.729 (0.491) & 0.734 (0.481) & 0.769 (0.538) & 0.476 (0.399) & 0.664 (0.484) \\
GPT-5 mini & 0.699 (0.446) & 0.701 (0.427) & 0.757 (0.507) & 0.427 (0.362) & 0.598 (0.405) \\
GPT-5 nano & 0.682 (0.425) & 0.672 (0.403) & 0.740 (0.466) & 0.387 (0.305) & 0.574 (0.380) \\
GPT-4o & 0.662 (0.409) & 0.658 (0.406) & 0.646 (0.457) & 0.339 (0.296) & 0.404 (0.325) \\
GPT-4o mini & 0.625 (0.363) & 0.592 (0.327) & 0.764 (0.475) & 0.282 (0.215) & 0.500 (0.311) \\
Claude Opus 4.6 & 0.734 (0.529) & 0.762 (0.570) & 0.793 (0.593) & 0.661 (0.597) & 0.742 (0.544) \\
Claude Sonnet 4.6 & 0.747 (0.530) & 0.750 (0.515) & 0.802 (0.584) & 0.613 (0.526) & 0.790 (0.572) \\
Claude Haiku 4.5 & 0.729 (0.466) & 0.719 (0.455) & 0.764 (0.521) & 0.419 (0.367) & 0.619 (0.387) \\
\midrule
Average & 0.734 (0.485) & 0.729 (0.478) & 0.779 (0.558) & 0.507 (0.444) & 0.637 (0.454) \\
\midrule
\rowcolor{examplegray}
\multicolumn{6}{@{}l}{\textbf{Open-Source LLMs}} \\
\midrule
Llama 3.3 70B & 0.715 (0.428) & 0.690 (0.428) & 0.795 (0.504) & 0.395 (0.285) & 0.708 (0.396) \\
Llama 3.1 8B & 0.522 (0.287) & 0.480 (0.282) & 0.658 (0.329) & 0.339 (0.182) & 0.543 (0.278) \\
Llama 3.2 3B & 0.549 (0.265) & 0.501 (0.224) & 0.600 (0.298) & 0.395 (0.220) & 0.533 (0.245) \\
Llama 3.2 1B & 0.570 (0.183) & 0.562 (0.180) & 0.667 (0.292) & 0.306 (0.136) & 0.533 (0.206) \\
Qwen3 32B & 0.638 (0.369) & 0.614 (0.387) & 0.778 (0.463) & 0.371 (0.255) & 0.536 (0.334) \\
Qwen3 14B & 0.580 (0.354) & 0.563 (0.362) & 0.641 (0.439) & 0.379 (0.331) & 0.362 (0.276) \\
Qwen3 8B & 0.719 (0.426) & 0.667 (0.391) & 0.720 (0.451) & 0.137 (0.116) & 0.289 (0.218) \\
\midrule
Average & 0.613 (0.330) & 0.582 (0.322) & 0.694 (0.397) & 0.332 (0.218) & 0.501 (0.279) \\
\bottomrule
\end{tabular*}
\caption{Main results on EconCausal benchmark. We report accuracy (Macro F1) for each task. ``Sign-Mismatch'' denotes the subset where the example sign contradicts the ground-truth sign of the target question. \textbf{Bold} indicates the best in each column.}
\label{tab:main-results}
\vspace{-3mm}
}
\end{table*}
% \subsection{Baseline models}

\subsection{Experimental Settings}

\mypara{Models.}
We evaluate a diverse set of LLMs, spanning both proprietary and open-source systems:
\emph{Gemini family} \cite{comanici2025gemini},
\emph{GPT family} \cite{achiam2023gpt},
\emph{Claude family} \cite{anthropic2026claude},
\emph{Llama family} \cite{grattafiori2024llama},
and \emph{Qwen family} \cite{yang2025qwen3}.

\mypara{Evaluation protocol.}
Each benchmark result is computed from a single model response per test instance under the corresponding task prompt. We use each provider's default decoding settings.

\subsection{Main Results}

\mypara{Overview.}
Table~\ref{tab:main-results} reports model performance
across Tasks~1--3 of the \econcausal{} benchmark.
Paper-cluster bootstrap 95\% confidence intervals
are reported in Appendix~\ref{app:ci-layouts}.
Among proprietary models, Gemini 3 Flash and
Gemini 3.1 Pro achieve the highest Task~1 accuracy
in Economics (83.9\%) and Finance (83.1\%),
respectively, while Gemini 3.1 Pro also leads
Tasks~2 and~3 (90.6\% and 87.7\%).
Aggregate Task~2 accuracy is 77.9\% for closed-source
and 69.4\% for open-source models.
For comparison, simply copying the source sign
achieves 70.12\% accuracy (macro-F1: 0.420),
highlighting the importance of evaluating cases
where source and target signs differ.
Under corrupted source signs in Task~3, the
corresponding averages are lower at 63.7\% and
50.1\%, respectively.
These aggregate results motivate a closer examination
of how models use signed source examples.

\mypara{Performance across sign categories.}
Despite high aggregate accuracy in Tasks~1--2,
macro-F1 scores remain much lower, revealing
systematic class-level imbalance.
Models perform substantially better on directional
signs (\(+\)/\(-\)) than on \texttt{None} and
\texttt{Mixed} cases.
For instance, averaging across the four task rows
in Table~\ref{tab:sign_accuracy_by_task},
accuracy on \texttt{None} is only 19.16\%,
and \texttt{Mixed} reaches just 35.22\%,
both far below 77.58\% for \((+)\) and
64.02\% for \((-)\).
The ground-truth sign distribution is also heavily
skewed toward directional signs
(Figure~\ref{fig:sign_distribution} in the Appendix).
Supplementary analyses of GPT-4o reveal poor
confidence calibration on \texttt{None}/\texttt{Mixed}
and limited abstention when context is removed
(Appendix~\ref{app:calibration}).

\mypara{Economist Reference Study.}
To provide a human reference for interpreting model
performance, we asked three economists to answer ten questions per task with the same inputs and task instructions as the LLMs.
On these questions, economist accuracy was numerically
similar to the closed-source LLM average on Task~2 overall
(70.0\% versus 68.9\%) and higher on sign-mismatch items
(47.2\% versus 36.3\%).
Economists scored lower on both Task~1 domains and Task~3
(23.3\% versus 50.3\% on Task~3).
These results provide a descriptive comparison on a small
sample, with relative performance varying across tasks.
Appendix~\ref{app:economist-eval} reports the results,
agreement on shared questions, and evaluation design.

\subsection{Context-Dependent Transfer Failures}
\label{sec:context_transfer_failures}

\mypara{Sign-mismatch transfer.}
We focus on 124 of 415 Task~2 pairs (29.9\%)
whose comparable treatment--outcome relations have
different reported signs across contexts, including
50 directional reversals ($+\leftrightarrow-$).
These cases test whether models distinguish valid
source findings from their applicability to the target.

\mypara{Source-sign over-transfer.}
We compare Task~2 with a target-only setting that
retains the target context and treatment--outcome
pair but omits the signed source example.

\begin{table}[t]
\centering
\footnotesize
\setlength{\tabcolsep}{1.5pt}
\renewcommand{\arraystretch}{1.03}
\begin{tabular*}{\columnwidth}{@{\extracolsep{\fill}}lrrr@{}}
\toprule
& Correct & \shortstack{Example\\sign error}
          & \shortstack{Other\\error} \\
\midrule
\shortstack[l]{Target only\\(Task~1 setting, $N=415$)}
    & 75.10 & 8.41 & 16.49 \\
\shortstack[l]{With example\\(Task~2, $N=415$)}
    & 74.75 & 13.16 & 12.09 \\
$\Delta$ (pp)
    & $-0.36$ & $+4.76$ & $-4.40$ \\
\midrule
Sign-matched ($N=291$)
    & 87.74 & 0.00 & 12.26 \\
Sign-mismatched ($N=124$)
    & 44.27 & 44.06 & 11.67 \\
\bottomrule
\end{tabular*}
\caption{
Prediction shares (\%) across 19 models, overall (top)
and by source--target sign agreement (bottom).
}
\label{tab:source_overtransfer}
\end{table}

Table~\ref{tab:source_overtransfer} shows little
change in overall accuracy, but errors shift
toward the source sign.
On mismatched pairs, adding the example reduces
accuracy by 11.54\,pp and increases source-sign
errors by 15.92\,pp.
The accuracy decline is larger for directional
reversals ($-17.26$\,pp). Mismatch accuracy stays below
overall accuracy under stricter matching
(Appendix~\ref{app:sensitivity_pair_matching}).

\mypara{Unrelated-source placebo.}
To distinguish source reliance from a generic
effect of extra input, we replace related sources
with unrelated empirical findings while preserving
the joint source--target sign distribution.

\begin{table}[t]
\centering
\footnotesize
\setlength{\tabcolsep}{1.5pt}
\renewcommand{\arraystretch}{1.03}
\begin{tabular*}{\columnwidth}{@{\extracolsep{\fill}}lrrr@{}}
\toprule
& \shortstack{Target\\only}
& \shortstack{Unrelated\\source}
& \shortstack{Related\\source} \\
\midrule
Sign-matched ($N=291$)
    & 83.32 & 79.42 & 87.74 \\
Sign-mismatched ($N=124$)
    & 55.81 & 51.83 & 44.27 \\
\bottomrule
\end{tabular*}
\caption{
Target-sign accuracy (\%) across 19 models.
Sign agreement is defined using the original related source.
}
\label{tab:unrelated_placebo}
\end{table}

Unrelated examples reduce accuracy by about
4\,pp relative to target-only predictions in
both groups (Table~\ref{tab:unrelated_placebo}).
Related examples instead improve accuracy for
matched pairs but worsen it for mismatched pairs,
suggesting reliance on relevant source findings
rather than a uniform cost of additional input.

\mypara{Within-instance source-sign sensitivity.}
Holding source and target descriptions fixed,
we compare the three Task~3 source-sign variants
of each pair.
Across 19 models, predictions differ across
variants in 48.82\% of model--pair instances.
Agreement with the displayed source sign exceeds
its within-instance permutation baseline
(32.29\% vs.\ 17.01\%; $+15.28$\,pp).
The excess persists on 291 pairs without
correction reruns ($+15.37$\,pp;
Appendix~\ref{app:within_instance_source_sign}),
consistent with systematic source-sign anchoring
beyond fixed label preferences.

\subsection{Robustness Analyses}
\label{sec:robustness_analyses}

We examine robustness to pretraining exposure,
prompt and context sensitivity, and lexical shortcuts.

\paragraph{Pretraining exposure and recall.}
On post-cutoff Task~2 papers, accuracy changes by
$-4.67$\,pp (closed) and $+1.01$\,pp (open),
showing no consistent decline
(Appendix~\ref{app:post_cutoff}).
For Task~1, joint bibliographic recall averages
22.7\% (closed) and 0.6\% (open), but shows no
robust association with correctness
(Appendix~\ref{app:source_recall}).
Source-paper citation counts likewise show no
significant association in a fixed-effects regression
(Appendix~\ref{app:citation_regression}).

\mypara{Prompt and context sensitivity.}
SIMPLE and EXPERT prompt variants preserve the
overall performance pattern, with Task~3 remaining
more difficult than Task~2
(Appendix~\ref{app:sensitivity_analyses}).
On 273 unique Task~2 targets, adding economic
context raises accuracy from 67.96\% to 73.61\%
(+5.65~pp) and macro-F1 from 0.383 to 0.480
across 19 models.
In contrast, extending source and target contexts
from 100 to 300 words improves Task~2 accuracy
by only 0.66~pp
(Appendix~\ref{app:context_length_ablation}).

\paragraph{Lexical shortcuts.}
Under paper-grouped nested cross-validation,
TF-IDF logistic regression achieves 0.465/0.473
accuracy in Economics/Finance, below the LLM
means of 0.690/0.675
(Appendix~\ref{app:tfidf_baseline},
Table~\ref{tab:tfidf_baseline}).

\section{Discussion and Future Directions}
% \subsection{Reliability and Uncertainty Awareness}
% Our findings in \S\ref{sec:calibration} reveal a significant gap between model performance and the ability to recognize uncertainty. Despite the inclusion of an explicit ``Unknown'' option, GPT-4o exhibited a profound failure to abstain, attempting to predict causal signs even when all contextual information was removed. This tendency toward over-commitment suggests that, in this GPT-4o case study, the model does not reliably abstain when contextual information is insufficient, instead often producing plausible-sounding directional answers. Such behavior is particularly problematic in economic settings, where insufficient context should ideally trigger a demand for more information rather than a speculative judgment.

% Furthermore, model calibration degrades substantially as causal complexity increases. While models appear relatively well-calibrated for simple directional effects, they are severely miscalibrated for complex categories such as \texttt{None} and \texttt{Mixed}. In these regimes, high confidence scores often mask systematically incorrect predictions, failing to signal when the underlying structural mechanism is ambiguous. These reliability concerns suggest that the model still exhibits important gaps in context-dependent economic reasoning, limiting its reliability for high-stakes economic decision-making.

\subsection{Plausibility versus Portability}
% rag를 적용해도 문제가 생길 수 있다를 강조
The source-sign over-transfer pattern in
Section~\ref{sec:context_transfer_failures}
suggests that the difficulty is not merely
predicting the target sign, but determining
whether a finding from a related study
generalizes to the target context,
echoing evidence-utilization challenges observed
in retrieval-augmented generation (RAG;
\citealp{joren2025sufficient}).
In Task~2, the source sign is a study-reported empirical
finding, not a random distractor.
Models may therefore treat it as a plausible precedent,
even when the target study reports a different effect.

This distinction is central to socio-economic reasoning.
The effects of a treatment need not remain invariant across
settings and populations.
Prior economic research highlights the limitations of generalizing economic causal findings across contexts \citep{vivalt2020how,pritchett2015learning}, while recent NLP studies also reveal limitations in language models' ability to adapt their reasoning to changing contexts \citep{wilie-etal-2024-belief,srikanth-rudinger-2025-nli}.
Thus, the key challenge is not merely generating a plausible
causal sign, but deciding whether that sign applies in the
target context.
The Task~2/3 failures suggest that current LLMs often conflate
plausibility with portability, over-transferring empirical
findings across contexts where they should be revised or rejected.

\subsection{Does More Reasoning Improve Robustness?}
To examine whether reasoning effort and instruction tuning
mitigate source-sign over-transfer, we vary GPT-5 reasoning
effort and Claude thinking on 100-item subsets and compare
Qwen3 Base and Instruct at 8B and 14B.

Minimal-to-low effort improves GPT mismatch and Task~3 accuracy,
but further reasoning offers no consistent gains.
Across GPT low-to-high and Haiku OFF-to-ON comparisons,
mean Task~3 accuracy falls by 9.7~pp as corrupted-sign
agreement rises equally, with little Task~2 change,
paralleling vulnerability to misleading code
hints~\citep{lam2025codecrash}.

In contrast, Qwen3 Instruct improves mean accuracy over Base
by 20~pp on Task~3 versus 5~pp on Task~2, alongside
30.5~pp lower corrupted-sign agreement.
This suggests reduced reliance on misleading source examples,
echoing findings on flipped labels after instruction
tuning~\citep{wei2023larger}.

\subsection{Error Patterns in Model Reasoning}
\label{sec:reasoning_errors}

To examine the reasoning failures underlying
source-sign over-transfer (Section~\ref{sec:context_transfer_failures}),
we classify explanations of up to 100 words from
GPT-5.2, Claude Sonnet 4.6, Gemini 3.1 Pro,
Qwen3 32B, and Llama 3.3 70B into five
non-exclusive error types with GPT-6 Luna (Table~\ref{tab:reasoning_errors}).

\begin{table}[t]
\centering
\vspace{-3mm}
\footnotesize
\setlength{\tabcolsep}{3pt}
\renewcommand{\arraystretch}{1.03}
\begin{tabular*}{\columnwidth}{@{\extracolsep{\fill}}clrr@{}}
\toprule
& Error type & Task~2 & Task~3 \\
\midrule
A & Input/variable misinterpretation & 9.6 & 5.9 \\
B & Unqualified source transfer & 26.1 & 10.4 \\
C & Unsupported study-specific claims & 17.5 & 15.2 \\
D & Invalid economic/logical inference & 7.9 & 4.9 \\
E & Effect-label/output inconsistency & 9.6 & 10.4 \\
\bottomrule
\end{tabular*}
\caption{
Non-exclusive error frequencies (\%) among incorrect predictions.
}
\label{tab:reasoning_errors}
\vspace{-3mm}
\end{table}

In Task~2, unqualified source transfer is the most
frequently identified error (26.1\%).
When correct Task~2 predictions become incorrect
under conflicting source signs in Task~3,
effect-label/output inconsistencies are newly flagged
more often than when predictions remain correct
(13.6\% vs.\ 2.2\%).
Invalid economic/logical inferences show a similar
pattern (5.2\% vs.\ 1.0\%).
These findings suggest that misleading source evidence
is associated not only with shifted predictions,
but also with inconsistencies in their explanations.

\section{Conclusion}

% - (future work 느낌) causal graph + mechanism 단위의 고차원 분석도 추후에 하면 좋을 것 같다.

% - context의 genralization 능력

% \section{Conclusion}
This paper presents \textsc{\econcausal{}}, a large-scale benchmark for evaluating context-dependent reasoning over empirical causal findings in socio-economic domains. Constructed through a systematic four-stage pipeline, it curates context-annotated triplets that link a treatment and outcome to a directional sign within a specific socio-economic context, providing a rigorous foundation for evaluating how LLMs integrate institutional and environmental constraints into their judgments.

Across three tasks requiring increasing contextual generalization, our experiments reveal a persistent gap. While LLMs perform strongly in explicit settings, they exhibit a persistent anchoring effect under context shifts and misleading premises, applying empirically valid signs to contexts where they no longer hold rather than re-evaluating them. These findings suggest that current systems may not yet be robust enough for reliable deployment in high-stakes economic or policy-sensitive analyses. \textsc{\econcausal{}} thus provides a meaningful step toward systems that judge not only whether a causal sign is plausible, but whether it transfers to the context at hand.

\section*{Limitations}

\mypara{Dependence on LLM-assisted extraction and filtering.}
The dataset construction pipeline relies on LLMs for triplet extraction,
context refinement, and multi-critic filtering. Multi-run consensus,
conservative filtering thresholds, and economist validation reduce noise, but
they cannot fully eliminate systematic extraction errors, model-specific biases,
or omissions of subtle identification details. LLM-based annotators may still
simplify nuanced empirical findings or assign contexts that are faithful at a
high level but incomplete for fine-grained causal interpretation.

\mypara{Corpus selection bias.}
\econcausal{} is constructed from NBER working papers that were subsequently
published in top economics and finance journals. This design improves empirical
quality, traceability, and consistency, but may overrepresent influential,
English-language, publication-oriented research. As a result, our benchmark
should be interpreted as a high-quality but selective subset of empirical
economics and finance, rather than the full distribution of socio-economic
causal claims.

\mypara{Granularity of causal labels.}
\econcausal{} represents empirical findings using a compact directional label space:
\texttt{+}, \texttt{-}, \texttt{None}, and \texttt{Mixed}. This abstraction enables
scalable benchmark construction and consistent evaluation, but omits information
such as effect size, economic magnitude, statistical strength, $p$-values, and
confidence intervals. Thus, correctly predicting the sign should not be
interpreted as fully understanding the empirical result.

\mypara{Compression of heterogeneous effects.}
The current \texttt{Mixed} label compresses several distinct empirical patterns
into a single category. Mixed effects may arise from heterogeneity across groups,
sign changes across specifications or institutional settings, nonlinear
responses, or temporal reversals. Future versions could split \texttt{Mixed}
into interpretable subtypes and record the dimension along which the effect
varies.

\mypara{Scope of context and external validity.}
Finally, the context field in \econcausal{} is tied to the specific empirical
setting of the source paper. This improves grounding and traceability, but limits
our ability to evaluate external validity directly. Our context-shift tasks test
whether models revise causal signs across paired settings, but they do not
explicitly annotate portability-relevant features such as institutions, market
structure, baseline conditions, or enforcement environment.

\mypara{Extending the human reference.}
The economist evaluation study
(Appendix~\ref{app:economist-eval}) provides a descriptive
comparison based on a small sample of economists and
questions.
Larger studies with systematic matching between question
topics and participants' specializations could provide
more precise human references and clarify how domain
familiarity relates to performance.
Such studies could further examine when relevant
expertise supports appropriate use of source evidence,
helping characterize the challenges that humans and
LLMs face in applying empirical findings across contexts.

\mypara{Behavioral scope of evaluation.} Our evaluation is primarily behavioral. Beyond output predictions and their associated confidence (Appendix~\ref{app:calibration}), we do not probe model internals such as intermediate representations or attention, nor attribute failures to specific computational mechanisms. Connecting these patterns to internal model processes is a direction for future work.

\mypara{Richer Context and Controlled Comparisons.}
Our robustness analyses provide initial evidence on the role of contextual information in causal sign prediction. Providing target context improves accuracy by 5.65 percentage points, while extending source and target contexts from 100 to 300 words yields only a modest additional gain of 0.66 percentage points (Appendix~H.5). These findings suggest that contextual information is valuable, but that simply increasing its length has limited impact on prediction performance. Building on these results, future work could explore more targeted contextual information, including detailed identification strategies and sample characteristics, to better understand which information is most useful for predicting empirical findings. Additionally, constructing more tightly controlled cross-study comparisons could help disentangle genuine contextual differences from variations in estimands or research designs, enabling a more fine-grained evaluation of context-dependent reasoning.

\section*{Ethical Considerations}
\econcausal{} is constructed from scholarly economics and finance papers, including NBER working papers and their matched published versions. NBER does not hold the copyright for its Working Paper Series; rather, copyright remains with the authors of each paper. We therefore use the source papers solely for research purposes and provide attribution through bibliographic metadata and source identifiers. We do not redistribute source PDFs, full-text papers, or verbatim excerpts from the original articles. Any verbatim evidence used during dataset construction is retained only for internal traceability and quality control, and is excluded from the public release. The released \econcausal{} artifact will be distributed under the Creative Commons Attribution-NonCommercial 4.0 International License (CC BY-NC 4.0).

\econcausal{} is intended as a research benchmark for evaluating context-dependent causal reasoning in language models. It should not be used as a substitute for reading the original empirical studies, nor as a basis for policy, financial, legal, medical, or investment decisions. Since causal signs in socio-economic settings are context-dependent and may omit effect sizes, uncertainty, and heterogeneous mechanisms, systems evaluated on this benchmark should not be deployed in high-stakes decision-making settings without expert review and source verification.

% Bibliography entries for the entire Anthology, followed by custom entries
%\bibliography{anthology,custom}
% Custom bibliography entries only
\bibliography{custom}

\begin{thebibliography}{35}
\providecommand{\natexlab}[1]{#1}

\bibitem[{Angrist and Pischke(2010)}]{angrist2010credibility}
Joshua~D. Angrist and J{\"o}rn-Steffen Pischke. 2010.
\newblock \href {https://doi.org/10.1257/jep.24.2.3} {The credibility revolution in empirical economics: How better research design is taking the con out of econometrics}.
\newblock \emph{Journal of Economic Perspectives}, 24(2):3--30.

\bibitem[{{Anthropic}(2026)}]{anthropic2026claude}
{Anthropic}. 2026.
\newblock \href {https://www.anthropic.com/system-cards} {Model system cards}.
\newblock Accessed: 2026-10-12.

\bibitem[{Ashida and Sugawara(2022)}]{ashida2022possible}
Mana Ashida and Saku Sugawara. 2022.
\newblock \href {https://aclanthology.org/2022.coling-1.319/} {{Possible Stories}: Evaluating situated commonsense reasoning under multiple possible scenarios}.
\newblock In \emph{Proceedings of the 29th International Conference on Computational Linguistics}, pages 3606--3630, Gyeongju, Republic of Korea. International Committee on Computational Linguistics.

\bibitem[{Chu et~al.(2024)Chu, Chen, Chen, Yu, Wang, Liu, and Qin}]{chu2024timebench}
Zheng Chu, Jingchang Chen, Qianglong Chen, Weijiang Yu, Haotian Wang, Ming Liu, and Bing Qin. 2024.
\newblock \href {https://doi.org/10.18653/v1/2024.acl-long.66} {{TimeBench}: A comprehensive evaluation of temporal reasoning abilities in large language models}.
\newblock In \emph{Proceedings of the 62nd Annual Meeting of the Association for Computational Linguistics (Volume 1: Long Papers)}, pages 1204--1228, Bangkok, Thailand. Association for Computational Linguistics.

\bibitem[{Comanici et~al.(2025)Comanici, Bieber, Schaekermann, Pasupat, Sachdeva, Dhillon, Blistein, Ram, Zhang, Rosen, Marris, Petulla, Gaffney, Aharoni, Lintz, Pais, Jacobsson, Szpektor, Jiang, Haridasan, Omran, Saunshi, Bahri, Mishra, Chu, Boyd, Hekman, Parisi, Zhang, Kawintiranon, Bedrax-Weiss, Wang, Xu, Purkiss, Mendlovic, Deutel, Nguyen, Langley, Korn, Rossazza, Ramé, Waghmare, Miller, Byrd, Sheshan, Hadsell, Bhardwaj, Janus, Rissa, Horgan, Abdagic, Belenki, Allingham, Singh, Guidroz, Srinivasan, Schmit, Chiafullo, Elisseeff, Jha, Kolhar, Berrada, Ding, Si, Mallick, Och, Erell, Ni, Latkar, Yang, Sirkovic, Feng, Leland, Hornung, Wu, Blundell, Alvari, Huang, Yip, Deur, Liu, Surita, Duque, Damen, Jia, Guez, Mircea, Sinha, Magni, Stradomski, Marian, Galić, Chen, Husain, Singhal, Grewe, Aubet, Song, Blanco, Rechis, Ho, Munoz, Zheng, Hamrick, Mather, Taitelbaum, Rutherford, Lei, Chen, Shukla, Moreira, Doi, Isik, Shabat, Rogozińska, Kolipaka, Chang, Vušak, Venkatachary, Noghabi, Bharti, Jun, Zaks, Green,
  Challagundla, Wong, Mohammad, Hirsch, Cheng, Naim, Proleev, Vincent, Singh, Krikun, Krishnan, Ghahramani, Atias, Aggarwal, Kirov, Vytiniotis, Koh, Chronopoulou, Dogra, Ion, Tyen, Lee, Weissenberger, Strohman, Balakrishna, Rae, Velic, de~Liedekerke, Elyada, Yuan, Liu, Shani, Kishchenko, Alessio, Li, Song, Kwei, Jankowski, Pappu, Namiki, Ma, Tripuraneni, Cherry, Ikonomidis, Ling, Ji, Westberg, Wright, Yu, Parkinson, Ramaswamy, Connor, Yeganeh, Grover, Kenwright, Litchev, Apps, Tomala, Halim, Castro-Ros, Li, Boral, Sho, Yarom, Malmi, Klinghoffer, Lin, Ansell, S, Zhao, Zuo, Santoro, Cheng, Demmessie, Liu, Brichtova, Culp, Braun, Graur, Ng, Mehta, Phillips, Sundberg, Godbole, Liu, Katariya, Rim, Seyedhosseini, Ammirati, Valfridsson, Malihi, Knight, Toor, Lampe, Ittycheriah, Chiang, Yeung, Fréchette, Rao, Wang, Srivastava, Zhang, Rhodes, Brand, Weesner, Figotin, Gimeno, Fellinger, Marcenac, Leal, Marcus, Cotruta, Cabrera, Luo, Garrette, Axelrod, Baltateanu, Barker, Chen, Toma, Ingram, Riesa, Kulkarni, Zhang,
  Liu, Wang, Polacek, Wu, Hui, Reyes, Su, Barnes, Malhi, Siddiqui, Feng, Damaschin, Pighin, Steiner, Yang, Boppana, Ivanov, Kandoor, Shah, Mujika, Huang, Choquette-Choo, Patel, Yu, Creswell, Liu, Barros, Razeghi, Roy, Culliton, Xiong, Pan, Strohmann, Powell, Seal, DeCarlo, Shyam, Katircioglu, Wang, Hardin, Odisho, Broder, Chang, Nair, Shtefan, O'Brien, Agarwal, Potluri, Goyal, Jhindal, Thakur, Stuken, Lyon, Toutanova, Feng, Wu, Horn, Wang, Cullum, Taubman, Shrivastava, Shi, Tomlinson, Patel, Tu, Oflazer, Pongetti, Yang, Taïga, Perot, Pierse, Han, Drori, Iturrate, Chakrabarti, Yeung, Dopson, ting Chen, Kulshreshtha, Guo, Pham, Schuster, Chen, Polozov, Xing, Zhou, Kacham, Kukliansky, Miech, Yaroshenko, Chi, Douglas, Fei, Blondel, Myla, Madmoni, Wu, Keysers, Kjems, Albuquerque, Yu, D'sa, Plantan, Ionescu, Elias, Gupta, Vuyyuru, Alcober, Zhou, Ji, Hartmann, Puttagunta, Song, Amid, Stefanoiu, Lee, Pucciarelli, Wang, Raul, Petrov, Tian, Anklin, Nti, Gomes, Schumacher, Vesom, Panagopoulos, Bousmalis, Andor, Jacob,
  Zhang, Rosgen, Kecman, Tung, Belias, Goodman, Covington, Wieder, Saxena, Davoodi, Huang, Maddineni, Roulet, Campbell-Ajala, Sessa, Wu, Lai, Collins, Haig, Sakenas, Xu, Giustina, Shafey, Charoenpanit, Garg, Ainslie, Severson, Arenas, Pathak, Rajayogam, Feng, Bakker, Li, Wichers, Rogers, Geng, Li, Jagerman, Jia, Olmert, Sharon, Mauger, Mariserla, Ma, Mohabey, Kim, Andreev, Pollom, Love, Jain, Agrawal, Schroecker, Fortin, Warmuth, Liu, Leach, Blok, Girirajan, Aharoni, Uria, Sozanschi, Goldberg, Ionita, Ribeiro, Zlocha, Birodkar, Lachgar, Yuan, Choudhury, Ginsberg, Zheng, Dibb, Graves, Lokhande, Rasskin, Muraru, Quick, Tata, Sermanet, Chawla, Karo, Wang, Zhang, Keller, Dragan, Su, Chou, Liu, Tao, Prabhakara, Wilson, Liu, Wang, Evans, Du, Castaño, Prasad, Mahdy, Gerlach, Reid, Kahn, Zait, Pillai, Ulrich, Wang, Wassenberg, Farkash, Yalasangi, Wang, Bauza, Bucher, Liu, Yan, Leung, Sindhwani, Barnes, Singh, Jurin, Chang, Bhumihar, Eiger, Citovsky, Withbroe, Li, Xue, Santo, Stoyanov, Raimond, Zheng, Gao, Listík,
  Kwasiborski, Saputro, Ozturel, Mallya, Majmundar, West, Caron, Wei, Castrejon, Vikram, Ramachandran, Dhawan, Park, Smoot, van~den Driessche, Blau, Malik, Liang, Hirsch, dos Santos, Weinstein, van~den Oord, Lall, FitzGerald, Jiang, Yang, Webster, Elqursh, Pope, Rotival, Raposo, Zhu, Dean, Alabed, Tran, Gupta, Gleicher, Austin, Rosseel, Umekar, Das, Sun, Chen, Misiunas, Zhou, Di, Loo, Newlan, Li, Ramasesh, Xu, Chen, Gandhe, Soricut, Gupta, Hu, El-Sayed, Garcia, Brusilovsky, Chen, Bolt, Huang, Gurney, Zhang, Pritzel, Wilkiewicz, Seybold, Shamanna, Fischer, Dean, Gill, Mcilroy, Bhowmick, Selier, Yang, Cheng, Magay, Tan, Varma, Walder, Kocisky, Nakashima, Natsev, Kwong, Gog, Zhang, Dieleman, Jimma, Ryabtsev, Brahma, Steiner, Du, Žužul, Žanić, Raghavachari, Gierke, Zheng, Petrova, Dauphin, Liu, Kessler, Hand, Duvarney, Kim, Lee, Hussenot, Hui, Smith, Jain, Xia, Tomar, Amiri, Phan, Fuchs, Weyand, Tomasev, Cordell, Liu, Mallinson, Joshi, Crawford, Suggala, Chien, Fernando, Sanchez-Vargas, Williams, Crone, Luo,
  Karpov, Shan, Thurk, Strudel, Voigtlaender, Patil, Dozat, Khodaei, Singla, Ambroszczyk, Wu, Chang, Roark, Hegde, Ding, Filos, Wu, Pinto, Liu, Khanna, Pandey, Mcloughlin, Li, Haves, Zhou, Buchatskaya, Leal, de~Boursac, Akazawa, Anderson, Chen, Somandepalli, Liang, Goenka, Winkler, Grushetsky, Ding, Smith, Ye, Pont-Tuset, Li, Li, Golany, Wegner, Jiang, Barak, Shangguan, Vértes, Wong, Bornschein, Tudor, Bevilacqua, Schaul, Rawat, Zhao, Axiotis, Meng, McLean, Lai, Beattie, Kushman, Liu, Kutzman, Lang, Ye, Netrapalli, Mishra, Khan, Goel, Willoughby, Tian, Zhuang, Chen, Tsai, Kementsietsidis, Khare, Keeling, Xu, Waters, Altché, Popat, Mittal, Saxton, Badawy, Mathieu, Zheng, Zhou, Ranka, Shin, Duan, Salimans, Mihailescu, Shaham, Chang, Assael, Dikkala, Izzard, Cohen-Addad, Graves, Feinberg, Chung, Strouse, Karmon, Sharifzadeh, Ashwood, Pham, Blanton, Vasiloff, Barber, Geller, Zhou, Zubach, Huang, Zhang, Gupta, Young, Proskurnia, Votel, Gabeur, Barcik, Tripathi, Yu, Yan, Changpinyo, Pavetić, Coyle, Fujii,
  Mendez, Zhou, Rajamani, Hechtman, Cao, Juan, Tan, Dalibard, Du, Clay, Yao, Jia, Vijaykumar, Zhou, Bai, Hung, Pecht, Todorov, Khadke, Gupta, Lahoti, Autef, Duddu, Lee-Thorp, Bykovsky, Misiunas, Flennerhag, Thangaraj, McGiffin, Nado, Kunesch, Noever, Hertz, Liang, Stone, Palmer, Daruki, Pramanik, Põder, Kyker, Khan, Sluzhaev, Ritter, Ruderman, Zhou, Nagpal, Vodrahalli, Necula, Barham, Pavlick, Hartford, Shafran, Zhao, Mikuła, Eccles, Shimokawa, Garg, Vilnis, Chen, Shumailov, Lee, Abdelhamed, Xie, Cohen, Hlavnova, Malkin, Sitawarin, Lottes, Coquinot, Yu, Kumar, Zhang, Mahendru, Ahmed, Martens, Chen, Boag, Peng, Devin, Klimovskiy, Phuong, Vainstein, Xie, Ramabhadran, Howard, Yu, Goswami, Cui, Shleifer, Pinto, Yeh, Yang, Javanmardi, Ethier, Lee, Orbay, Kotecha, Bromberg, Shaw, Thornton, Rosenthal, Gu, Thomas, Gemp, Ayyar, Ushio, Selvan, Wee, Liu, Majzoubi, Yu, Abernethy, Liechty, Pan, Nguyen, Hu, Perrin, Arora, Pitler, Wang, Shivakumar, Prost, Limonchik, Wang, Gao, Cour, Buch, Gui, Ivanova, Neubeck, Chan, Kim,
  Chen, Goyal, Chung, Liu, Su, Petrushkina, Shen, Joulin, Xu, Lin, Kulizhskaya, Chelba, Vasudevan, Collins, Bashlovkina, Lu, Fritz, Park, Zhou, Su, Tanburn, Sushkov, Rasquinha, Li, Prendki, Li, LV, Sharma, Fitoussi, Huang, Dai, Dao, Burrows, Prior, Qin, Pundak, Sjoesund, Khurshudov, Zhu, Webson, Kemp, Tan, Agrawal, Sargsyan, Cheng, Stephan, Kwiatkowski, Reid, Byravan, Michaely, Heess, Zhou, Goenka, Carpenter, Levskaya, Wang, Roberts, Leblond, Chikkerur, Ginzburg, Chang, Riachi, Xu, Borsos, Pliskin, Pawar, Lustman, Kirkwood, Anand, Chaudhary, Kalb, Milan, Augenstein, Goldie, Prince, Raman, Sun, Xia, Cohen, Huo, Camp, Ellis, Zilka, Torres, Patel, Arora, Chan, Adler, Ayoub, Liang, Jamil, Jiang, Baumgartner, Sun, Karov, Akulov, Zheng, Cai, Fantacci, Rubin, Acha, Wang, D'Souza, Sathyanarayana, Dai, Rowe, Simanovsky, Goldman, Kuang, Pan, Rosenberg, Rojas-Esponda, Dutta, Zeng, Jurenka, Farquhar, Bansal, Iqbal, Roelofs, Joung, Beak, Ryu, Poplin, Wu, Alayrac, Buthpitiya, Ronneberger, Habtegebriel, Li, Cavallaro, Wei,
  Bensky, Denk, Ganapathy, Stanway, Joshi, Bertolini, Lo, Ma, Charles, Sampemane, Sahni, Chen, Askham, Gaddy, Young, Tan, Eyal, Bražinskas, Zhong, Wu, Epstein, Bailey, Hard, Lee, Goldshtein, Ruiz, Badawi, Lochbrunner, Kearns, Brown, Pardo, Weber, Yang, Jiang, Akin, Fu, Wainwright, Zou, Gaba, Manzagol, Kan, Song, Zainullina, Lin, Ko, Deshmukh, Jindal, Svensson, Tyam, Zhao, Kaeser-Chen, Baird, Moradi, Hall, Guo, Tsang, Liang, Pereira, Ganesh, Korotkov, Adamek, Thiagarajan, Tran, Chen, Tar, Jain, Dasgupta, Bilal, Reitter, Zhao, Vezzani, Gehman, Mehta, Beltrone, Dotiwalla, Guadarrama, Abbas, Karp, Georgiev, Ferng, Brockschmidt, Peng, Hirnschall, Verma, Bi, Xiao, Dabush, Xu, Wallis, Parker, Wang, Xu, Safarli, Tewari, Zhang, Kim, Gesmundo, Thomas, Levi, Chowdhury, Rao, Garst, Conway-Rahman, Ran, McKinney, Xiao, Yu, Agrawal, Stjerngren, Ionescu, Chen, Sharma, Chiu, Liu, Franko, Sanford, Cai, Michel, Ganapathy, Labanowski, Garrett, Vargas, Sun, Gale, Buschmann, Desjardins, Ghelani, Jain, Verma, Asawaroengchai,
  Eisenschlos, Harlalka, Kazawa, Metzler, Howland, Jian, Ades, Shah, Gangwani, Lee, Ring, Hernandez, Reich, Sinha, Sathe, Kovac, Gill, Kannan, D'olimpio, Sevenich, Whang, Kim, Sim, Chen, Zhang, Lall, Matias, Jia, Friesen, Nasso, Thapliyal, Perozzi, Yu, Shekhawat, Huda, Grabowski, Wang, Sreevatsa, Dib, Hassen, Schuh, Milutinovic, Welty, Quinn, Shah, Wang, Barth-Maron, Frye, Axelsson, Zhu, Ma, Giannoumis, Sedghi, Ye, Luan, Aydin, Chandra, Sampathkumar, Huang, Lavrenko, Eleryan, Hong, Hansen, Carthy, Samanta, Ćevid, Wang, Li, Voznesensky, Hoffman, Terzis, Sehwag, Fidel, He, Cai, He, Feng, Nikoltchev, Phatale, Chase, Lawton, Zhang, Ouyang, Tragut, Manshadi, Narayanan, Shen, Gao, Bolukbasi, Roy, Li, Golovin, Panait, Qin, Han, Anthony, Kudugunta, Patraucean, Ray, Chen, Yang, Bhatia, Talluri, Morris, Ražnatović, Brownfield, An, Peng, Kane, Zheng, Duduta, Kessinger, Noraky, Liu, Rong, Veličković, Rush, Goldin, Wei, Garlapati, Pantofaru, Kwon, Ni, Noland, Trapani, Beaufays, Roy, Chow, Turker, Cideron, Mei, Clark,
  Dou, Bošnjak, Leith, Du, Yazdanbakhsh, Nasr, Kwak, Sheth, Kaskasoli, Anand, Lakshminarayanan, Jerome, Bieber, Chu, Senges, Shen, Sridhar, Ndebele, Beyret, Mohamed, Chen, Freitag, Guo, Liu, Roit, Chen, Yan, Stone, Co-Reyes, Cole, Scellato, Azizi, Hashemi, Jin, Iyer, Valentine, György, Ahuja, Diaz, Lee, Clement, Kong, Garmon, Watts, Bhatia, Gupta, Miecnikowski, Vallet, Taly, Loper, Joshi, Atwood, Chick, Collier, Iliopoulos, Trostle, Gunel, Leal-Cavazos, Hrafnkelsson, Guzman, Ju, Forbes, Emond, Chauhan, Caine, Xiao, Zeng, Moufarek, Murphy, Meng, Gupta, Riedel, Das, Lawal, Narayan, Sosea, Swirhun, Friso, Neyshabur, Lu, Girgin, Wunder, Yvinec, Pyne, Carbune, Rijhwani, Guo, Doshi, Briukhov, Bain, Hitron, Wang, Gupta, Chen, Du, Zhang, Shah, Akula, Dylla, Kachra, Kuo, Zou, Wang, Xu, Zhu, Snyder, Menon, Firat, Mordatch, Yuan, Ponomareva, Blevins, Moore, Wang, Chen, Scholz, Dwornik, Lin, Li, Antognini, I, Song, Miller, Kalra, Raveret, Akerlund, Wu, Nystrom, Godbole, Liu, DeBalsi, Zhao, Liu, Caciularu, Lax,
  Khandelwal, Langston, Bailey, Lattanzi, Wang, Kovelamudi, Mondal, Guruganesh, Hua, Roval, Wesołowski, Ingale, Halcrow, Sohn, Angermueller, Raad, Stickgold, Lu, Kosik, Xie, Lillicrap, Huang, Zhang, Paulus, Farabet, Wertheim, Wang, Joshi, ling Ko, Wu, Agrawal, Lin, Sheng, Sung, Breland-King, Butterfield, Gawde, Singh, Zhang, Apte, Shetty, Hutter, Li, Salesky, Lebron, Kanerva, Paganini, Nguyen, Vallu, Peter, Velury, Kao, Hoover, Bortsova, Bishop, Jakobovits, Agostini, Agarwal, Liu, Kwong, Tavakkol, Bica, Greve, GP, Marcus, Hou, Duerig, Moroshko, Lacey, Davis, Amelot, Wang, Kim, Strinopoulos, Wan, Lan, Krishnan, Tang, Humphreys, Bai, Shtacher, Machado, Pang, Burke, Liu, Aravamudhan, Song, Hirst, Singh, Jou, Bai, Piccinno, Fu, Alazard, Meiri, Winter, Chen, Zhang, Heitkaemper, Lambert, Lee, Frömmgen, Rogulenko, Nair, Niemczyk, Bulyenov, Xu, Shemtov, Zadimoghaddam, Toropov, Wirth, Dai, Gollapudi, Zheng, Kurakin, Lee, Bullard, Serrano, Balazevic, Li, Schalkwyk, Murphy, Zhang, Sequeira, Datta, Agrawal, Sutton,
  Attaluri, Chiang, Farhan, Thornton, Lin, Choma, Nguyen, Dasgupta, Robinson, Comşa, Riley, Pillai, Mustafa, Golan, Zandieh, Lespiau, Porter, Ross, Rajayogam, Agarwal, Venugopalan, Shahriari, Yan, Xu, Tobin, Dubov, Shi, Recasens, Kovsharov, Borgeaud, Dery, Vasanth, Gribovskaya, Qiu, Mahdieh, Skut, Nielsen, Zheng, Yu, Bostock, Gupta, Archer, Rawles, Davies, Svyatkovskiy, Tsai, Halpern, Reisswig, Wydrowski, Chang, Puigcerver, Taege, Li, Schnider, Li, Dena, Xu, Telang, Shi, Zen, Kastner, Ko, Subramaniam, Kumar, Blois, Dai, Wieting, Lu, Zeldes, Xie, Hauth, Ţifrea, Li, El-Husseini, Abolafia, Zhou, Ding, Ghalebikesabi, Guía, Maksai, Ágoston Weisz, Arik, Sukhanov, Świetlik, Jia, Yu, Wang, Brand, Bloxwich, Kirmani, Chen, Go, Sprechmann, Kannen, Carin, Sandhu, Edkins, Nooteboom, Gupta, Maggiore, Azizi, Pritch, Yin, Gupta, Tarlow, Smith, Ivanov, Babaeizadeh, Goel, Kambala, Chu, Kastelic, Liu, Soltau, Stone, Agrawal, Kim, Soparkar, Tadepalli, Bunyan, Soh, Kannan, Kim, Chen, Halumi, Roy, Wang, Sercinoglu, Gibson,
  Bhatnagar, Sano, von Dincklage, Ren, Mitrevski, Olšák, She, Doersch, Wang, Liu, Tan, Yakar, Warkentin, Ramirez, Lebsack, Dillon, Mathews, Cobley, Wu, Chen, Simon, Nath, Sainath, Bendebury, Julian, Mankalale, Ćurko, Zacchello, Brown, Sodhia, Howard, Caelles, Gupta, Evans, Bulanova, Katzen, Goldenberg, Tsitsulin, Stanton, Schillings, Kovalev, Fry, Shah, Lin, Upadhyay, Li, Radpour, Maggioni, Xiong, Haas, Brennan, Kamath, Savinov, Nagrani, Yacovone, Kappedal, Andriopoulos, Lao, Li, Rozhdestvenskiy, Hashimoto, Audibert, Austin, Rodriguez, Ruoss, Honke, Karkhanis, Xiong, Wei, Huang, Leng, Premachandran, Bileschi, Evangelopoulos, Mensink, Pavagadhi, Teplyashin, Chang, Xue, Tanzer, Goldman, Patel, Li, Wiesner, Zheng, Stewart-Binks, Han, Li, Luo, Lenc, Lučić, Xue, Mullins, Guseynov, Chang, Galatzer-Levy, Zhang, Bingham, Hu, Hartman, Ma, Griffith, Irpan, Radebaugh, Yue, Fan, Ungureanu, Sorokin, Teufel, Li, Anil, Paparas, Wang, Lin, Peng, Shum, Petrovic, Brady, Nguyen, Macherey, Li, Singh, Yenugula, Iinuma, Chen,
  Kopparapu, Stern, Dave, Thekkath, Perot, Kumar, Li, Xiao, Bilotti, Bateni, Noble, Lee, Vázquez-Reina, Salazar, Yang, Wang, Gruzewska, Rao, Raghuram, Xu, Ben-David, Mei, Dalmia, Zhang, Liu, Bansal, Pankov, Schwarcz, Burns, Chan, Sanghai, Liang, Liang, He, Stuart, Narayanan, Zhu, Frank, Fatemi, Sabne, Lang, Bhattacharya, Settle, Wang, McMahan, Tacchetti, Soares, Hadian, Cabi, Chung, Putikhin, Li, Chen, Tarango, Michalewski, Kazemi, Masoom, Sheftel, Shivanna, Vadali, Comanescu, Reid, Moore, Neelakantan, Sander, Herzig, Rosenberg, Dehghani, Choi, Fink, Hayes, Ge, Weng, Ho, Karro, Krishna, Thiet, Skerry-Ryan, Eppens, Andreetto, Sarma, Bonacina, Ayan, Nawhal, Shan, Dusenberry, Thakoor, Gubbi, Nguyen, Tsarfaty, Albanie, Mitrović, Gandhi, Chen, Epasto, Stephanov, Jin, Gehman, Amini, Weber, Behbahani, Xu, Allamanis, Chen, Ott, Sha, Jastrzebski, Qi, Greene, Wu, Toki, Vlasic, Shapiro, Kotikalapudi, Shen, Saeki, Xie, Cassirer, Bharadwaj, Kiyono, Bhojanapalli, Rosenfeld, Ritter, Mao, Oliveira, Egyed, Bandemer,
  Parisotto, Kinoshita, Pluto, Maniatis, Li, Guo, Ghiasi, Tarbouriech, Chatterjee, Jin, Xu, Palomaki, Arnold, Sewak, Piccinini, Sharma, Albrecht, Purser-haskell, Vaswani, Chen, Wisniewski, Cao, Aslanides, Phu, Sieb, Agubuzu, Zheng, Sohn, Selvi, Andreassen, Subudhi, Eruvbetine, Woodman, Mery, Krause, Ren, Ma, Luo, Chen, Fan, Griffiths, Schuler, Li, Zhang, Sarr, Luo, Patana, Watson, Naboulsi, Collins, Sidhwani, Hoogeboom, Silver, Caveness, Zhao, Rodriguez, Deines, Bai, Griffin, Tagliasacchi, Xue, Babbula, Pang, Ding, Shen, Peake, Crocker, Raghvendra, Swisher, Han, Singh, Wu, Pchelin, Munkhdalai, Alon, Bacon, Robles, Bulian, Johnson, Powell, Ferreira, Li, Benzing, Velimirović, Soyer, Kong, Nguyên, Yang, Liu, van Amersfoort, Gillick, Sun, Rauschmayr, Zhang, Zhan, Zhou, Frolov, Yang, Vnukov, Rouillard, Li, Mandhane, Fallen, Venkataraman, Hu, Brennan, Lee, Chang, Sundermeyer, Pan, Ke, Tong, Fabrikant, Bono, Gu, Foley, Mao, Delakis, Bhaswar, Frostig, Li, Zipori, Hope, Kozlova, Mishra, Djolonga, Schiff, Merey,
  Briakou, Morgan, Wan, Hassidim, Skerry-Ryan, Sengupta, Jasarevic, Kallakuri, Kunkle, Brennan, Lieber, Mansoor, Walker, Zhang, Xie, Žužić, Chukwuka, Druinsky, Cho, Yao, Naeem, Butt, Kim, Jia, Jordan, Lelkes, Kurzeja, Wang, Zhao, Over, Chakladar, Prasetya, Jha, Ganapathy, Cong, Shroff, Saroufim, Miryoosefi, Hammad, Nasir, Xi, Gao, Maeng, Hora, Cheng, Haghani, Lewenberg, Lu, Matysiak, Raisinghani, Wang, Baugher, Sukthankar, Giang, Schultz, Fiedel, Chen, Lee, Dey, Zheng, Paul, Smith, Ly, Wang, Bansal, Perz, Ricco, Blank, Keshava, Sharma, Chow, Lad, Jalan, Osindero, Swanson, Scott, Ilić, Li, Jonnalagadda, Soudagar, Xiong, Batsaikhan, Jarrett, Kumar, Shah, Lawlor, Waters, Graham, May, Ramos, Lefdal, Cankara, Cano, O'Donoghue, Borovik, Liu, Grimstad, Alnahlawi, Tsihlas, Hudson, Grigorev, Jia, Huang, Igwe, Lebedev, Tang, Krivokon, Garcia, Tan, Jia, Stys, Vashishth, Liang, Venkatraman, Gu, Kementsietsidis, Zhu, Jung, Bai, Hosseini, Ahmed, Gupta, Yuan, Ashraf, Nigam, Vasudevan, Awasthi, Gilady, Mariet, Eskander,
  Li, Hu, Garrido, Schlattner, Zhang, Saxena, Dević, Muralidharan, Murthy, Zhou, Choi, Wongpanich, Wang, Shah, Xu, Huang, Spencer, Chen, Cohan, Wang, Tompson, Wu, Haroun, Li, Huergo, Yang, Yin, Wendt, Bendersky, Chaabouni, Snaider, Ferret, Jindal, Thompson, Xue, Bishop, Phal, Sharma, Sung, Radhakrishnan, Shomrat, Ingle, Vij, Gilmer, Istin, Sobell, Lu, Nottage, Sadigh, Willcock, Zhang, Xu, Brown, Lee, Wang, Zhu, Tay, Kim, Gutierrez, Sharma, Xian, Seo, Cui, Pochernina, Baetu, Jastrzębski, Ly, Elhawaty, Suh, Sezener, Wang, Yuen, Tucker, Cai, Yang, Wang, Muzio, Qian, Yoo, Lockhart, McKee, Guo, Mehrotra, Mendonça, Mehta, Ben, Tekur, Mu, Zhu, Krakovna, Lee, Maschinot, Cevey, Choe, Bai, Srinivasan, Gasaway, Young, Siegler, Holtmann-Rice, Piratla, Baumli, Yogev, Hofer, van Hasselt, Grant, Chervonyi, Silver, Hogue, Agarwal, Wang, Singh, Flynn, Lipschultz, David, Bellot, Yang, Le, Graziano, Olszewska, Hui, Maurya, Parotsidis, Chen, Oguntebi, Kelley, Baddepudi, Mauerer, Shaw, Siegman, Yang, Shetty, Roy, Song,
  Stokowiec, Burnell, Savant, Busa-Fekete, Miao, Ghosh, MacDermed, Lippe, Dektiarev, Behrman, Mentzer, Nguyen, Wei, Verma, Knutsen, Dasari, Yan, Mitrichev, Wang, Shejwalkar, Austin, Sunkara, Potti, Virin, Wright, Liu, Riva, Pot, Kochanski, Le, Balasubramaniam, Dhar, Liao, Bloniarz, Shukla, Cole, Lee, Zhang, Kafle, Vashishtha, Mahmoudieh, Chen, Hoffmann, Srinivasan, Lago, Shalom, Wang, Elabd, Sharma, Oh, Kothawade, Le, Monteiro, Yang, Alarakyia, Geirhos, Mincu, Garnes, Kobayashi, Mariooryad, Krasowiak, Lai, Mourad, Wang, Bu, Aharoni, Chen, Goyal, Zubov, Bapna, Dabir, Kothari, Lamerigts, Cao, Shar, Yew, Kulkarni, Mahaarachchi, Joshi, Zhu, Lichtarge, Zhou, Muckenhirn, Selo, Vinyals, Chen, Brohan, Mehta, Cogan, Wang, Geri, Ko, Chen, Viola, Shivam, Wang, Elish, Popa, Pereira, Liu, Koster, Kim, Zhang, Ebrahimi, Talukdar, Zheng, Poklukar, Mikhalap, Johnson, Vijayakumar, Omernick, Dibb, Dubey, Hu, Suman, Aggarwal, Kornakov, Xia, Lowe, Kolganov, Xiao, Nikolaev, Hemingray, Li, Iljazi, Rybiński, Sandhu, Lu, Luong,
  Jenatton, Govindaraj, Li, Dulac-Arnold, Park, Wang, Modi, Pouget-Abadie, Greller, Gupta, Berry, Ramachandran, Xie, McCafferty, Wang, Gupta, Lim, Bratanič, Brock, Akolzin, Sproch, Karliner, Kim, Goedeckemeyer, Shazeer, Schmid, Calandriello, Bhatia, Choromanski, Montgomery, Dua, Ramalho, King, Gao, Nguyen, Lindner, Pitta, Johnson, Salama, Ardila, Han, Farnese, Odoom, Wang, Ding, Rink, Smith, Lehri, Cohen, Vats, He, Gopavarapu, Paszke, Patel, Gansbeke, Loher, Castro, Voitovich, von Glehn, George, Niklaus, Eaton-Rosen, Rakićević, Jue, Perel, Zhang, Bahat, Pouget, Xing, Huot, Shenoy, Bos, Coriou, Richter, Noy, Wang, Ontanon, Qin, Makarchuk, Hassabis, Li, Sharma, Venkatesan, Kemaev, Daniel, Huang, Shah, Ponce, Chen, Faruqui, Wu, Andačić, Payrits, McDuff, Hume, Cao, Tessler, Wang, Wang, Rendulic, Agustsson, Johnson, Lando, Howard, Padmanabhan, Daswani, Banino, Kilgore, Heek, Ji, Caceres, Li, Kassner, Vlaskin, Liu, Grills, Hou, Sukkerd, Cheon, Shetty, Markeeva, Stanczyk, Iyer, Gong, Gao, Gopalakrishnan, Blyth,
  Reynolds, Bhoopchand, Bilenko, Gharibian, Zayats, Faust, Singh, Ma, Jiao, Vijayanarasimhan, Aroyo, Yadav, Chakera, Kakarla, Meshram, Gregor, Botea, Senter, Jia, Kovacs, Sharma, Baur, Kang, He, Zhuo, Kostelac, Laish, Peng, O'Bryan, Kasenberg, Rao, Leurent, Zhang, Stevens, Salazar, Zhang, Lobov, Walker, Porter, Redshaw, Ke, Rao, Lee, Lam, Moffitt, Kim, Qiao, Koo, Dadashi, Song, Sundararajan, Xu, Kawamoto, Zhong, Barbu, Reddy, Verzetti, Li, Papamakarios, Klimczak-Plucińska, Cassin, Kavukcuoglu, Swavely, Vaucher, Zhao, Hemsley, Tschannen, Ge, Menghani, Yu, Ha, He, Wu, Song, Sterneck, Zinke, Calian, Marsden, Ruiz, Hessel, Gueta, Lee, Farris, Gupta, Li, Saleh, Misra, Xiao, Mendolicchio, Buttimore, Krayvanova, Nayakanti, Wiethoff, Pande, Mirhoseini, Lao, Liu, Hua, Chen, Malkov, Kalashnikov, Gupta, Audhkhasi, Zhai, Kopalle, Jain, Ofek, Meyer, Baatarsukh, Strejček, Qian, Freedman, Figueira, Sokolik, Bachem, Lin, Kharrat, Hidey, Xu, Duan, Li, Ersoy, Everett, Cen, Santamaria-Fernandez, Taubenfeld, Mackinnon, Deng,
  Zablotskaia, Viswanadha, Goel, Yates, Deng, Choy, Chen, Sinha, Mossin, Wang, Szlam, Hao, Rubenstein, Toksoz-Exley, Aperghis, Zhong, Ahn, Isard, Lacombe, Luisier, Anastasiou, Kalley, Prabhu, Dunleavy, Bijwadia, Mao-Jones, Chen, Pasumarthi, Wood, Dostmohamed, Hurley, Simsa, Parrish, Pajarskas, Harvey, Skopek, Kochinski, Rey, Rieser, Zhou, Lee, Acharya, Li, Jiang, Zhang, Gipson, Mahintorabi, Gelmi, Khajehnouri, Yeh, Lee, Matthey, Baker, Pham, Fu, Pak, Gupta, Vasconcelos, Sadovsky, Walker, Hsiao, Zochbauer, Marzoca, Velan, Zeng, Baechler, Driess, Jain, Huang, Tao, Maggs, Levine, Schneider, Gemzer, Petit, Han, Fisher, Zelle, Biles, Ie, Fadeeva, Liu, Franco, Collister, Zhang, Wang, Zhao, Kieliger, Shuster, Zhu, Gong, Chan, Sun, Basu, Zimmermann, Hayes, Bapna, Snoek, Yang, Datta, Abdallah, Kilgour, Li, Mah, Jun, Rivière, Karmarkar, Spalink, Huang, Gonzalez, Tran, Nowak, Palowitch, Chadwick, Talius, Mehta, Sellam, Fränken, Nicosia, He, Kini, Amos, Basu, Jobe, Shaw, Xu, Evans, Ikeda, Yan, Jin, Wang, Yadav,
  Labzovsky, Sampath, Ma, Schumann, Siddhant, Shah, Youssef, Agarwal, Dabney, Tonioni, Ambar, Li, Guyon, Li, Soergel, Fang, Karadzhov, Udrescu, Trinh, Raunak, Noury, Guo, Gupta, Finkelstein, Petek, Liang, Billock, Sun, Wood, Song, Yu, Matejovicova, Cohen, Andra, D'Ambrosio, Deng, Nallatamby, Songhori, Dangovski, Lampinen, Botadra, Hillier, Cao, Baddi, Kuncoro, Yoshino, Bhagatwala, Ranzato, Schaeffer, Liu, Ye, Sarvana, Nham, Kuang, Gao, Baek, Mittal, Wahid, Gergely, Ni, Feldman, Muir, Lamblin, Macherey, Dyer, Kilpatrick, Campos, Bhutani, Fort, Ahmad, Severyn, Chatziprimou, Ferludin, Dimarco, Kusupati, Heyward, Bahir, Villela, Millican, Marcus, Bahargam, Unlu, Roth, Wei, Gopal, Ghoshal, Lee, Lin, Lees, Lee, Hosseini, Fan, Neel, Wu, Altun, Cai, Piqueras, Woodward, Bissacco, Haykal, Bordbar, Sundaram, Hodkinson, Toyama, Polovets, Myers, Sinha, Levinboim, Krishnakumar, Chhaparia, Sholokhova, Gundavarapu, Jawahar, Qureshi, Hu, Momchev, Rahtz, Wu, S, Dhamdhere, Guo, Gupta, Eslami, Schain, Blokzijl, Welling, Orr,
  Bolelli, Perez-Nieves, Sirotenko, Prasad, Kar, Pigem, Terzi, Weisz, Ghosh, Mavalankar, Madeka, Daugaard, Adam, Shah, Berman, Tran, Baker, Andrejczuk, Chole, Raboshchuk, Mirzazadeh, Kagohara, Wu, Schallhart, Orlando, Wang, Rrustemi, Xiong, Liu, Vezer, Ramsden, yiin Chang, Mudgal, Li, Vieillard, Hoshen, Ahmad, Slone, Hua, Potikha, Rossini, Stritar, Prakash, Wang, Dong, Nazari, Nehoran, Tekelioglu, Li, Badola, Funkhouser, Li, Yerram, Ganeshan, Formoso, Langner, Shi, Li, Yamamori, Panda, Saade, Scarpati, Breaux, Carey, Zhou, Hsieh, Bridgers, Butryna, Gupta, Tulsyan, Woo, Eltyshev, Grathwohl, Parks, Benjamin, Panigrahy, Dodhia, Freitas, Sauer, Song, Alet, Tolins, Paduraru, Zhou, Albert, Zhang, Shu, Bansal, Nguyen, Globerson, Xiao, Manyika, Hennigan, Rong, Matak, Bakalov, Sharma, Sinopalnikov, Pierson, Roller, Brown, Gao, Fukuzawa, Ghafouri, Vassigh, Barr, Wang, Korsun, Jayaram, Ren, Zaman, Khan, Lunts, Deutsch, Uthus, Katz, Samsikova, Khalifa, Sethi, Sun, Tang, Alon, Luo, Yu, Nayyar, Petrini, Truong,
  Hellendoorn, Chinaev, Alberti, Wang, Hu, Mirrokni, Balashankar, Aharon, Mehta, Iscen, Kready, Manning, Mohananey, Chen, Tripathi, Wu, Petrovski, Hwang, Baeuml, Chandrakaladharan, Liu, Coaguila, Chen, Ma, Tafti, Tatineni, Spitz, Ye, Vicol, Rosca, Puigdomènech, Yahav, Ghemawat, Lin, Kirk, Nabulsi, Brin, Bohnet, Caluwaerts, Veerubhotla, Zheng, Dai, Petrov, Xu, Mehran, Xu, Zintgraf, Choi, Hombaiah, Thoppilan, Reddi, Lew, Li, Webster, Sawhney, Lamprou, Shakeri, Lunayach, Chen, Bagri, Salcianu, Chen, Donchev, Magister, Nørly, Rodrigues, Izo, Noga, Zou, Köppe, Zhou, Lee, Long, Eisenbud, Chen, Schenck, To, Zhong, Taropa, Truong, Levy, Martins, Zhang, Semturs, Zhang, Yakubovich, Moreno, McConnaughey, Lu, Redmond, Weerts, Bitton, Refice, Lacasse, Conmy, Tallec, Odell, Forbes-Pollard, Socala, Hoech, Kohli, Walton, Wang, Sazanovich, Zhu, Kapishnikov, Galt, Denton, Murdoch, Sikora, Mohamed, Wei, First, McConnell, Cobo, Qin, Avrahami, Balle, Watanabe, Louis, Kraft, Ariafar, Gu, Rives, Yoon, Rusu, Cobon-Kerr, Hahn,
  Luo, Zhu, Ahuja, Benenson, Kaufman, Yu, Hightower, Zhang, Ni, Hendricks, Wang, Yona, Jain, Barrio, Bhupatiraju, Velusamy, Dafoe, Riedel, Thomas, Yuan, Bellaiche, Panthaplackel, Kloboves, Jauhari, Akbulut, Davchev, Gladchenko, Madras, Chuklin, Hill, Yuan, Madhavan, Leonhard, Scandinaro, Chen, Niu, Douillard, Damoc, Onoe, Pedregosa, Bertsch, Leichner, Pagadora, Malmaud, Ponda, Twigg, Duzhyi, Shen, Wang, Garg, Chen, Evci, Lee, Liu, Kojima, Yamaguchi, Rajendran, Piergiovanni, Rajendran, Fornoni, Ibagon, Ragan, Khan, Blitzer, Bunner, Sun, Kosakai, Lundberg, Elue, Guu, Park, Park, Narayanaswamy, Wu, Mudigonda, Cohn, Mu, Kumar, Graesser, Zhang, Killam, Zhuang, Giménez, Jishi, Ley-Wild, Zhai, Osawa, Cedillo, Liu, Upadhyay, Sieniek, Sharma, Paine, Angelova, Addepalli, Parada, Majumder, Lamp, Kumar, Deng, Myaskovsky, Sabolić, Dudek, York, de~Chaumont~Quitry, Nie, Cattle, Gunjan, Piot, Khawaja, Bang, Wang, Khodadadeh, R, Rawlani, Powell, Lee, Griesser, Oh, Magalhaes, Li, Tokumine, Vogel, Hsu, BC, Jindal, Cohen,
  Yang, Yuan, de~Cesare, Bruguier, Xu, Roy, Jacovi, Belov, Arya, Meadowlark, Cohen-Ganor, Ye, Morris-Suzuki, Banzal, Song, Ponnuramu, Zhang, Scrivener, Zaiem, Rochman, Han, Ghazi, Lee, Drath, Suo, Girgis, Shenoy, Nguyen, Eck, Gupta, Yan, Carreira, Gulati, Sang, Mirylenka, Cooney, Chou, Ling, Fan, Coleman, Tubone, Kumar, Baldridge, Hernandez-Campos, Lazaridou, Besley, Yona, Bulut, Wellens, Pierigiovanni, George, Green, Han, Tao, Clark, You, Abdolmaleki, Fu, Chen, Chaugule, Chandorkar, Rahman, Thompson, Koanantakool, Bernico, Ren, Vlasov, Vassilvitskii, Kula, Liang, Kim, Huang, Ye, Lepikhin, and Helmholz}]{comanici2025gemini}
Gheorghe Comanici, Eric Bieber, Mike Schaekermann, Ice Pasupat, Noveen Sachdeva, Inderjit Dhillon, Marcel Blistein, Ori Ram, Dan Zhang, Evan Rosen, Luke Marris, Sam Petulla, Colin Gaffney, Asaf Aharoni, Nathan Lintz, Tiago~Cardal Pais, Henrik Jacobsson, Idan Szpektor, Nan-Jiang Jiang, and 3405 others. 2025.
\newblock \href {https://doi.org/10.48550/arXiv.2507.06261} {{Gemini 2.5}: Pushing the frontier with advanced reasoning, multimodality, long context, and next generation agentic capabilities}.
\newblock \emph{arXiv preprint arXiv:2507.06261}.

\bibitem[{Emelin et~al.(2021)Emelin, Le~Bras, Hwang, Forbes, and Choi}]{emelin2021moral}
Denis Emelin, Ronan Le~Bras, Jena~D. Hwang, Maxwell Forbes, and Yejin Choi. 2021.
\newblock \href {https://doi.org/10.18653/v1/2021.emnlp-main.54} {{Moral Stories}: Situated reasoning about norms, intents, actions, and their consequences}.
\newblock In \emph{Proceedings of the 2021 Conference on Empirical Methods in Natural Language Processing}, pages 698--718, Online and Punta Cana, Dominican Republic. Association for Computational Linguistics.

\bibitem[{Garg and Fetzer(2025)}]{garg2025causal}
Prashant Garg and Thiemo Fetzer. 2025.
\newblock \href {https://doi.org/10.48550/arXiv.2501.06873} {Causal claims in economics}.
\newblock \emph{Preprint}, arXiv:2501.06873.

\bibitem[{Grattafiori et~al.(2024)Grattafiori, Dubey, Jauhri, Pandey, Kadian, Al-Dahle, Letman, Mathur, Schelten, Vaughan, Yang, Fan, Goyal, Hartshorn, Yang, Mitra, Sravankumar, Korenev, Hinsvark, Rao, Zhang, Rodriguez, Gregerson, Spataru, Roziere, Biron, Tang, Chern, Caucheteux, Nayak, Bi, Marra, McConnell, Keller, Touret, Wu, Wong, Ferrer, Nikolaidis, Allonsius, Song, Pintz, Livshits, Wyatt, Esiobu, Choudhary, Mahajan, Garcia-Olano, Perino, Hupkes, Lakomkin, AlBadawy, Lobanova, Dinan, Smith, Radenovic, Guzmán, Zhang, Synnaeve, Lee, Anderson, Thattai, Nail, Mialon, Pang, Cucurell, Nguyen, Korevaar, Xu, Touvron, Zarov, Ibarra, Kloumann, Misra, Evtimov, Zhang, Copet, Lee, Geffert, Vranes, Park, Mahadeokar, Shah, van~der Linde, Billock, Hong, Lee, Fu, Chi, Huang, Liu, Wang, Yu, Bitton, Spisak, Park, Rocca, Johnstun, Saxe, Jia, Alwala, Prasad, Upasani, Plawiak, Li, Heafield, Stone, El-Arini, Iyer, Malik, Chiu, Bhalla, Lakhotia, Rantala-Yeary, van~der Maaten, Chen, Tan, Jenkins, Martin, Madaan, Malo, Blecher,
  Landzaat, de~Oliveira, Muzzi, Pasupuleti, Singh, Paluri, Kardas, Tsimpoukelli, Oldham, Rita, Pavlova, Kambadur, Lewis, Si, Singh, Hassan, Goyal, Torabi, Bashlykov, Bogoychev, Chatterji, Zhang, Duchenne, Çelebi, Alrassy, Zhang, Li, Vasic, Weng, Bhargava, Dubal, Krishnan, Koura, Xu, He, Dong, Srinivasan, Ganapathy, Calderer, Cabral, Stojnic, Raileanu, Maheswari, Girdhar, Patel, Sauvestre, Polidoro, Sumbaly, Taylor, Silva, Hou, Wang, Hosseini, Chennabasappa, Singh, Bell, Kim, Edunov, Nie, Narang, Raparthy, Shen, Wan, Bhosale, Zhang, Vandenhende, Batra, Whitman, Sootla, Collot, Gururangan, Borodinsky, Herman, Fowler, Sheasha, Georgiou, Scialom, Speckbacher, Mihaylov, Xiao, Karn, Goswami, Gupta, Ramanathan, Kerkez, Gonguet, Do, Vogeti, Albiero, Petrovic, Chu, Xiong, Fu, Meers, Martinet, Wang, Wang, Tan, Xia, Xie, Jia, Wang, Goldschlag, Gaur, Babaei, Wen, Song, Zhang, Li, Mao, Coudert, Yan, Chen, Papakipos, Singh, Srivastava, Jain, Kelsey, Shajnfeld, Gangidi, Victoria, Goldstand, Menon, Sharma, Boesenberg,
  Baevski, Feinstein, Kallet, Sangani, Teo, Yunus, Lupu, Alvarado, Caples, Gu, Ho, Poulton, Ryan, Ramchandani, Dong, Franco, Goyal, Saraf, Chowdhury, Gabriel, Bharambe, Eisenman, Yazdan, James, Maurer, Leonhardi, Huang, Loyd, Paola, Paranjape, Liu, Wu, Ni, Hancock, Wasti, Spence, Stojkovic, Gamido, Montalvo, Parker, Burton, Mejia, Liu, Wang, Kim, Zhou, Hu, Chu, Cai, Tindal, Feichtenhofer, Gao, Civin, Beaty, Kreymer, Li, Adkins, Xu, Testuggine, David, Parikh, Liskovich, Foss, Wang, Le, Holland, Dowling, Jamil, Montgomery, Presani, Hahn, Wood, Le, Brinkman, Arcaute, Dunbar, Smothers, Sun, Kreuk, Tian, Kokkinos, Ozgenel, Caggioni, Kanayet, Seide, Florez, Schwarz, Badeer, Swee, Halpern, Herman, Sizov, Zhang, Lakshminarayanan, Inan, Shojanazeri, Zou, Wang, Zha, Habeeb, Rudolph, Suk, Aspegren, Goldman, Zhan, Damlaj, Molybog, Tufanov, Leontiadis, Veliche, Gat, Weissman, Geboski, Kohli, Lam, Asher, Gaya, Marcus, Tang, Chan, Zhen, Reizenstein, Teboul, Zhong, Jin, Yang, Cummings, Carvill, Shepard, McPhie, Torres,
  Ginsburg, Wang, Wu, U, Saxena, Khandelwal, Zand, Matosich, Veeraraghavan, Michelena, Li, Jagadeesh, Huang, Chawla, Huang, Chen, Garg, A, Silva, Bell, Zhang, Guo, Yu, Moshkovich, Wehrstedt, Khabsa, Avalani, Bhatt, Mankus, Hasson, Lennie, Reso, Groshev, Naumov, Lathi, Keneally, Liu, Seltzer, Valko, Restrepo, Patel, Vyatskov, Samvelyan, Clark, Macey, Wang, Hermoso, Metanat, Rastegari, Bansal, Santhanam, Parks, White, Bawa, Singhal, Egebo, Usunier, Mehta, Laptev, Dong, Cheng, Chernoguz, Hart, Salpekar, Kalinli, Kent, Parekh, Saab, Balaji, Rittner, Bontrager, Roux, Dollar, Zvyagina, Ratanchandani, Yuvraj, Liang, Alao, Rodriguez, Ayub, Murthy, Nayani, Mitra, Parthasarathy, Li, Hogan, Battey, Wang, Howes, Rinott, Mehta, Siby, Bondu, Datta, Chugh, Hunt, Dhillon, Sidorov, Pan, Mahajan, Verma, Yamamoto, Ramaswamy, Lindsay, Lindsay, Feng, Lin, Zha, Patil, Shankar, Zhang, Zhang, Wang, Agarwal, Sajuyigbe, Chintala, Max, Chen, Kehoe, Satterfield, Govindaprasad, Gupta, Deng, Cho, Virk, Subramanian, Choudhury, Goldman,
  Remez, Glaser, Best, Koehler, Robinson, Li, Zhang, Matthews, Chou, Shaked, Vontimitta, Ajayi, Montanez, Mohan, Kumar, Mangla, Ionescu, Poenaru, Mihailescu, Ivanov, Li, Wang, Jiang, Bouaziz, Constable, Tang, Wu, Wang, Wu, Gao, Kleinman, Chen, Hu, Jia, Qi, Li, Zhang, Zhang, Adi, Nam, Wang, Zhao, Hao, Qian, Li, He, Rait, DeVito, Rosnbrick, Wen, Yang, Zhao, and Ma}]{grattafiori2024llama}
Aaron Grattafiori, Abhimanyu Dubey, Abhinav Jauhri, Abhinav Pandey, Abhishek Kadian, Ahmad Al-Dahle, Aiesha Letman, Akhil Mathur, Alan Schelten, Alex Vaughan, Amy Yang, Angela Fan, Anirudh Goyal, Anthony Hartshorn, Aobo Yang, Archi Mitra, Archie Sravankumar, Artem Korenev, Arthur Hinsvark, and 540 others. 2024.
\newblock \href {https://doi.org/10.48550/arXiv.2407.21783} {The {Llama 3} herd of models}.
\newblock \emph{arXiv preprint arXiv:2407.21783}.

\bibitem[{Guo et~al.(2017)Guo, Pleiss, Sun, and Weinberger}]{guo2017calibration}
Chuan Guo, Geoff Pleiss, Yu~Sun, and Kilian~Q. Weinberger. 2017.
\newblock \href {https://proceedings.mlr.press/v70/guo17a.html} {On calibration of modern neural networks}.
\newblock In \emph{Proceedings of the 34th International Conference on Machine Learning}, volume~70 of \emph{Proceedings of Machine Learning Research}, pages 1321--1330. PMLR.

\bibitem[{Guo and Yang(2024)}]{guo2024econnli}
Yue Guo and Yi~Yang. 2024.
\newblock \href {https://doi.org/10.18653/v1/2024.findings-acl.58} {{EconNLI}: Evaluating large language models on economics reasoning}.
\newblock In \emph{Findings of the Association for Computational Linguistics: ACL 2024}, pages 982--994, Bangkok, Thailand. Association for Computational Linguistics.

\bibitem[{Huang et~al.(2024)Huang, Hong, Zhang, Shao, Yang, Yu, Zhang, Liang, and Song}]{huang2024clomo}
Yinya Huang, Ruixin Hong, Hongming Zhang, Wei Shao, Zhicheng Yang, Dong Yu, Changshui Zhang, Xiaodan Liang, and Linqi Song. 2024.
\newblock \href {https://doi.org/10.18653/v1/2024.acl-long.593} {{CLOMO}: Counterfactual logical modification with large language models}.
\newblock In \emph{Proceedings of the 62nd Annual Meeting of the Association for Computational Linguistics (Volume 1: Long Papers)}, pages 11012--11034, Bangkok, Thailand. Association for Computational Linguistics.

\bibitem[{Jin et~al.(2023)Jin, Chen, Leeb, Gresele, Kamal, Lyu, Blin, Gonzalez~Adauto, Kleiman-Weiner, Sachan, and Sch\"{o}lkopf}]{jin2023cladder}
Zhijing Jin, Yuen Chen, Felix Leeb, Luigi Gresele, Ojasv Kamal, Zhiheng Lyu, Kevin Blin, Fernando Gonzalez~Adauto, Max Kleiman-Weiner, Mrinmaya Sachan, and Bernhard Sch\"{o}lkopf. 2023.
\newblock \href {https://doi.org/10.52202/075280-1353} {{CLadder}: Assessing causal reasoning in language models}.
\newblock In \emph{Advances in Neural Information Processing Systems}, volume~36, pages 31038--31065. Curran Associates, Inc.

\bibitem[{Joren et~al.(2025)Joren, Zhang, Ferng, Juan, Taly, and Rashtchian}]{joren2025sufficient}
Hailey Joren, Jianyi Zhang, Chun-Sung Ferng, Da-Cheng Juan, Ankur Taly, and Cyrus Rashtchian. 2025.
\newblock \href {https://proceedings.iclr.cc/paper_files/paper/2025/hash/33dffa2e3d2ab74a783d1a8c292f66d9-Abstract-Conference.html} {Sufficient context: A new lens on retrieval augmented generation systems}.
\newblock In \emph{International Conference on Learning Representations}, pages 20310--20334.

\bibitem[{Kojima et~al.(2022)Kojima, Gu, Reid, Matsuo, and Iwasawa}]{kojima2022large}
Takeshi Kojima, Shixiang~(Shane) Gu, Machel Reid, Yutaka Matsuo, and Yusuke Iwasawa. 2022.
\newblock \href {https://doi.org/10.52202/068431-1613} {Large language models are zero-shot reasoners}.
\newblock In \emph{Advances in Neural Information Processing Systems}, volume~35, pages 22199--22213. Curran Associates, Inc.

\bibitem[{Lam et~al.(2025)Lam, Wang, Huang, and Lyu}]{lam2025codecrash}
Man~Ho Lam, Chaozheng Wang, Jen-Tse Huang, and Michael~R. Lyu. 2025.
\newblock \href {https://doi.org/10.52202/085713-4029} {{CodeCrash}: Exposing {LLM} fragility to misleading natural language in code reasoning}.
\newblock In \emph{Advances in Neural Information Processing Systems}, volume~38, pages 133886--133932. Neural Information Processing Systems Foundation, Inc.

\bibitem[{Min et~al.(2020)Min, Michael, Hajishirzi, and Zettlemoyer}]{min2020ambigqa}
Sewon Min, Julian Michael, Hannaneh Hajishirzi, and Luke Zettlemoyer. 2020.
\newblock \href {https://doi.org/10.18653/v1/2020.emnlp-main.466} {{AmbigQA}: Answering ambiguous open-domain questions}.
\newblock In \emph{Proceedings of the 2020 Conference on Empirical Methods in Natural Language Processing (EMNLP)}, pages 5783--5797, Online. Association for Computational Linguistics.

\bibitem[{{National Bureau of Economic Research}(2026)}]{nber_wp_series}
{National Bureau of Economic Research}. 2026.
\newblock \href {https://www.nber.org/papers} {Working papers}.
\newblock Accessed: 2026-10-12.

\bibitem[{{OpenAI} et~al.(2023){OpenAI}, Achiam, Adler, Agarwal, Ahmad, Akkaya, Aleman, Almeida, Altenschmidt, Altman, Anadkat, Avila, Babuschkin, Balaji, Balcom, Baltescu, Bao, Bavarian, Belgum, Bello, Berdine, Bernadett-Shapiro, Berner, Bogdonoff, Boiko, Boyd, Brakman, Brockman, Brooks, Brundage, Button, Cai, Campbell, Cann, Carey, Carlson, Carmichael, Chan, Chang, Chantzis, Chen, Chen, Chen, Chen, Chen, Chess, Cho, Chu, Chung, Cummings, Currier, Dai, Decareaux, Degry, Deutsch, Deville, Dhar, Dohan, Dowling, Dunning, Ecoffet, Eleti, Eloundou, Farhi, Fedus, Felix, Fishman, Forte, Fulford, Gao, Georges, Gibson, Goel, Gogineni, Goh, Gontijo-Lopes, Gordon, Grafstein, Gray, Greene, Gross, Gu, Guo, Hallacy, Han, Harris, He, Heaton, Heidecke, Hesse, Hickey, Hickey, Hoeschele, Houghton, Hsu, Hu, Hu, Huizinga, Jain, Jain, Jang, Jiang, Jiang, Jin, Jin, Jomoto, Jonn, Jun, Kaftan, Łukasz Kaiser, Kamali, Kanitscheider, Keskar, Khan, Kilpatrick, Kim, Kim, Kim, Kirchner, Kiros, Knight, Kokotajlo, Łukasz Kondraciuk,
  Kondrich, Konstantinidis, Kosic, Krueger, Kuo, Lampe, Lan, Lee, Leike, Leung, Levy, Li, Lim, Lin, Lin, Litwin, Lopez, Lowe, Lue, Makanju, Malfacini, Manning, Markov, Markovski, Martin, Mayer, Mayne, McGrew, McKinney, McLeavey, McMillan, McNeil, Medina, Mehta, Menick, Metz, Mishchenko, Mishkin, Monaco, Morikawa, Mossing, Mu, Murati, Murk, Mély, Nair, Nakano, Nayak, Neelakantan, Ngo, Noh, Ouyang, O'Keefe, Pachocki, Paino, Palermo, Pantuliano, Parascandolo, Parish, Parparita, Passos, Pavlov, Peng, Perelman, de~Avila Belbute~Peres, Petrov, de~Oliveira~Pinto, Pokorny, Pokrass, Pong, Powell, Power, Power, Proehl, Puri, Radford, Rae, Ramesh, Raymond, Real, Rimbach, Ross, Rotsted, Roussez, Ryder, Saltarelli, Sanders, Santurkar, Sastry, Schmidt, Schnurr, Schulman, Selsam, Sheppard, Sherbakov, Shieh, Shoker, Shyam, Sidor, Sigler, Simens, Sitkin, Slama, Sohl, Sokolowsky, Song, Staudacher, Such, Summers, Sutskever, Tang, Tezak, Thompson, Tillet, Tootoonchian, Tseng, Tuggle, Turley, Tworek, Uribe, Vallone,
  Vijayvergiya, Voss, Wainwright, Wang, Wang, Wang, Ward, Wei, Weinmann, Welihinda, Welinder, Weng, Weng, Wiethoff, Willner, Winter, Wolrich, Wong, Workman, Wu, Wu, Wu, Xiao, Xu, Yoo, Yu, Yuan, Zaremba, Zellers, Zhang, Zhang, Zhao, Zheng, Zhuang, Zhuk, and Zoph}]{achiam2023gpt}
{OpenAI}, Josh Achiam, Steven Adler, Sandhini Agarwal, Lama Ahmad, Ilge Akkaya, Florencia~Leoni Aleman, Diogo Almeida, Janko Altenschmidt, Sam Altman, Shyamal Anadkat, Red Avila, Igor Babuschkin, Suchir Balaji, Valerie Balcom, Paul Baltescu, Haiming Bao, Mohammad Bavarian, Jeff Belgum, and 261 others. 2023.
\newblock \href {https://doi.org/10.48550/arXiv.2303.08774} {{GPT-4} technical report}.
\newblock \emph{arXiv preprint arXiv:2303.08774}.

\bibitem[{Pritchett and Sandefur(2015)}]{pritchett2015learning}
Lant Pritchett and Justin Sandefur. 2015.
\newblock \href {https://doi.org/10.1257/aer.p20151016} {Learning from experiments when context matters}.
\newblock \emph{American Economic Review}, 105(5):471--475.

\bibitem[{Quan and Liu(2024)}]{quan2024econlogicqa}
Yinzhu Quan and Zefang Liu. 2024.
\newblock \href {https://doi.org/10.18653/v1/2024.findings-emnlp.125} {{EconLogicQA}: A question-answering benchmark for evaluating large language models in economic sequential reasoning}.
\newblock In \emph{Findings of the Association for Computational Linguistics: EMNLP 2024}, pages 2273--2282, Miami, Florida, USA. Association for Computational Linguistics.

\bibitem[{Raman et~al.(2025)Raman, Lundy, Amin, Leyton-Brown, and Perla}]{raman2026steer}
Narun Raman, Taylor Lundy, Thiago Amin, Kevin Leyton-Brown, and Jesse Perla. 2025.
\newblock \href {https://doi.org/10.52202/085713-2826} {{STEER-ME}: Assessing the microeconomic reasoning of large language models}.
\newblock In \emph{Advances in Neural Information Processing Systems}, volume~38. Curran Associates, Inc.

\bibitem[{Raman et~al.(2024)Raman, Lundy, Amouyal, Levine, Leyton-Brown, and Tennenholtz}]{raman2024steer}
Narun~Krishnamurthi Raman, Taylor Lundy, Samuel~Joseph Amouyal, Yoav Levine, Kevin Leyton-Brown, and Moshe Tennenholtz. 2024.
\newblock \href {https://proceedings.mlr.press/v235/raman24b.html} {{STEER}: Assessing the economic rationality of large language models}.
\newblock In \emph{Proceedings of the 41st International Conference on Machine Learning}, volume 235 of \emph{Proceedings of Machine Learning Research}, pages 42026--42047. PMLR.

\bibitem[{Srikanth and Rudinger(2025)}]{srikanth-rudinger-2025-nli}
Neha Srikanth and Rachel Rudinger. 2025.
\newblock \href {https://doi.org/10.18653/v1/2025.naacl-long.130} {{NLI} under the microscope: What atomic hypothesis decomposition reveals}.
\newblock In \emph{Proceedings of the 2025 Conference of the Nations of the Americas Chapter of the Association for Computational Linguistics: Human Language Technologies (Volume 1: Long Papers)}, pages 2574--2589, Albuquerque, New Mexico. Association for Computational Linguistics.

\bibitem[{Srivastava et~al.(2023)Srivastava, Rastogi, Rao, Shoeb, Abid, Fisch, Brown, Santoro, Gupta, Garriga-Alonso, Kluska, Lewkowycz, Agarwal, Power, Ray, Warstadt, Kocurek, Safaya, Tazarv, Xiang, Parrish, Nie, Hussain, Askell, Dsouza, Slone, Rahane, Iyer, Andreassen, Madotto, Santilli, Stuhlm{\"u}ller, Dai, La, Lampinen, Zou, Jiang, Chen, Vuong, Gupta, Gottardi, Norelli, Venkatesh, Gholamidavoodi, Tabassum, Menezes, Kirubarajan, Mullokandov, Sabharwal, Herrick, Efrat, Erdem, Karaka{\c{s}}, Roberts, Loe, Zoph, Bojanowski, {\"O}zyurt, Hedayatnia, Neyshabur, Inden, Stein, Ekmekci, Lin, Howald, Orinion, Diao, Dour, Stinson, Argueta, Ferri, Singh, Rathkopf, Meng, Baral, Wu, Callison-Burch, Waites, Voigt, Manning, Potts, Ramirez, Rivera, Siro, Raffel, Ashcraft, Garbacea, Sileo, Garrette, Hendrycks, Kilman, Roth, Freeman, Khashabi, Levy, Gonz{\'a}lez, Perszyk, Hernandez, Chen, Ippolito, Gilboa, Dohan, Drakard, Jurgens, Datta, Ganguli, Emelin, Kleyko, Yuret, Chen, Tam, Hupkes, Misra, Buzan, Mollo, Yang, Lee,
  Schrader, Shutova, Cubuk, Segal, Hagerman, Barnes, Donoway, Pavlick, Rodol{\`a}, Lam, Chu, Tang, Erdem, Chang, Chi, Dyer, Jerzak, Kim, Manyasi, Zheltonozhskii, Xia, Siar, Mart{\'\i}nez-Plumed, Happ{\'e}, Chollet, Rong, Mishra, Winata, de~Melo, Kruszewski, Parascandolo, Mariani, Wang, Jaimovitch-Lopez, Betz, Gur-Ari, Galijasevic, Kim, Rashkin, Hajishirzi, Mehta, Bogar, Shevlin, Schuetze, Yakura, Zhang, Wong, Ng, Noble, Jumelet, Geissinger, Kernion, Hilton, Lee, Fisac, Simon, Koppel, Zheng, Zou, Kocon, Thompson, Wingfield, Kaplan, Radom, Sohl-Dickstein, Phang, Wei, Yosinski, Novikova, Bosscher, Marsh, Kim, Taal, Engel, Alabi, Xu, Song, Tang, Waweru, Burden, Miller, Balis, Batchelder, Berant, Frohberg, Rozen, Hernandez-Orallo, Boudeman, Guerr, Jones, Tenenbaum, Rule, Chua, Kanclerz, Livescu, Krauth, Gopalakrishnan, Ignatyeva, Markert, Dhole, Gimpel, Omondi, Mathewson, Chiafullo, Shkaruta, Shridhar, McDonell, Richardson, Reynolds, Gao, Zhang, Dugan, Qin, Contreras-Ochando, Morency, Moschella, Lam, Noble,
  Schmidt, He, Oliveros-Col{\'o}n, Metz, Senel, Bosma, Sap, Hoeve, Farooqi, Faruqui, Mazeika, Baturan, Marelli, Maru, Ramirez-Quintana, Tolkiehn, Giulianelli, Lewis, Potthast, Leavitt, Hagen, Schubert, Baitemirova, Arnaud, McElrath, Yee, Cohen, Gu, Ivanitskiy, Starritt, Strube, Sw{\k{e}}drowski, Bevilacqua, Yasunaga, Kale, Cain, Xu, Suzgun, Walker, Tiwari, Bansal, Aminnaseri, Geva, Gheini, T, Peng, Chi, Lee, Krakover, Cameron, Roberts, Doiron, Martinez, Nangia, Deckers, Muennighoff, Keskar, Iyer, Constant, Fiedel, Wen, Zhang, Agha, Elbaghdadi, Levy, Evans, Casares, Doshi, Fung, Liang, Vicol, Alipoormolabashi, Liao, Liang, Chang, Eckersley, Htut, Hwang, Mi{\l}kowski, Patil, Pezeshkpour, Oli, Mei, Lyu, Chen, Banjade, Rudolph, Gabriel, Habacker, Risco, Milli{\`e}re, Garg, Barnes, Saurous, Arakawa, Raymaekers, Frank, Sikand, Novak, Sitelew, Bras, Liu, Jacobs, Zhang, Salakhutdinov, Chi, Lee, Stovall, Teehan, Yang, Singh, Mohammad, Anand, Dillavou, Shleifer, Wiseman, Gruetter, Bowman, Schoenholz, Han, Kwatra, Rous,
  Ghazarian, Ghosh, Casey, Bischoff, Gehrmann, Schuster, Sadeghi, Hamdan, Zhou, Srivastava, Shi, Singh, Asaadi, Gu, Pachchigar, Toshniwal, Upadhyay, Debnath, Shakeri, Thormeyer, Melzi, Reddy, Makini, Lee, Torene, Hatwar, Dehaene, Divic, Ermon, Biderman, Lin, Prasad, Piantadosi, Shieber, Misherghi, Kiritchenko, Mishra, Linzen, Schuster, Li, Yu, Ali, Hashimoto, Wu, Desbordes, Rothschild, Phan, Wang, Nkinyili, Schick, Kornev, Tunduny, Gerstenberg, Chang, Neeraj, Khot, Shultz, Shaham, Misra, Demberg, Nyamai, Raunak, Ramasesh, vinay~uday prabhu, Padmakumar, Srikumar, Fedus, Saunders, Zhang, Vossen, Ren, Tong, Zhao, Wu, Shen, Yaghoobzadeh, Lakretz, Song, Bahri, Choi, Yang, Hao, Chen, Belinkov, Hou, Hou, Bai, Seid, Zhao, Wang, Wang, Wang, and Wu}]{srivastava2023beyond}
Aarohi Srivastava, Abhinav Rastogi, Abhishek Rao, Abu Awal~Md Shoeb, Abubakar Abid, Adam Fisch, Adam~R. Brown, Adam Santoro, Aditya Gupta, Adri{\`a} Garriga-Alonso, Agnieszka Kluska, Aitor Lewkowycz, Akshat Agarwal, Alethea Power, Alex Ray, Alex Warstadt, Alexander~W. Kocurek, Ali Safaya, Ali Tazarv, and 431 others. 2023.
\newblock \href {https://openreview.net/forum?id=uyTL5Bvosj} {Beyond the imitation game: Quantifying and extrapolating the capabilities of language models}.
\newblock \emph{Transactions on Machine Learning Research}.

\bibitem[{Tian et~al.(2023)Tian, Mitchell, Zhou, Sharma, Rafailov, Yao, Finn, and Manning}]{tian-etal-2023-just}
Katherine Tian, Eric Mitchell, Allan Zhou, Archit Sharma, Rafael Rafailov, Huaxiu Yao, Chelsea Finn, and Christopher Manning. 2023.
\newblock \href {https://doi.org/10.18653/v1/2023.emnlp-main.330} {Just ask for calibration: Strategies for eliciting calibrated confidence scores from language models fine-tuned with human feedback}.
\newblock In \emph{Proceedings of the 2023 Conference on Empirical Methods in Natural Language Processing}, pages 5433--5442, Singapore. Association for Computational Linguistics.

\bibitem[{Vivalt(2020)}]{vivalt2020how}
Eva Vivalt. 2020.
\newblock \href {https://doi.org/10.1093/jeea/jvaa019} {How much can we generalize from impact evaluations?}
\newblock \emph{Journal of the European Economic Association}, 18(6):3045--3089.

\bibitem[{Wang et~al.(2024)Wang, Ma, Zhang, Ni, Chandra, Guo, Ren, Arulraj, He, Jiang, Li, Ku, Wang, Zhuang, Fan, Yue, and Chen}]{wang2024mmlu}
Yubo Wang, Xueguang Ma, Ge~Zhang, Yuansheng Ni, Abhranil Chandra, Shiguang Guo, Weiming Ren, Aaran Arulraj, Xuan He, Ziyan Jiang, Tianle Li, Max Ku, Kai Wang, Alex Zhuang, Rongqi Fan, Xiang Yue, and Wenhu Chen. 2024.
\newblock \href {https://doi.org/10.52202/079017-3018} {{MMLU-Pro}: A more robust and challenging multi-task language understanding benchmark}.
\newblock In \emph{Advances in Neural Information Processing Systems}, volume~37, pages 95266--95290. Curran Associates, Inc.

\bibitem[{Wang(2024)}]{wang2024causalbench}
Zeyu Wang. 2024.
\newblock \href {https://aclanthology.org/2024.sighan-1.17/} {{CausalBench}: A comprehensive benchmark for evaluating causal reasoning capabilities of large language models}.
\newblock In \emph{Proceedings of the 10th SIGHAN Workshop on Chinese Language Processing (SIGHAN-10)}, pages 143--151, Bangkok, Thailand. Association for Computational Linguistics.

\bibitem[{Wei et~al.(2022)Wei, Wang, Schuurmans, Bosma, Ichter, Xia, Chi, Le, and Zhou}]{wei2022chain}
Jason Wei, Xuezhi Wang, Dale Schuurmans, Maarten Bosma, Brian Ichter, Fei Xia, Ed~Chi, Quoc~V Le, and Denny Zhou. 2022.
\newblock \href {https://doi.org/10.52202/068431-1800} {Chain-of-thought prompting elicits reasoning in large language models}.
\newblock In \emph{Advances in Neural Information Processing Systems}, volume~35, pages 24824--24837. Curran Associates, Inc.

\bibitem[{Wei et~al.(2023)Wei, Wei, Tay, Tran, Webson, Lu, Chen, Liu, Huang, Zhou, and Ma}]{wei2023larger}
Jerry Wei, Jason Wei, Yi~Tay, Dustin Tran, Albert Webson, Yifeng Lu, Xinyun Chen, Hanxiao Liu, Da~Huang, Denny Zhou, and Tengyu Ma. 2023.
\newblock \href {https://doi.org/10.48550/arXiv.2303.03846} {Larger language models do in-context learning differently}.
\newblock \emph{Computing Research Repository}, arXiv:2303.03846.
\newblock Version 2.

\bibitem[{Wilie et~al.(2024)Wilie, Cahyawijaya, Ishii, He, and Fung}]{wilie-etal-2024-belief}
Bryan Wilie, Samuel Cahyawijaya, Etsuko Ishii, Junxian He, and Pascale Fung. 2024.
\newblock \href {https://doi.org/10.18653/v1/2024.emnlp-main.586} {Belief revision: The adaptability of large language models reasoning}.
\newblock In \emph{Proceedings of the 2024 Conference on Empirical Methods in Natural Language Processing}, pages 10480--10496, Miami, Florida, USA. Association for Computational Linguistics.

\bibitem[{Xiong et~al.(2024)Xiong, Hu, Lu, Li, Fu, He, and Hooi}]{xiong2024uncertainty}
Miao Xiong, Zhiyuan Hu, Xinyang Lu, Yifei Li, Jie Fu, Junxian He, and Bryan Hooi. 2024.
\newblock \href {https://proceedings.iclr.cc/paper_files/paper/2024/hash/6733cf15e10e2cd1d59af033c3bb8507-Abstract-Conference.html} {Can {LLMs} express their uncertainty? {An} empirical evaluation of confidence elicitation in {LLMs}}.
\newblock In \emph{International Conference on Learning Representations}, pages 23650--23678.

\bibitem[{Yang et~al.(2025)Yang, Li, Yang, Zhang, Hui, Zheng, Yu, Gao, Huang, Lv, Zheng, Liu, Zhou, Huang, Hu, Ge, Wei, Lin, Tang, Yang, Tu, Zhang, Yang, Yang, Zhou, Zhou, Lin, Dang, Bao, Yang, Yu, Deng, Li, Xue, Li, Zhang, Wang, Zhu, Men, Gao, Liu, Luo, Li, Tang, Yin, Ren, Wang, Zhang, Ren, Fan, Su, Zhang, Zhang, Wan, Liu, Wang, Cui, Zhang, Zhou, and Qiu}]{yang2025qwen3}
An~Yang, Anfeng Li, Baosong Yang, Beichen Zhang, Binyuan Hui, Bo~Zheng, Bowen Yu, Chang Gao, Chengen Huang, Chenxu Lv, Chujie Zheng, Dayiheng Liu, Fan Zhou, Fei Huang, Feng Hu, Hao Ge, Haoran Wei, Huan Lin, Jialong Tang, and 41 others. 2025.
\newblock \href {https://doi.org/10.48550/arXiv.2505.09388} {{Qwen3} technical report}.
\newblock \emph{arXiv preprint arXiv:2505.09388}.

\bibitem[{Yin et~al.(2023)Yin, Sun, Guo, Wu, Qiu, and Huang}]{yin-etal-2023-large}
Zhangyue Yin, Qiushi Sun, Qipeng Guo, Jiawen Wu, Xipeng Qiu, and Xuanjing Huang. 2023.
\newblock \href {https://doi.org/10.18653/v1/2023.findings-acl.551} {Do large language models know what they don't know?}
\newblock In \emph{Findings of the Association for Computational Linguistics: ACL 2023}, pages 8653--8665, Toronto, Canada. Association for Computational Linguistics.

\bibitem[{Zhang and Choi(2021)}]{zhang2021situatedqa}
Michael Zhang and Eunsol Choi. 2021.
\newblock \href {https://doi.org/10.18653/v1/2021.emnlp-main.586} {{SituatedQA}: Incorporating extra-linguistic contexts into {QA}}.
\newblock In \emph{Proceedings of the 2021 Conference on Empirical Methods in Natural Language Processing}, pages 7371--7387, Online and Punta Cana, Dominican Republic. Association for Computational Linguistics.

\end{thebibliography}
\clearpage
\appendix
\begingroup
\raggedbottom
\setlength{\floatsep}{9pt plus 2pt minus 2pt}
\setlength{\textfloatsep}{11pt plus 2pt minus 2pt}
\setlength{\intextsep}{10pt plus 2pt minus 2pt}
\renewcommand{\topfraction}{0.95}
\renewcommand{\bottomfraction}{0.85}
\renewcommand{\textfraction}{0.05}
\renewcommand{\floatpagefraction}{0.80}
\renewcommand{\dbltopfraction}{0.95}
\renewcommand{\dblfloatpagefraction}{0.80}

% Short prompt panels fit inside one column and move with their captions.
% The minipage moves to the next column when the remaining space is too small.
\newenvironment{econpromptpanel}{%
  \par\addvspace{8pt}\noindent\begin{minipage}{\columnwidth}%
  \ifdefined\nolinenumbers\nolinenumbers\fi
  \setlength{\emergencystretch}{1em}%
  \captionsetup{font=footnotesize,hypcap=false}%
  \tcbset{boxrule=0.5pt,boxsep=1pt,left=4pt,right=4pt,top=4pt,bottom=4pt,
    before skip=0pt,after skip=3pt}%
}{%
  \end{minipage}\par\addvspace{8pt}%
}

\FloatBarrier
\section{Triplet Extraction Details}
\label{app:extraction}

\paragraph{Embedding model.}
We use OpenAI's \texttt{text-embedding-3-small} model to encode each extracted treatment and outcome phrase into a dense vector representation.

\mypara{Similarity-based merging.}
Across the three independent extraction runs, minor surface-level wording differences (e.g., ``physical activity'' vs.\ ``physical exercise'') are common.
To merge such near-duplicate pairs, we compute the cosine similarity between the embedding vectors of each treatment phrase and each outcome phrase separately.
A candidate pair from one run is considered identical to a pair from another run if and only if \emph{both} the treatment similarity and the outcome similarity meet or exceed a threshold of $0.8$:
\begin{equation}
    \begin{aligned}
    \operatorname{sim}_{\cos}(\mathbf{t}_i, \mathbf{t}_j)
        &\geq 0.8 \\
    {}\land\quad \operatorname{sim}_{\cos}(\mathbf{o}_i, \mathbf{o}_j)
        &\geq 0.8,
    \end{aligned}
\end{equation}
where $\mathbf{t}$ and $\mathbf{o}$ denote the embedding vectors of the treatment and outcome phrases, respectively.

\mypara{Consensus filtering.}
After merging, we retain only those treatment--outcome pairs that appear in at least two out of the three runs ($\geq 2/3$ consensus).
This multi-run consensus strategy filters out hallucinated or spurious extractions that arise in only a single run.

\mypara{Sign assignment.}
For each retained pair, the directional sign label $s \in \{+,\; -,\; \texttt{None},\; \texttt{Mixed}\}$ is determined by majority vote across the runs in which the pair appears.

\FloatBarrier
\section{Multi-Critic Evaluation Details}
\label{app:critics}

\subsection{Critic Models}

Each extracted triplet is independently evaluated by three LLM critic models from different providers to reduce single-model bias:

\begin{itemize}\raggedright
    \item \textbf{Gemini} (Google): \texttt{gemini-3-flash-preview}
    \item \textbf{Grok} (xAI): \texttt{grok-4-1-fast-reasoning}
    \item \textbf{Qwen} (OpenRouter): \texttt{qwen3-vl-30b-a3b-thinking}
\end{itemize}

\noindent Each critic receives the extracted triplet (treatment, outcome, sign), the verbatim evidence paragraph(s), selected context from the extraction pipeline, and the original paper.
Scores from the three critics are averaged per criterion.

\subsection{Scoring Rubric (0--3 Scale)}

All six criteria are scored on a uniform four-point scale:

\begin{itemize}
    \item \textbf{3} -- \textit{Clearly supported / correct.} The aspect is unambiguously correct and well-grounded in the paper.
    \item \textbf{2} -- \textit{Mostly supported / minor ambiguity.} The aspect is largely correct but contains minor imprecision or ambiguity.
    \item \textbf{1} -- \textit{Weak support / substantial ambiguity.} The aspect has notable issues, e.g., mis-specification, vague grounding, or debatable interpretation.
    \item \textbf{0} -- \textit{Not supported / contradicted / clearly wrong.} The aspect is unsupported by or contradicts the paper.
\end{itemize}

\subsection{Quality Criteria}

% Each triplet is scored on the following six dimensions.\footnote{The full critic prompt is provided in Appendix~\ref{app:prompts}.}

\begin{enumerate}

    \item \textbf{Variable Extraction.}
    Whether the treatment and outcome are extracted as concise, concrete noun phrases explicitly mentioned or defined in the paper, with pronouns, acronyms, or shorthand correctly expanded.

    \item \textbf{Direction.}
    Whether the triplet correctly captures the intended causal direction (Treatment $\rightarrow$ Outcome) as asserted by the authors, without reversal due to ambiguous wording.

    \item \textbf{Sign.}
    Whether the assigned sign ($+$/$-$/None/Mixed) matches the authors' preferred or baseline estimate. Signs of $+$ or $-$ require statistical significance; \texttt{None} applies when the preferred estimate is insignificant; \texttt{Mixed} is reserved for genuinely heterogeneous headline results, not mere sensitivity to alternative specifications.

    \item \textbf{Causality.}
    Whether the relationship is presented as a causal effect claim supported by an identification strategy (e.g., instrumental variables, difference-in-differences), rather than a correlation, descriptive statistic, or theoretical conjecture.

    \item \textbf{Main Claim.}
    Whether the triplet represents a core causal claim emphasized by the authors as a central contribution (e.g., in the abstract, introduction, or conclusion), rather than a peripheral finding.

    \item \textbf{Context Appropriateness.}
    Whether the accompanying context includes the key elements required to interpret the causal claim without omitting paper-critical setting information, while avoiding encoding or implying the correctness of the triplet's sign, direction, or causal validity.

\end{enumerate}

\subsection{Conservative Filtering}

After scoring, we apply a conjunctive filtering rule: a triplet is retained only if (i) its critic-averaged score is at least 2.0 on \emph{every} individual criterion, and (ii) the sum of its six criterion-averaged scores is at least 15 (i.e., a mean of 2.5 or above across all criteria).
A triplet failing either condition is discarded.
This conservative strategy removes 27.3\% of candidate triplets, prioritizing precision over recall in the final dataset.

\FloatBarrier
\section{Source Paper Distribution}
\label{app:source_distribution}

Table~\ref{tab:venue_domain_distribution} reports the distribution of the final source papers by journal and domain. The final corpus contains 2,595 papers, drawn from top economics and finance journals matched from the NBER Working Paper series.

\begin{table}[!htbp]
\centering
\footnotesize
\setlength{\tabcolsep}{3pt}
\begin{tabular*}{\columnwidth}{@{\extracolsep{\fill}}lc@{}}
\toprule
\textbf{Journal} & \textbf{\# Papers} \\
\midrule
\multicolumn{2}{@{}l}{\textit{Economics journals}} \\
American Economic Review       & 751 \\
Quarterly Journal of Economics & 440 \\
Journal of Political Economy   & 258 \\
Review of Economic Studies     & 177 \\
Econometrica                   & 123 \\
\addlinespace[2pt]
\multicolumn{2}{@{}l}{\textit{Finance journals}} \\
Journal of Financial Economics & 356 \\
Journal of Finance             & 255 \\
Review of Financial Studies    & 235 \\
\midrule
\textbf{Total} & \textbf{2,595} \\
\bottomrule
\end{tabular*}
\caption{Distribution of source papers by journal and domain.}
\label{tab:venue_domain_distribution}
\end{table}
\FloatBarrier
\section{Economist Validation Details}
\label{app:validation}

Three economics professors individually assessed \textbf{206} triplets within their areas of expertise, using the same six 0--3 criteria described in Appendix~\ref{app:critics}. These expert evaluations allow us to validate both the LLM-critic scores and the downstream Step~4 filtering outcomes.

\mypara{Expert validation protocol.}
The validation was conducted by three economics professors: two author-evaluators and one external expert recruited through the authors' professional network. The evaluators were not separately compensated. The external expert was informed that the ratings would be used to validate the dataset construction pipeline and consented to this use.

\mypara{Validation summary.}
Table~\ref{tab:validation} summarizes score-level agreement, economist approval rates, and Step~4 filtering outcomes on the expert-rated triplets. The Step~4 acceptance rate is 76.2\%, lower than the economist approval rate of 83.5\%, suggesting that the filtering rule is relatively conservative.
The 86.6\% represents economist approval of
Step~4-retained triplets under six quality criteria,
rather than sign-label accuracy or an upper bound
on benchmark performance.

\begin{table}[!htbp]
\centering
\footnotesize
\setlength{\tabcolsep}{5pt}
\renewcommand{\arraystretch}{1.05}

\begin{tabular*}{\columnwidth}{@{\extracolsep{\fill}}>{\raggedright\arraybackslash}p{0.79\columnwidth}r@{}}
\toprule
Metric & Value \\
\midrule
\multicolumn{2}{@{}l}{\textit{Score-level agreement}} \\
Per-criterion MAE & 0.229 \\
Sum-of-scores MAE & 1.018 \\
\addlinespace[2pt]

\multicolumn{2}{@{}l}{\textit{Filtering and approval outcomes}} \\
Economist approval rate before filtering & 83.5\% \\
Step~4 acceptance rate & 76.2\% \\
Economist approval among Step~4 accepted triplets & 86.6\% \\
\bottomrule
\end{tabular*}
\caption{Economist validation summary for LLM-critic scores and Step~4 filtering.}
\label{tab:validation}
\end{table}

\mypara{Criterion-level expert scores.}
Table~\ref{tab:validation_by_criterion} reports economist scores for each of the six validation criteria. Across all criteria, mean scores are high, and most triplets receive scores of at least 2 out of 3.

\begin{table}[!htbp]
\centering
\footnotesize
\setlength{\tabcolsep}{3pt}
\renewcommand{\arraystretch}{1.05}

\begin{tabular*}{\columnwidth}{@{\extracolsep{\fill}}lcc@{}}
\toprule
Criterion & Mean score & Share $\geq 2$ \\
\midrule
Variable extraction & 2.87 / 3 & 97.57\% \\
Direction & 2.99 / 3 & 99.51\% \\
Sign score & 2.93 / 3 & 97.57\% \\
Causality & 2.83 / 3 & 94.66\% \\
Main claim & 2.72 / 3 & 90.78\% \\
Context & 2.94 / 3 & 98.06\% \\
\bottomrule
\end{tabular*}

\caption{Criterion-level economist validation scores. Mean scores are reported on a 0--3 scale, and the final column reports the share of triplets receiving a score of at least 2.}
\label{tab:validation_by_criterion}
\end{table}

\FloatBarrier
\section{Cross-Context Pair Construction and Validation}
\label{app:pair_construction}

\subsection{Cross-Context Pair Construction}
\label{app:pair_construction_details}

We construct the source--target pairs used in Task~2 in four stages:

\begin{itemize}

    \item \textbf{Citation-linked source papers.}
    For each candidate pair of causal triplets, we require the source
    papers from which the two triplets were extracted to have a direct
    citation relationship in either direction: the source paper cites
    the target paper, or the target paper cites the source paper.
    This restricts candidate relations to studies that are directly
    connected in the economics or finance literature.

    \item \textbf{Shared JEL classification.}
    We further require the two source papers to share at least one
    three-character JEL code, such as \texttt{J31} (wage level and structure)
    or \texttt{G12} (asset pricing). This provides a coarse topical
    restriction before matching individual causal relations.

    \item \textbf{Embedding-based candidate selection.}
    Treatment and outcome phrases are embedded separately using
    \texttt{text-embedding-3-small}. For each candidate pair, we use the
    smaller of the treatment and outcome cosine similarities as the
    matching score. Among 67,128 candidate pairs, we retain the top
    10\%, corresponding to a similarity threshold of approximately
    0.36 and leaving 6,712 pairs for further evaluation.

    \item \textbf{LLM-based comparability filtering.}
    The remaining pairs are evaluated by GPT-5.6 Luna using a
    five-question rubric:
    \begin{itemize}
        \item \textbf{Q1/Q3:} conceptual containment of the treatment
        and outcome, respectively;
        \item \textbf{Q2/Q4:} economic comparability of the treatment
        and outcome, respectively;
        \item \textbf{Q5:} economic comparability of the
        treatment--outcome causal relationship.
    \end{itemize}

    A pair is rejected if either the treatment or outcome is judged
    \texttt{DISTINCT} under Q1/Q3. Q2, Q4, and Q5 are scored from
    0 to 3, and we retain only pairs receiving a score of at least 2
    on all three criteria. This procedure yields 415 source--target pairs covering 273 distinct target triplets.

\end{itemize}

The full matching prompt, including the definitions and examples used
for all five criteria, is provided in
Appendix~\ref{app:pair_matching_prompt}.

\subsection{Economist Assessment of Pair Matching}
\label{app:economist-matching}

Three economists independently evaluated ten candidate pairs,
balanced between five accepted and five rejected by the
LLM-based filter. Table~\ref{tab:economist-matching-validation}
reports agreement with the economist majority for each
criterion and the final inclusion decision. For final inclusion,
all five criteria were applied within each evaluator before
majority voting.

\begin{table}[!htbp]
\centering
\footnotesize
\setlength{\tabcolsep}{2pt}
\renewcommand{\arraystretch}{1.1}
\begin{tabular*}{\columnwidth}{@{\extracolsep{\fill}}lc@{}}
\toprule
Criterion & Agreement with LLM \\
\midrule
Treatment containment (Q1)   & 6/10 \\
Treatment comparability (Q2) & 9/10 \\
Outcome containment (Q3)     & 9/10 \\
Outcome comparability (Q4)   & 10/10 \\
Relation comparability (Q5)  & 8/10 \\
\midrule
\textbf{Overall Agreement}   & \textbf{8/10} \\
\bottomrule
\end{tabular*}
\caption{Agreement between the economist majority and the LLM across ten candidate pairs.}
\label{tab:economist-matching-validation}
\end{table}

% As shown in Table~\ref{tab:economist-matching-validation},
% the validator agreed with the economist majority on 8/10
% final inclusion decisions (Cohen's $\kappa = 0.60$).
% The economists endorsed three of the five accepted pairs
% and rejected all five rejected pairs.
% However, the small, balanced audit and limited
% inter-economist consensus preclude conclusions about
% the reliability of the full matching pool.

\subsection{Sensitivity to Pair Comparability Thresholds}
\label{app:sensitivity_pair_matching}
We examine whether Task 2 results are sensitive to
the strictness of source--target pair matching.
We vary the minimum sum of Q2 (treatment comparability),
Q4 (outcome comparability), and Q5 (relation comparability)
scores, while requiring each score to be at least 2.
Thresholds of 6, 7, and 8 retain 415, 170, and 17
source--target pairs, respectively.
As shown in Table~\ref{tab:matching_sensitivity},
sign-mismatch accuracy remains substantially below
overall Task 2 accuracy across all thresholds for
both model groups. This suggests that the observed
sign-mismatch difficulty is not driven solely by
loosely matched pairs.
\begin{table}[!htbp]
\centering
\footnotesize
\setlength{\tabcolsep}{2pt}
\begin{tabular}{@{}llccc@{}}
\toprule
Cutoff ($N$) & Group & \shortstack{T2\\Overall} & \shortstack{T2\\Mismatch} & T3 \\
\midrule
$\geq 6$ (415)
& Closed & \shortstack{0.779\\(0.558)} & \shortstack{0.507\\(0.444)} & \shortstack{0.637\\(0.454)} \\
& Open & \shortstack{0.694\\(0.397)} & \shortstack{0.332\\(0.218)} & \shortstack{0.501\\(0.279)} \\
\midrule
$\geq 7$ (170)
& Closed & \shortstack{0.759\\(0.556)} & \shortstack{0.464\\(0.412)} & \shortstack{0.588\\(0.431)} \\
& Open & \shortstack{0.696\\(0.437)} & \shortstack{0.302\\(0.222)} & \shortstack{0.480\\(0.281)} \\
\midrule
$\geq 8$ (17)
& Closed & \shortstack{0.868\\(0.588)} & \shortstack{0.417\\(0.201)} & \shortstack{0.556\\(0.393)} \\
& Open & \shortstack{0.815\\(0.505)} & \shortstack{0.286\\(0.095)} & \shortstack{0.426\\(0.245)} \\
\bottomrule
\end{tabular}
\caption{Sensitivity to pair-comparability thresholds
based on the sum of Q2, Q4, and Q5 scores.
Values report accuracy on the first line and macro-F1 in
parentheses on the second line, averaged across
closed-source (12) and open-source (7) models.
$N$ denotes the number of retained source--target pairs.}
\label{tab:matching_sensitivity}
\end{table}

\FloatBarrier
\section{Economist Reference Study}
\label{app:economist-eval}

\mypara{Purpose and overview.}
We conducted a small-scale economist evaluation to provide
a human reference for interpreting LLM performance on
the benchmark tasks.
Three economists each answered ten questions per task,
using the same benchmark inputs and task instructions
as the LLMs.

\mypara{Comparison with LLMs.}
Table~\ref{tab:economist-eval} compares economist and LLM
accuracy on the questions assigned to the economists.
Economists scored lower than both LLM groups on Task~1
in Economics and Finance.
On Task~2, their overall accuracy was numerically similar
to the closed-source LLM average
(70.0\% versus 68.9\%), while their accuracy on
sign-mismatch items was higher
(47.2\% versus 36.3\%).
On Task~3, economist accuracy was 23.3\%, compared with
50.3\% for closed-source LLMs and 39.0\% for open-source
LLMs.
The relative performance of economists and LLMs thus
varied across tasks in this sample.

% Before finalizing, confirm the aggregation used for
% economist accuracy and ensure comparable weighting
% in the human--LLM comparison.
\begin{table}[!htbp]
\centering
\footnotesize
\setlength{\tabcolsep}{2pt}
\renewcommand{\arraystretch}{1.1}
\begin{tabular*}{\columnwidth}
{@{\extracolsep{\fill}}lccccc@{}}
\toprule
& \multicolumn{2}{c}{Task 1}
& \multicolumn{2}{c}{Task 2}
& Task 3 \\
\cmidrule(lr){2-3}
\cmidrule(lr){4-5}
& Econ
& Finance
& Overall
& \shortstack{Sign\\mismatch}
& \\
\midrule
Economists  & 35.6 & 70.0 & 70.0 & 47.2 & 23.3 \\
Closed LLMs & 55.8 & 91.7 & 68.9 & 36.3 & 50.3 \\
Open LLMs   & 47.9 & 85.5 & 61.0 & 22.6 & 39.0 \\
\bottomrule
\end{tabular*}
\caption{
Accuracy (\%) of economists and LLMs on the
economist-evaluation questions.
Three economists each answered ten questions per task.
LLM results are averaged across models within each group.
}
\label{tab:economist-eval}
\end{table}

\mypara{Results by question allocation.}
Each economist answered five shared questions and five
individualized questions per task.
Table~\ref{tab:economist-allocation} reports response-pooled
accuracy for these two sets and agreement among economists
on the shared questions.
Each set contributes 15 responses per task; the shared
set contains five distinct questions answered by all
three economists.

On the shared questions, agreement was higher on Task~2
than on Task~3.
Task~3 combined low accuracy with substantial disagreement,
indicating that this setting was also challenging for
the economists in this sample.
For each task, the individualized set yielded one
additional correct response out of 15 compared with
the shared set.
These differences are descriptive comparisons between
question sets, not estimates of an expertise effect.

% Before finalizing, specify the agreement statistic
% and its calculation in the caption.
\begin{table}[!htbp]
\centering
\footnotesize
\setlength{\tabcolsep}{3pt}
\renewcommand{\arraystretch}{1.2}
\begin{tabular*}{\columnwidth}
{@{\extracolsep{\fill}}lccc@{}}
\toprule
Task
& \shortstack{Shared\\accuracy}
& \shortstack{Individualized\\accuracy}
& \shortstack{Agreement\\(\%)} \\
\midrule
Task 1 & 7/15 (46.7)  & 8/15 (53.3)  & 73.3 \\
Task 2 & 10/15 (66.7) & 11/15 (73.3) & 86.7 \\
Task 3 & 3/15 (20.0)  & 4/15 (26.7)  & 33.3 \\
\bottomrule
\end{tabular*}
\caption{
Economist results by question allocation.
Accuracy entries show correct/total responses
(percentage), pooled across the three economists.
Each economist answered five shared and five
individualized questions per task.
Task~1 combines Economics and Finance.
Agreement is reported only for the five shared questions
answered by all three economists.
}
\label{tab:economist-allocation}
\end{table}

\mypara{Participants and question selection.}
The participants were the same three economists who
conducted the earlier dataset validation, and they
specialized in different subfields.
This task evaluation took place more than eight months
after the dataset validation.

For each task, five questions were randomly sampled
and shared across all evaluators, providing a common
basis for comparing responses and examining agreement.
The five individualized questions were selected to
include topics connected to each evaluator's earlier
validation experience.
Specifically, they were drawn from papers with keywords
similar to those of papers that the corresponding
economist had previously reviewed.
These were different papers from those used in the
earlier validation.
This selection criterion linked questions to previously
encountered topics, but did not directly match them to
each economist's research specialization.

Given the small number of evaluators and questions,
we did not design separate within-subfield and
outside-subfield comparisons: subdividing the sample
would leave very few observations for each evaluator
and category.
Questions were therefore not restricted to each
economist's own subfield.
The shared--individualized distinction describes
question allocation rather than a division by expertise.

\mypara{Interpretation.}
These results provide an exploratory reference for
economists' performance under the benchmark conditions.
The small number of evaluators and questions limits
precision and generalization, while familiarity with
the relevant empirical literature may vary across
evaluators and items.
The comparison does not establish a general human--LLM
ranking or a performance ceiling.
Agreement on shared questions describes consistency
among evaluators; it does not by itself establish
correctness or the sufficiency of the supplied information.

\FloatBarrier

\FloatBarrier
\section{Calibration and Uncertainty Analysis}
\label{app:calibration}

Accuracy alone does not indicate whether a model
recognizes when its predictions are uncertain
or insufficiently supported. 
Prior work has examined LLMs' ability to
recognize unanswerable questions
\citep{yin-etal-2023-large} and to express
calibrated uncertainty
\citep{tian-etal-2023-just,xiong2024uncertainty}.
We therefore conduct two supplementary analyses
of GPT-4o on Task~1: whether it abstains
when contextual information is withheld,
and whether its reported confidence is
calibrated to empirical accuracy.

\mypara{Abstention under missing context.}
We remove the context from each Task~1 instance, provide only the
treatment--outcome pair, and add an explicit \texttt{Unknown} option.
Despite the missing contextual information, GPT-4o selects
\texttt{Unknown} in only 6.8\% of Economics and 9.4\% of Finance
instances (Figure~\ref{fig:unknown_ratio}). Thus, the model generally
commits to a causal sign even when the contextual information available
in the original task is removed.

\mypara{Confidence calibration.}
We next evaluate whether GPT-4o's confidence is aligned with accuracy
under the original Task~1 setting. Using log-probabilities over
$\mathcal{Y}=\{+,-,\texttt{None},\texttt{Mixed}\}$, we compute Expected
Calibration Error (ECE)~\cite{guo2017calibration}:
\begin{equation}
    \mathrm{ECE}
    =
    \sum_{m=1}^{M}
    \frac{|B_m|}{N}
    \left|
    \mathrm{acc}(B_m)-\mathrm{conf}(B_m)
    \right|,
\end{equation}
where $B_m$ denotes the instances in confidence bin $m$. We additionally
compute label-specific ECE by restricting the evaluation to instances
with each gold label.

As shown in Figure~\ref{fig:ece_error}, calibration is substantially
worse for \texttt{None} and \texttt{Mixed} than for directional labels.
The largest errors occur for \texttt{None} in Finance
(ECE = 0.839) and \texttt{Mixed} in Economics (ECE = 0.743).
Together with the lower accuracy on these labels
(Table~\ref{tab:sign_accuracy_by_task}), these results suggest that
GPT-4o's weakest causal-sign categories are also associated with poor
confidence calibration.

\begin{figure}[!htbp]
    \centering
    \begin{subfigure}[t]{0.98\columnwidth}
        \centering
        \includegraphics[width=\linewidth]{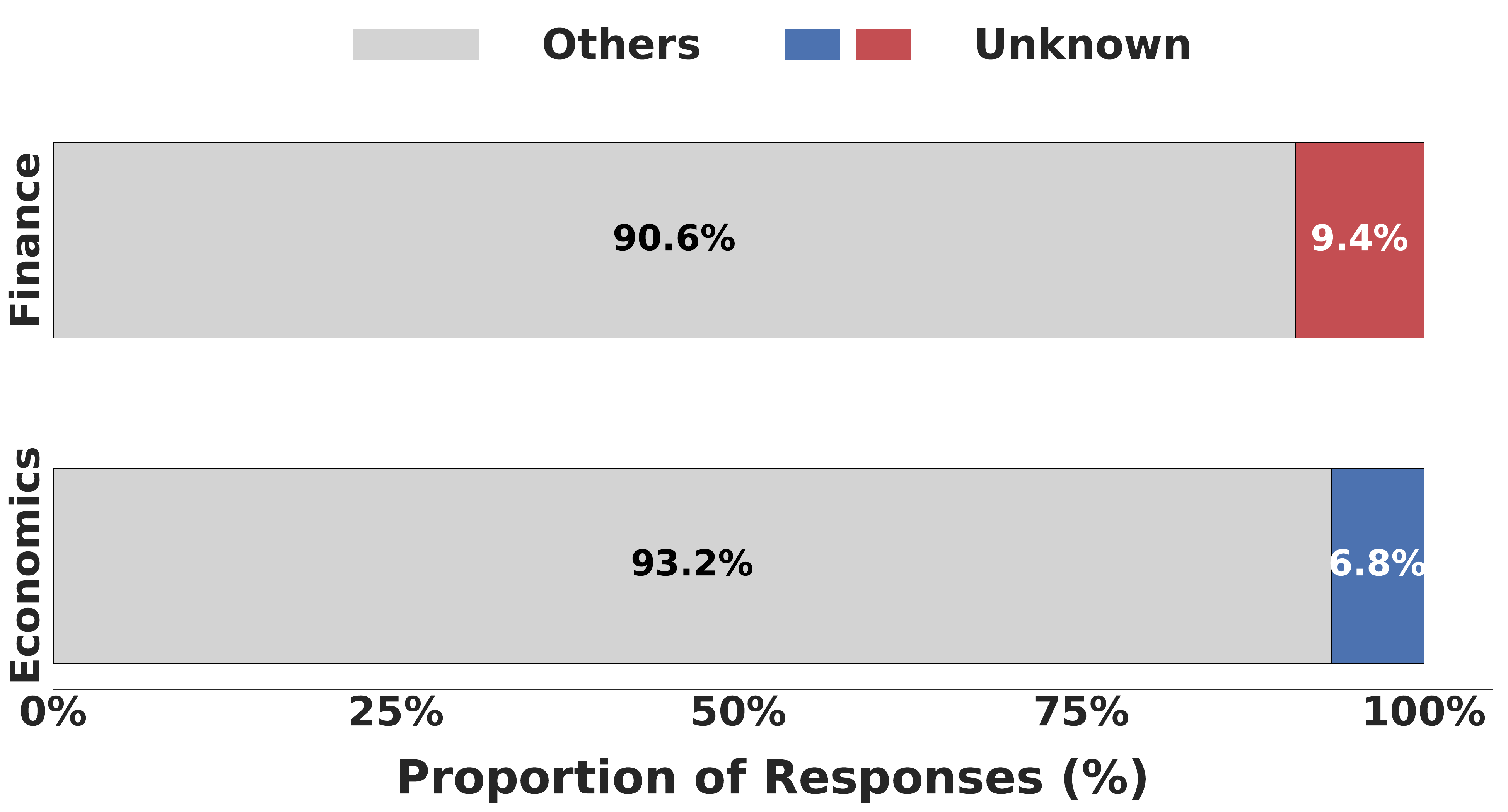}
        \caption{Rate of \texttt{Unknown} responses.}
        \label{fig:unknown_ratio}
    \end{subfigure}
    \par\smallskip
    \begin{subfigure}[t]{0.98\columnwidth}
        \centering
        \includegraphics[width=\linewidth]{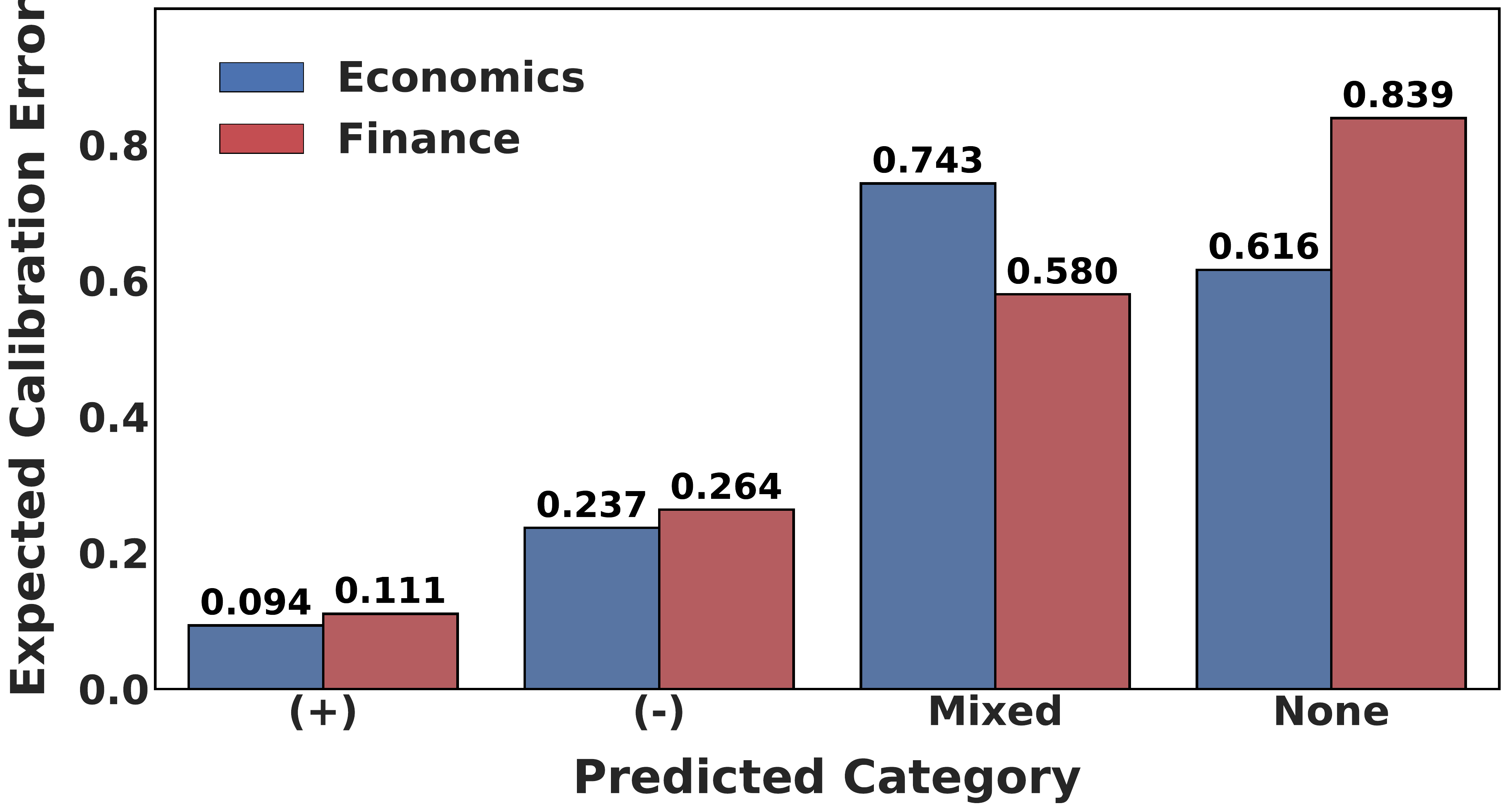}
        \caption{ECE by causal-sign category.}
        \label{fig:ece_error}
    \end{subfigure}
    \caption{GPT-4o uncertainty analysis on Task~1.}
    \label{fig:calibration}
\end{figure}

\FloatBarrier
\setlength{\intextsep}{6pt plus 1pt minus 1pt}
\setlength{\textfloatsep}{8pt plus 1pt minus 1pt}
\section{Robustness Analyses}
\label{app:robustness_analyses}

We conduct targeted robustness analyses to examine
potential pretraining exposure, source memorization,
prompt sensitivity, context sufficiency, and lexical
shortcuts. Each analysis focuses on the task most
directly relevant to the corresponding concern.

\FloatBarrier
\subsection{Post-Cutoff Evaluation}
\label{app:post_cutoff}

We evaluate Task~2 on 101 pairs whose target NBER
working papers were issued after August 31, 2025,
examining whether cross-context predictions generalize
to more recent findings with potentially lower
pretraining exposure.

Pair construction follows the main Task~2 procedure:
source and target papers share a three-character JEL code,
the minimum treatment/outcome embedding similarity
is at least 0.36, and all three LLM-based comparability
scores are at least 2. Source papers may also come
from NBER studies outside the original top-five
economics and top-three finance journals.

\begin{table}[!htbp]
\centering
\footnotesize
\renewcommand{\arraystretch}{0.92}
\setlength{\tabcolsep}{3pt}
\begin{tabular*}{\columnwidth}{@{\extracolsep{\fill}}lrrr@{}}
\toprule
Model & Main & Post-cutoff & $\Delta$ \\
\midrule
Gemini 3.1 Pro    & 90.60 & 82.18 & -8.42 \\
Gemini 3 Flash    & 86.27 & 79.21 & -7.06 \\
Gemini 2.5 Pro    & 77.59 & 72.28 & -5.31 \\
Gemini 2.5 Flash  & 76.39 & 80.20 & +3.81 \\
GPT-5.2           & 76.87 & 69.31 & -7.56 \\
GPT-5 mini        & 75.66 & 78.22 & +2.56 \\
GPT-5 nano        & 73.98 & 75.25 & +1.27 \\
GPT-4o            & 64.58 & 58.42 & -6.16 \\
GPT-4o mini       & 76.39 & 72.28 & -4.11 \\
Claude Opus 4.6   & 79.28 & 84.16 & +4.88 \\
Claude Sonnet 4.6 & 80.24 & 71.29 & -8.95 \\
Claude Haiku 4.5  & 76.39 & 55.45 & -20.94 \\
\midrule
\textbf{Closed mean (12)}
& \textbf{77.85} & \textbf{73.18} & \textbf{-4.67} \\
\midrule
Llama 3.3 70B     & 79.52 & 79.21 & -0.31 \\
Llama 3.1 8B      & 65.78 & 73.27 & +7.48 \\
Llama 3.2 3B      & 60.00 & 52.48 & -7.52 \\
Llama 3.2 1B      & 66.75 & 50.50 & -16.25 \\
Qwen3 32B         & 77.83 & 86.14 & +8.31 \\
Qwen3 14B         & 64.10 & 71.29 & +7.19 \\
Qwen3 8B          & 72.05 & 80.20 & +8.15 \\
\midrule
\textbf{Open mean (7)}
& \textbf{69.43} & \textbf{70.44} & \textbf{+1.01} \\
\midrule
\textbf{Overall mean (19)}
& \textbf{74.75} & \textbf{72.17} & \textbf{-2.58} \\
\bottomrule
\end{tabular*}
\caption{Task~2 accuracy (\%) on the main 415-pair
and post-cutoff 101-pair sets. $\Delta$ denotes
post-cutoff minus main accuracy in percentage points.}
\label{tab:post_cutoff_models}
\end{table}

As shown in Table~\ref{tab:post_cutoff_models},
performance does not decline consistently across model
groups. However, the sets differ in composition and
source coverage, so these differences cannot be
attributed solely to pretraining exposure.

\FloatBarrier
\subsection{Source-Paper Recall}
\label{app:source_recall}

To probe whether Task~1 accuracy reflects explicit
memorization of source studies, we test whether models
can identify the underlying papers from the standalone
Task~1 inputs, without source examples.

We stratify eligible items by NBER issue year
(1991--2025) and domain, sampling one paper and
one item per available stratum. This yields 68 items
from distinct papers (34 Economics, 34 Finance).
All 19 models receive the context, treatment, and
outcome without the gold sign or bibliographic metadata.
Search and browsing tools are unavailable.

We score four recall criteria: correct NBER or
published title (Title), at least one correct coauthor
(Author), correct journal (Journal), and all three
jointly correct (Joint). Minor title variations and
standard journal abbreviations are accepted; unknown
responses count as failures. Ambiguous matches are
manually audited.

\begin{table}[!htbp]
\centering
\scriptsize
\setlength{\tabcolsep}{2.2pt}
\renewcommand{\arraystretch}{0.92}
\begin{tabular*}{\columnwidth}{@{\extracolsep{\fill}}lrrrrr@{}}
\toprule
Model & Title & Author & Journal & Joint & $r$ \\
\midrule
GPT-4o mini       & 8.8 & 7.4 & 19.1 & 1.5 & 0.105 \\
GPT-4o            & 8.8 & 29.4 & 17.6 & 5.9 & 0.008 \\
GPT-5 nano        & 5.9 & 10.3 & 5.9 & 1.5 & 0.087 \\
GPT-5 mini        & 19.1 & 25.0 & 22.1 & 14.7 & 0.144 \\
GPT-5.2           & 11.8 & 25.0 & 19.1 & 7.4 & -0.098 \\
Gemini 3.1 Pro    & 82.4 & 86.8 & 77.9 & 77.9 & 0.012 \\
Gemini 3 Flash    & 64.7 & 82.4 & 67.6 & 58.8 & -0.124 \\
Gemini 2.5 Flash  & 14.7 & 33.8 & 29.4 & 4.4 & 0.124 \\
Gemini 2.5 Pro    & 55.9 & 72.1 & 63.2 & 47.1 & 0.075 \\
Claude Haiku 4.5  & 13.2 & 30.9 & 13.2 & 4.4 & 0.134 \\
Claude Sonnet 4.6 & 22.1 & 54.4 & 33.8 & 22.1 & 0.094 \\
Claude Opus 4.6   & 32.4 & 58.8 & 42.6 & 26.5 & -0.084 \\
\midrule
Llama 3.2 1B      & 1.5 & 1.5 & 5.9 & 0.0 & -- \\
Llama 3.2 3B      & 2.9 & 2.9 & 14.7 & 0.0 & -- \\
Llama 3.1 8B      & 0.0 & 8.8 & 0.0 & 0.0 & -- \\
Llama 3.3 70B     & 10.3 & 23.5 & 25.0 & 4.4 & 0.134 \\
Qwen3 8B          & 8.8 & 4.4 & 14.7 & 0.0 & -- \\
Qwen3 14B         & 8.8 & 7.4 & 14.7 & 0.0 & -- \\
Qwen3 32B         & 7.4 & 5.9 & 19.1 & 0.0 & -- \\
\bottomrule
\end{tabular*}
\caption{Source-paper recall on 68 Task~1 items.
Recall rates are percentages. $r$ denotes the
within-model correlation between joint recall
and Task~1 correctness; undefined values are
shown as --.}
\label{tab:source_recall_model}
\end{table}

Mean joint recall is 22.7\% for closed-source models
and 0.6\% for open-source models. Recall is not
consistently associated with Task~1 correctness,
and no association survives multiple-comparison
correction. Failure to identify a source, however,
does not exclude implicit recall of its findings.

\FloatBarrier
\subsection{Citation Frequency and Task 1 Accuracy}
\label{app:citation_regression}

We examine the association between source-paper
citation counts and Task~1 correctness across
19 models. Citation counts are obtained from
OpenAlex (July 13, 2026), taking the maximum
count across matched NBER and published versions.

We estimate a logistic regression:

\begin{equation}
\begin{aligned}
&\operatorname{logit} P(Y_{im}=1) \\
&= \alpha + \beta\,\mathrm{Citations}_i \\
&\quad + \mathrm{Year\,FE}
+ \mathrm{Model\,FE} \\
&\quad + \mathrm{Gold\text{-}sign\,FE}
+ \mathrm{Journal\,FE},
\end{aligned}
\end{equation}

where $Y_{im}$ indicates whether model $m$
correctly predicts item $i$. Year fixed effects
refer to NBER working-paper issue years.
Standard errors are clustered by source paper
(merged-paper entity).

The analysis includes 34,276 observations
(1,804 items $\times$ 19 models) from 1,133
paper clusters, excluding three items with
missing citation counts. Invalid responses
are counted as incorrect.

\begin{table}[!htbp]
\centering
\footnotesize
\renewcommand{\arraystretch}{0.92}
\setlength{\tabcolsep}{3pt}
\begin{tabular}{@{}lc@{}}
\toprule
Statistic & Estimate \\
\midrule
Coefficient ($\beta$) & $0.000256$ \\
Clustered SE & $0.000357$ \\
95\% CI & $[-0.000443,\;0.000955]$ \\
$p$-value & $0.4724$ \\
\bottomrule
\end{tabular}
\caption{Logistic regression of Task~1 correctness
on raw source-paper citation counts, controlling
for year, model, gold-sign, and journal fixed effects.
Standard errors are clustered at the paper level.}
\label{tab:citation_regression}
\end{table}

\FloatBarrier
\subsection{Prompt Sensitivity}
\label{app:sensitivity_analyses}

We examine whether source-conditioned predictions
depend on prompt formulation. We focus on Tasks~2
and~3, which share the same prompt structure but
provide correct and misleading source signs,
respectively.

BASE includes an expert-economist role and explicit
sign definitions. SIMPLE removes these elements
and shortens the task instruction, while EXPERT adds
an instruction emphasizing the target context over
source examples.

\begin{table}[!htbp]
\centering
\footnotesize
\renewcommand{\arraystretch}{0.92}
\setlength{\tabcolsep}{3pt}
\begin{tabular}{@{}llrrrr@{}}
\toprule
 & & \multicolumn{2}{c}{Task 2}
     & \multicolumn{2}{c}{Task 3} \\
\cmidrule(lr){3-4}\cmidrule(l){5-6}
Group & Prompt & Acc. & F1 & Acc. & F1 \\
\midrule
Closed (12) & BASE   & 77.85 & 55.80 & 63.69 & 45.42 \\
            & SIMPLE & 80.54 & 53.39 & 68.17 & 44.99 \\
            & EXPERT & 78.55 & 56.55 & 66.73 & 47.46 \\
\midrule
Open (7)    & BASE   & 69.43 & 39.67 & 50.05 & 27.91 \\
            & SIMPLE & 71.12 & 38.36 & 57.68 & 29.31 \\
            & EXPERT & 69.29 & 39.09 & 55.17 & 29.20 \\
\bottomrule
\end{tabular}
\caption{Prompt sensitivity on Tasks~2 and~3.
Accuracy and macro-F1 (\%) are equally weighted
model averages. }
\label{tab:prompt_sensitivity}
\end{table}

\FloatBarrier
\subsection{Context Length Ablation}
\label{app:context_length_ablation}

To assess whether the default 100-word contexts
omit information needed for Task~2 prediction,
we regenerate both source and target contexts
with a 300-word cap by repeating Steps~2 and~3
of the Task~1 extraction pipeline.
We evaluate the same 415 source--target pairs
across 19 models (12 closed and 7 open),
keeping the treatment--outcome pairs, source signs,
and evaluation prompts unchanged.
Extending the context length yields only modest
accuracy gains, averaging 0.66 percentage points
across all models.

\begin{table}[!htbp]
\centering
\footnotesize
\renewcommand{\arraystretch}{0.92}
\setlength{\tabcolsep}{3pt}
\begin{tabular*}{\columnwidth}{@{\extracolsep{\fill}}lrrr@{}}
\toprule
Model & 100 words & 300 words & $\Delta$ \\
\midrule
\multicolumn{4}{@{}l}{\textit{Closed models}} \\
Gemini 3.1 Pro Preview   & 90.60 & 91.08 & +0.48 \\
Gemini 3 Flash Preview  & 86.27 & 88.43 & +2.17 \\
Gemini 2.5 Pro          & 77.59 & 79.52 & +1.93 \\
Gemini 2.5 Flash        & 76.39 & 76.63 & +0.24 \\
GPT-5.2                & 76.87 & 79.76 & +2.89 \\
GPT-5 mini             & 75.66 & 76.63 & +0.96 \\
GPT-5 nano             & 73.98 & 74.94 & +0.96 \\
GPT-4o                 & 64.58 & 59.04 & -5.54 \\
GPT-4o mini            & 76.39 & 78.55 & +2.17 \\
Claude Opus 4.6        & 79.28 & 79.28 & +0.00 \\
Claude Sonnet 4.6      & 80.24 & 80.48 & +0.24 \\
Claude Haiku 4.5       & 76.39 & 76.14 & -0.24 \\
Closed mean (12)       & 77.85 & 78.37 & +0.52 \\
\midrule
\multicolumn{4}{@{}l}{\textit{Open models}} \\
Llama 3.3 70B          & 79.52 & 80.72 & +1.20 \\
Llama 3.1 8B           & 65.78 & 63.13 & -2.65 \\
Llama 3.2 3B           & 60.00 & 67.71 & +7.71 \\
Llama 3.2 1B           & 66.75 & 60.72 & -6.02 \\
Qwen3 32B              & 77.83 & 80.72 & +2.89 \\
Qwen3 14B              & 64.10 & 74.94 & +10.84 \\
Qwen3 8B               & 72.05 & 64.34 & -7.71 \\
Open mean (7)          & 69.43 & 70.33 & +0.90 \\
\midrule
Overall mean (19)      & 74.75 & 75.41 & +0.66 \\
\bottomrule
\end{tabular*}
\caption{Task~2 accuracy (\%) under different
context-length caps on the same 415 pairs.
$\Delta$ is in percentage points.}
\label{tab:context_length_ablation}
\end{table}

\FloatBarrier
\subsection{TF-IDF Lexical Baseline}
\label{app:tfidf_baseline}

To test whether simple lexical associations explain
Task~1 performance, we train multinomial logistic
regression on TF-IDF features extracted from the
context, treatment/outcome (T/O), or both.
We focus on Task~1 because it predicts the causal
sign from target text alone, without source-example
interactions.

We use paper-grouped nested cross-validation to
prevent triplets from the same source paper appearing
in both training and test folds. Regularization is
selected using validation macro-F1.

\begin{table}[!htbp]
\centering
\footnotesize
\setlength{\tabcolsep}{3pt}
\renewcommand{\arraystretch}{0.92}
\begin{tabular}{@{}lcc@{}}
\toprule
Model / Input & Economics & Finance \\
\midrule
LLM mean (19) & 0.690 (0.428) & 0.675 (0.421) \\
Closed LLMs (12) & 0.734 (0.485) & 0.729 (0.478) \\
Open LLMs (7) & 0.613 (0.330) & 0.582 (0.322) \\
\midrule

TF-IDF: Context & 0.460 (0.284) & 0.473 (0.296) \\
TF-IDF: T/O & 0.436 (0.275) & 0.405 (0.263) \\
TF-IDF: Context + T/O & 0.465 (0.281) & 0.467 (0.282) \\
\bottomrule
\end{tabular}
\caption{Task~1 accuracy (macro-F1), on a 0--1
scale.}
\label{tab:tfidf_baseline}
\end{table}

The best TF-IDF accuracies are 46.5\% for Economics
and 47.3\% for Finance, substantially below the
corresponding LLM means. A temporal split trained
on 1991--2019 and evaluated on 2020--2025
achieves 50.19\% accuracy. These results suggest
that simple lexical cues alone are insufficient
to explain Task~1 performance, although they do
not exclude more complex associative strategies.

\subsection{Within-Instance Source-Sign Sensitivity}
\label{app:within_instance_source_sign}

We compare stored Task~3 responses across three
corrupted source-sign variants per pair, holding
both contexts and treatment--outcome pairs fixed.
Of 7,885 model--pair instances, 7,862 have three
valid predictions.
To distinguish source-sign reliance from fixed
label preferences, we compare observed agreement
with a baseline that permutes the displayed signs
while holding predictions fixed.

\begin{table}[t]
\centering
\footnotesize
\setlength{\tabcolsep}{2pt}
\renewcommand{\arraystretch}{1.05}
\begin{tabular*}{\columnwidth}{@{\extracolsep{\fill}}lrrrr@{}}
\toprule
& \shortstack{Changed\\(\%)}
& \shortstack{Observed\\agree. (\%)}
& \shortstack{Permuted\\agree. (\%)}
& \shortstack{Excess\\(pp)} \\
\midrule
Closed (12)  & 41.77 & 29.17 & 16.02 & +13.15 \\
Open (7)    & 60.91 & 37.63 & 18.71 & +18.92 \\
Overall (19)& 48.82 & 32.29 & 17.01 & +15.28 \\
\bottomrule
\end{tabular*}
\caption{
Task~3 within-instance source-sign sensitivity.
Changed denotes at least one prediction change
across the three variants.
}
\label{tab:within_instance_source_sign}
\end{table}

As shown in Table~\ref{tab:within_instance_source_sign},
48.82\% of instances exhibit prediction changes.
Source-sign agreement exceeds the permutation
baseline by 15.28\,pp (95\% CI: [13.51, 17.08]),
with larger excess agreement among open models.
The excess persists on 291 pairs without
correction reruns (+15.37\,pp;
95\% CI: [13.41, 17.34]).
Confidence intervals use 10,000 target-paper-cluster
bootstrap resamples.

\setlength{\intextsep}{10pt plus 2pt minus 2pt}
\setlength{\textfloatsep}{11pt plus 2pt minus 2pt}
\FloatBarrier
\section{Confidence Intervals and Performance Breakdowns}% \subsection{Domain and Temporal Analysis}

\subsection{Confidence Intervals for Main Results}
\label{app:ci-layouts}
We report 95\% percentile confidence intervals from 10,000 paper-cluster bootstrap resamples. Accuracy and four-class macro-F1 are recomputed in each resample.

\begin{table*}[!t]
\ifdefined\nolinenumbers\nolinenumbers\fi
\captionsetup{hypcap=false}
\centering
{\footnotesize
\setlength{\tabcolsep}{1.8pt}
\renewcommand{\arraystretch}{0.92}
\begin{tabular*}{\linewidth}{@{\extracolsep{\fill}} l c c c c c @{}}
\toprule
Model & T1 Econ & T1 Finance & T2 Overall & T2 Mismatch & T3 \\
\midrule
Gemini 3.1 Pro & 0.838\,[0.813,0.863] & \textbf{0.831}\,[0.803,0.860] & \textbf{0.906}\,[0.866,0.939] & \textbf{0.806}\,[0.725,0.881] & \textbf{0.877}\,[0.827,0.917] \\
Gemini 3 Flash & \textbf{0.839}\,[0.815,0.864] & 0.821\,[0.791,0.850] & 0.863\,[0.813,0.905] & 0.621\,[0.525,0.725] & 0.736\,[0.678,0.787] \\
Gemini 2.5 Pro & 0.774\,[0.744,0.803] & 0.757\,[0.724,0.790] & 0.776\,[0.721,0.827] & 0.581\,[0.488,0.678] & 0.685\,[0.620,0.742] \\
Gemini 2.5 Flash & 0.752\,[0.721,0.782] & 0.748\,[0.717,0.778] & 0.764\,[0.698,0.824] & 0.476\,[0.381,0.583] & 0.453\,[0.402,0.501] \\
GPT-5.2 & 0.729\,[0.697,0.761] & 0.734\,[0.702,0.764] & 0.769\,[0.702,0.830] & 0.476\,[0.384,0.573] & 0.664\,[0.586,0.736] \\
GPT-5 mini & 0.699\,[0.666,0.732] & 0.701\,[0.667,0.735] & 0.757\,[0.692,0.815] & 0.427\,[0.331,0.540] & 0.598\,[0.539,0.657] \\
GPT-5 nano & 0.682\,[0.650,0.715] & 0.672\,[0.638,0.706] & 0.740\,[0.684,0.791] & 0.387\,[0.301,0.481] & 0.574\,[0.515,0.634] \\
GPT-4o & 0.662\,[0.630,0.694] & 0.658\,[0.624,0.693] & 0.646\,[0.573,0.713] & 0.339\,[0.250,0.445] & 0.404\,[0.354,0.449] \\
GPT-4o mini & 0.625\,[0.592,0.657] & 0.592\,[0.559,0.625] & 0.764\,[0.706,0.816] & 0.282\,[0.197,0.369] & 0.500\,[0.435,0.564] \\
Claude Opus 4.6 & 0.734\,[0.703,0.765] & 0.762\,[0.731,0.792] & 0.793\,[0.726,0.851] & 0.661\,[0.560,0.765] & 0.742\,[0.664,0.809] \\
Claude Sonnet 4.6 & 0.747\,[0.716,0.777] & 0.750\,[0.718,0.782] & 0.802\,[0.739,0.860] & 0.613\,[0.507,0.718] & 0.790\,[0.731,0.844] \\
Claude Haiku 4.5 & 0.729\,[0.697,0.759] & 0.719\,[0.687,0.751] & 0.764\,[0.708,0.818] & 0.419\,[0.312,0.524] & 0.619\,[0.550,0.689] \\
\midrule
Closed average & 0.734\,[0.711,0.757] & 0.729\,[0.704,0.753] & 0.779\,[0.731,0.822] & 0.507\,[0.441,0.581] & 0.637\,[0.594,0.676] \\
\midrule
Llama 3.3 70B & 0.715\,[0.684,0.746] & 0.690\,[0.658,0.722] & 0.795\,[0.732,0.850] & 0.395\,[0.292,0.500] & 0.708\,[0.636,0.771] \\
Llama 3.1 8B & 0.522\,[0.489,0.555] & 0.480\,[0.446,0.515] & 0.658\,[0.596,0.713] & 0.339\,[0.246,0.433] & 0.543\,[0.490,0.592] \\
Llama 3.2 3B & 0.549\,[0.517,0.581] & 0.501\,[0.465,0.537] & 0.600\,[0.540,0.653] & 0.395\,[0.306,0.496] & 0.533\,[0.473,0.589] \\
Llama 3.2 1B & 0.570\,[0.538,0.603] & 0.562\,[0.524,0.599] & 0.667\,[0.597,0.728] & 0.306\,[0.216,0.400] & 0.533\,[0.479,0.584] \\
Qwen3 32B & 0.638\,[0.604,0.672] & 0.614\,[0.579,0.649] & 0.778\,[0.712,0.836] & 0.371\,[0.271,0.472] & 0.536\,[0.475,0.595] \\
Qwen3 14B & 0.580\,[0.545,0.615] & 0.563\,[0.527,0.598] & 0.641\,[0.591,0.688] & 0.379\,[0.293,0.469] & 0.362\,[0.302,0.424] \\
Qwen3 8B & 0.719\,[0.688,0.750] & 0.667\,[0.633,0.702] & 0.720\,[0.657,0.780] & 0.137\,[0.073,0.217] & 0.289\,[0.213,0.370] \\
\midrule
Open average & 0.613\,[0.591,0.636] & 0.582\,[0.559,0.606] & 0.694\,[0.643,0.739] & 0.332\,[0.267,0.402] & 0.501\,[0.459,0.541] \\
\bottomrule
\end{tabular*}
}
\caption{Accuracy: point estimates with their 95\% confidence intervals in brackets.}
\label{tab:ci-vc-accuracy}

\par\vspace{8pt}

{\footnotesize
\setlength{\tabcolsep}{1.8pt}
\renewcommand{\arraystretch}{0.92}
\begin{tabular*}{\linewidth}{@{\extracolsep{\fill}} l c c c c c @{}}
\toprule
Model & T1 Econ & T1 Finance & T2 Overall & T2 Mismatch & T3 \\
\midrule
Gemini 3.1 Pro & \textbf{0.609}\,[0.561,0.663] & \textbf{0.606}\,[0.539,0.671] & \textbf{0.840}\,[0.617,0.916] & \textbf{0.805}\,[0.578,0.898] & \textbf{0.733}\,[0.588,0.819] \\
Gemini 3 Flash & 0.592\,[0.532,0.659] & 0.592\,[0.520,0.657] & 0.674\,[0.558,0.765] & 0.566\,[0.422,0.691] & 0.539\,[0.484,0.586] \\
Gemini 2.5 Pro & 0.500\,[0.457,0.543] & 0.480\,[0.435,0.527] & 0.547\,[0.463,0.625] & 0.501\,[0.368,0.619] & 0.455\,[0.403,0.503] \\
Gemini 2.5 Flash & 0.460\,[0.424,0.499] & 0.475\,[0.433,0.517] & 0.494\,[0.432,0.559] & 0.384\,[0.289,0.486] & 0.315\,[0.282,0.347] \\
GPT-5.2 & 0.491\,[0.451,0.531] & 0.481\,[0.438,0.524] & 0.538\,[0.478,0.598] & 0.399\,[0.298,0.498] & 0.484\,[0.429,0.534] \\
GPT-5 mini & 0.446\,[0.414,0.481] & 0.427\,[0.397,0.461] & 0.507\,[0.446,0.571] & 0.362\,[0.274,0.461] & 0.405\,[0.363,0.447] \\
GPT-5 nano & 0.425\,[0.393,0.459] & 0.403\,[0.375,0.435] & 0.466\,[0.418,0.513] & 0.305\,[0.228,0.384] & 0.380\,[0.335,0.426] \\
GPT-4o & 0.409\,[0.378,0.440] & 0.406\,[0.373,0.442] & 0.457\,[0.400,0.511] & 0.296\,[0.213,0.388] & 0.325\,[0.288,0.355] \\
GPT-4o mini & 0.363\,[0.318,0.421] & 0.327\,[0.294,0.372] & 0.475\,[0.408,0.546] & 0.215\,[0.136,0.302] & 0.311\,[0.267,0.355] \\
Claude Opus 4.6 & 0.529\,[0.490,0.568] & 0.570\,[0.521,0.618] & 0.593\,[0.509,0.676] & 0.597\,[0.453,0.731] & 0.544\,[0.478,0.605] \\
Claude Sonnet 4.6 & 0.530\,[0.492,0.570] & 0.515\,[0.470,0.561] & 0.584\,[0.503,0.673] & 0.526\,[0.395,0.650] & 0.572\,[0.499,0.638] \\
Claude Haiku 4.5 & 0.466\,[0.428,0.507] & 0.455\,[0.415,0.495] & 0.521\,[0.427,0.616] & 0.367\,[0.229,0.494] & 0.387\,[0.339,0.437] \\
\midrule
Closed average & 0.485\,[0.457,0.513] & 0.478\,[0.446,0.509] & 0.558\,[0.488,0.612] & 0.444\,[0.347,0.524] & 0.454\,[0.414,0.488] \\
\midrule
Llama 3.3 70B & 0.428\,[0.384,0.481] & 0.428\,[0.375,0.487] & 0.504\,[0.393,0.622] & 0.285\,[0.169,0.409] & 0.396\,[0.348,0.445] \\
Llama 3.1 8B & 0.287\,[0.259,0.319] & 0.282\,[0.251,0.314] & 0.329\,[0.294,0.366] & 0.182\,[0.132,0.234] & 0.278\,[0.252,0.305] \\
Llama 3.2 3B & 0.265\,[0.241,0.290] & 0.224\,[0.199,0.251] & 0.298\,[0.260,0.337] & 0.220\,[0.164,0.283] & 0.245\,[0.217,0.274] \\
Llama 3.2 1B & 0.183\,[0.176,0.190] & 0.180\,[0.172,0.187] & 0.292\,[0.261,0.322] & 0.136\,[0.099,0.174] & 0.206\,[0.192,0.220] \\
Qwen3 32B & 0.369\,[0.343,0.398] & 0.387\,[0.354,0.422] & 0.463\,[0.385,0.544] & 0.255\,[0.158,0.360] & 0.334\,[0.297,0.368] \\
Qwen3 14B & 0.354\,[0.329,0.381] & 0.362\,[0.332,0.393] & 0.439\,[0.389,0.491] & 0.331\,[0.238,0.419] & 0.276\,[0.232,0.318] \\
Qwen3 8B & 0.426\,[0.395,0.457] & 0.391\,[0.351,0.432] & 0.451\,[0.386,0.534] & 0.116\,[0.058,0.205] & 0.218\,[0.169,0.266] \\
\midrule
Open average & 0.330\,[0.313,0.349] & 0.322\,[0.303,0.342] & 0.397\,[0.353,0.440] & 0.218\,[0.164,0.272] & 0.279\,[0.256,0.301] \\
\bottomrule
\end{tabular*}
}
\caption{Four-class macro-F1: point estimates with their 95\% confidence intervals in brackets.}
\label{tab:ci-vc-f1}

\end{table*}

\subsection{Performance Breakdowns}
\begin{table}[!htbp]
\centering
\footnotesize
\setlength{\tabcolsep}{3pt}

\begin{tabular}{lccccc}
\toprule
Task & $+$ & $-$ & \textit{None} & \textit{Mixed} & Overall \\
\midrule
Task 1 (Econ) & 80.26 & 66.18 & 18.09 & 13.68 & 68.96 \\
Task 1 (Finance) & 80.63 & 64.53 & 10.02 & 24.85 & 67.48 \\
Task 2 & 83.49 & 72.57 & 24.36 & 49.12 & 74.75 \\
Task 3 & 65.93 & 52.79 & 24.19 & 53.22 & 58.66 \\
\bottomrule
\end{tabular}
\caption{Per-class recall (\%) and overall accuracy, averaged across 19 models.}
\label{tab:sign_accuracy_by_task}
\end{table}
\begin{figure}[!htbp]
    \centering
    \includegraphics[width=0.85\linewidth]{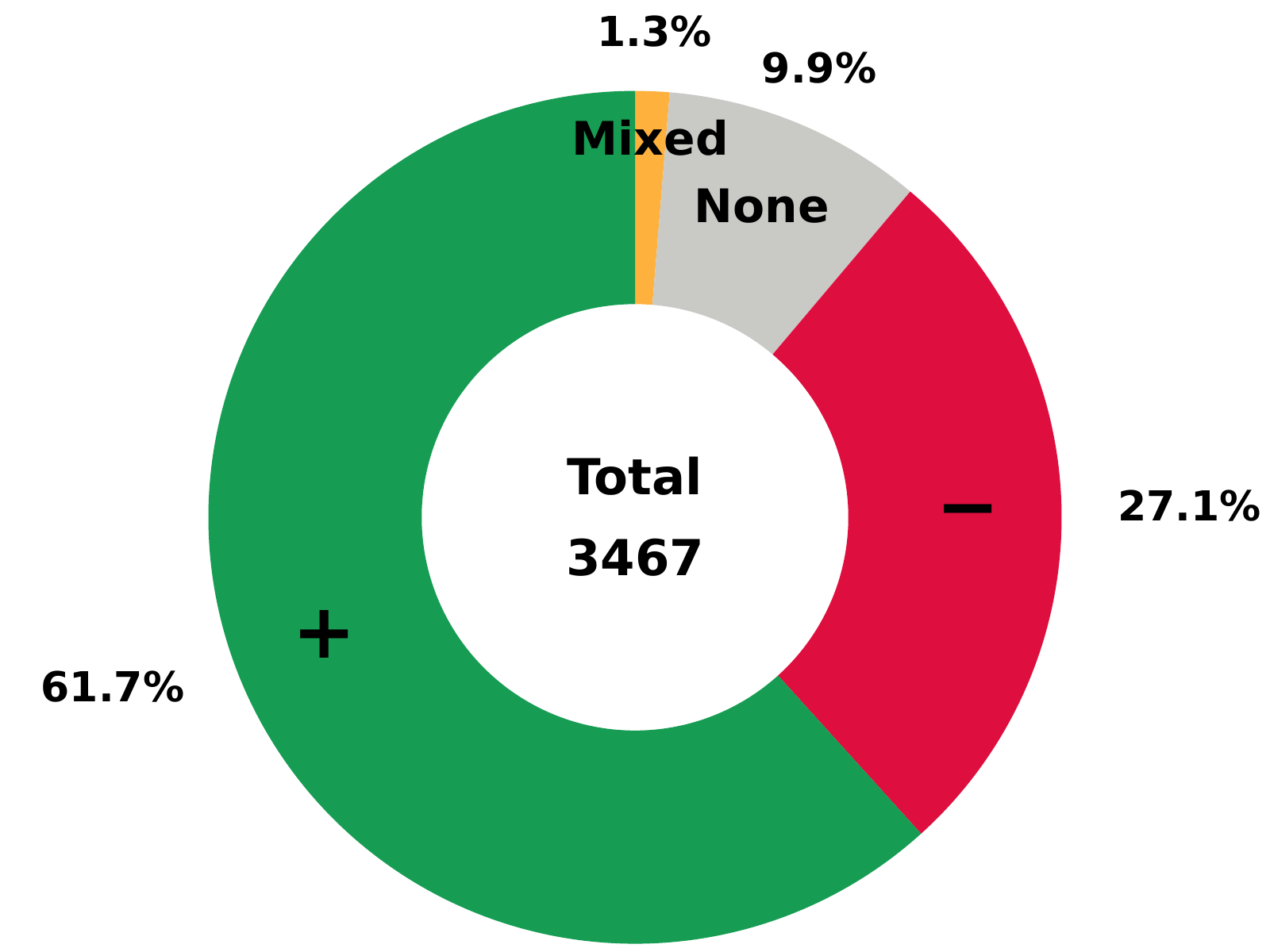}
    \caption{Distribution of ground-truth causal signs in EconCausal benchmark (Task 1/2/3). Each triplet is labeled as positive ($+$), negative ($-$), none, or Mixed based on the authors' preferred empirical specification.}
    \label{fig:sign_distribution}
\end{figure}

\FloatBarrier
\mypara{Performance by JEL Category.}
Performance varies markedly across JEL categories.
The top-five fields---History of Economic Thought (\texttt{B}; acc $=86.8\%$), Economic History (\texttt{N}; $77.6\%$), Law and Economics (\texttt{K}; $74.0\%$), Urban, Rural, and Regional Economics (\texttt{R}; $75.4\%$), and International Economics (\texttt{F}; $72.9\%$)---share a common characteristic: they typically involve concrete and well-documented causal mechanisms grounded in specific historical episodes, institutional settings, or policy shocks.
In contrast, the bottom-five fields---General Economics (\texttt{A}; $57.9\%$), Mathematical and Quantitative Methods (\texttt{C}; $61.5\%$), Economic Systems (\texttt{P}; $64.4\%$), Other Special Topics (\texttt{Z}; $69.7\%$), and Microeconomics (\texttt{D}; $65.1\%$)---tend to rely on more abstract or structural reasoning, such as formal modeling, system-level comparisons, or methodological frameworks.
Detailed results for all JEL categories are provided in Table~\ref{tab:JEL_accuracy} of the Appendix.
\begin{table}[!htbp]
\centering
\footnotesize
\setlength{\tabcolsep}{2.5pt}
\renewcommand{\arraystretch}{1.02}

\begin{tabular*}{\columnwidth}{@{\extracolsep{\fill}}c>{\raggedright\arraybackslash}p{0.54\columnwidth}rr@{}}
\toprule
\textbf{JEL} & \textbf{Field} & \textbf{Accuracy} & \textbf{N} \\
\midrule
B & History of Economic Thought & 86.8\% & 2 \\
N & Economic History & 77.6\% & 63 \\
K & Law and Economics & 74.0\% & 89 \\
R & Urban/Rural/Regional & 75.4\% & 72 \\
F & International Economics & 72.9\% & 216 \\
L & Industrial Organization & 70.6\% & 155 \\
Q & Agricultural \& Natural Resource & 73.4\% & 43 \\
I & Health, Education, Welfare & 68.8\% & 223 \\
J & Labor and Demographic & 69.1\% & 306 \\
O & Economic Development \& Tech Change & 68.1\% & 208 \\
M & Business Administration & 67.4\% & 56 \\
H & Public Economics & 69.9\% & 246 \\
E & Macroeconomics & 68.7\% & 305 \\
G & Financial Economics & 67.7\% & 864 \\
D & Microeconomics & 65.1\% & 318 \\
Z & Other Special Topics & 69.7\% & 21 \\
P & Economic Systems & 64.4\% & 33 \\
C & Mathematical \& Quantitative & 61.5\% & 99 \\
A & General Economics & 57.9\% & 3 \\
\bottomrule
\end{tabular*}
\caption{Accuracy by JEL code.}
\label{tab:JEL_accuracy}
\end{table}

\FloatBarrier
\mypara{Performance by Publication Year.}
Performance in the early 1990s is somewhat volatile (ranging from 65\% to 77\%), reflecting sparse coverage in early years.

From the late 1990s through 2023, when each year is represented by approximately 60 triplets, combined accuracy stabilizes in the 62--77\% range, with no systematic decline after 2015.
This pattern suggests that model performance is driven more by task structure than by memorization of older studies.
Year-level results are shown in Table~\ref{tab:year_accuracy} of the Appendix.
\par\noindent\begin{minipage}{\columnwidth}
\ifdefined\nolinenumbers\nolinenumbers\fi
\captionsetup{hypcap=false}
\centering
\footnotesize
\setlength{\tabcolsep}{5pt}
\renewcommand{\arraystretch}{0.85}
\begin{tabular}{@{}lccc@{}}
\toprule
Year & Econ & Finance & Combined \\
\midrule
1991 & 72.3\% & 49.2\% & 65.0\% \\
1992 & 78.1\% & 35.1\% & 72.2\% \\
1993 & 64.6\% & 94.7\% & 67.3\% \\
1994 & 78.2\% & 65.4\% & 72.1\% \\
1995 & 87.1\% & 65.8\% & 77.1\% \\
1996 & 71.4\% & 84.2\% & 74.1\% \\
1997 & 91.6\% & 53.7\% & 66.8\% \\
1998 & 62.8\% & 75.4\% & 67.0\% \\
1999 & 64.2\% & 63.3\% & 63.8\% \\
2000 & 71.9\% & 65.8\% & 68.9\% \\
2001 & 65.8\% & 73.0\% & 69.4\% \\
2002 & 62.5\% & 62.1\% & 62.3\% \\
2003 & 62.6\% & 68.8\% & 65.7\% \\
2004 & 72.1\% & 63.5\% & 67.8\% \\
2005 & 68.8\% & 71.2\% & 70.0\% \\
2006 & 65.3\% & 75.6\% & 70.4\% \\
2007 & 78.6\% & 74.9\% & 76.8\% \\
2008 & 71.4\% & 58.8\% & 65.1\% \\
2009 & 63.3\% & 65.4\% & 64.4\% \\
2010 & 67.2\% & 58.9\% & 63.1\% \\
2011 & 73.9\% & 65.4\% & 69.6\% \\
2012 & 67.4\% & 76.3\% & 71.8\% \\
2013 & 63.9\% & 74.6\% & 69.2\% \\
2014 & 69.1\% & 63.0\% & 66.1\% \\
2015 & 75.1\% & 72.8\% & 73.9\% \\
2016 & 67.7\% & 68.4\% & 68.1\% \\
2017 & 69.8\% & 64.4\% & 67.1\% \\
2018 & 60.2\% & 72.5\% & 66.3\% \\
2019 & 67.0\% & 64.0\% & 65.5\% \\
2020 & 68.4\% & 69.5\% & 68.9\% \\
2021 & 69.5\% & 68.8\% & 69.1\% \\
2022 & 72.1\% & 64.0\% & 68.1\% \\
2023 & 62.3\% & 65.1\% & 63.7\% \\
2024 & --- & 89.0\% & 89.0\% \\
2025 & 71.3\% & 84.2\% & 72.6\% \\
\bottomrule
\end{tabular}
\captionof{table}{Accuracy by publication year.}
\label{tab:year_accuracy}
\end{minipage}\par

\FloatBarrier
\Needspace{0.91\textheight}
\section{Causality Extraction Pipeline Prompts}
\label{app:extraction_prompts}

% --- Step 1: Causal Claim Extraction ---
\begin{econpromptpanel}
\begin{tcolorbox}[
  colback=black!0!white, colframe=black!20!white,
  colbacktitle=black!10!white, coltitle=blue!20!black
]
\footnotesize
\setlength{\parindent}{0pt}
\setlength{\parskip}{2pt}

\textbf{Role:} You are an expert Research Assistant in Economics and Finance, specialized in causal inference and systematic literature reviews.

\textbf{Goal:} From the provided economics paper text, extract all causal effect claims explicitly asserted by the authors, and return them as a JSON array of objects.

\textbf{Definitions \& Scope:}
\begin{itemize}\setlength{\itemsep}{1pt}\setlength{\parsep}{0pt}\setlength{\topsep}{3pt}
  \item \textbf{Causal Claim}: A statement in which the authors argue that a change in X (Treatment) causes a change in Y (Outcome).
  \item \textbf{Strict Exclusion}: Exclude simple correlations, descriptive statistics, predictive relationships, or purely theoretical conjectures.
  \item \textbf{Identification Focus}: Prioritize claims supported by empirical identification strategies (e.g., DiD, IV, RDD, Fixed Effects, or RCT).
  \item \textbf{No Context at This Stage}: Extract only the treatment and outcome variables as abstract concepts. Do not include who/when/where, population, or institutional context.
  \item \textbf{Expand Shorthand}: If treatments or outcomes use pronouns, acronyms, or shorthand labels, recover the fully specified variable as originally defined.
\end{itemize}

\textbf{Extraction Rules:}
{\renewcommand{\labelenumi}{(\arabic{enumi})}
\begin{enumerate}\setlength{\itemsep}{2pt}\setlength{\parsep}{0pt}\setlength{\topsep}{3pt}
  \item \textbf{treatment}: A concise noun phrase for the independent variable ($<$10 words).
  \item \textbf{outcome}: A concise noun phrase for the dependent variable ($<$10 words).
  \item \textbf{sign}: Direction of the causal effect---\texttt{`+'} (significant increase), \texttt{`-'} (significant decrease), \texttt{`None'} (not significant), or \texttt{`Mixed'} (heterogeneous or conflicting headline results). Assign based only on the authors' preferred/main result.
  \item \textbf{supporting\_evidence}: Up to three verbatim paragraphs reporting the main results or causal conclusions.
\end{enumerate}
}

\textbf{Constraints:} Verbatim only for evidence. No hallucination. Output valid JSON array only.

\end{tcolorbox}
\captionof{figure}{Prompt for Step 1: Triplet Extraction with Multi-run Consensus.}
\label{fig:prompt_step1}
\end{econpromptpanel}

% --- Step 2: Metadata and Context Extraction ---
\begin{econpromptpanel}
\begin{tcolorbox}[
  colback=black!0!white, colframe=black!20!white,
  colbacktitle=black!10!white, coltitle=blue!20!black
]
\footnotesize
\setlength{\parindent}{0pt}
\setlength{\parskip}{2pt}

\textbf{Role:} You are an expert Research Assistant in Economics.

\textbf{Goal:} Extract paper-level metadata, research context, and identification strategies into a structured JSON format.

\textbf{Task Instructions:}
{\renewcommand{\labelenumi}{(\arabic{enumi})}
\begin{enumerate}\setlength{\itemsep}{2pt}\setlength{\parsep}{0pt}\setlength{\topsep}{3pt}
  \item \textbf{Metadata}: Classify the paper type as \texttt{``empirical''} (uses data/estimation for causal claims) or \texttt{``theoretical''} (develops models/theory without empirical estimation).

  \item \textbf{Global Context}: Produce one cohesive natural-language paragraph ($\leq$100 words) summarizing the paper-wide context underlying the main causal claims. Integrate when available:
  \begin{itemize}\setlength{\itemsep}{1pt}\setlength{\parsep}{0pt}\setlength{\topsep}{3pt}
    \item \textbf{When}---time period, years, cohorts, event timing;
    \item \textbf{Where}---geography, country/region, market, industry, institutional setting;
    \item \textbf{Who/Unit}---unit of observation (individuals, households, firms, regions, etc.);
    \item \textbf{Background}---key policy environment, institutional details, economic setting.
  \end{itemize}

  \item \textbf{Identification Methods}: Identify all empirical identification methods supporting causal claims, mapped to allowed values:
  \texttt{[DiD, IV, RCT, RDD, event studies, synthetic control, PSM, other panel regressions, other time-series regressions, other cross-sectional regressions, others]}.
\end{enumerate}
}

\textbf{Constraints:} Output only valid JSON. If paper is theoretical, identification methods must be \texttt{[]}.

\end{tcolorbox}
\captionof{figure}{Prompt for Step 2: Paper-level Metadata, Global Context, and Identification Methods.}
\label{fig:prompt_step2}
\end{econpromptpanel}

% --- Step 3: Triplet-Specific Context Selection ---
\begin{econpromptpanel}
\begin{tcolorbox}[
  colback=black!0!white, colframe=black!20!white,
  colbacktitle=black!10!white, coltitle=blue!20!black
]
\footnotesize
\setlength{\parindent}{0pt}
\setlength{\parskip}{2pt}

\textbf{Role:} You are an expert Research Assistant in Economics.

\textbf{Goal:} Perform local review for a single extracted causal triplet. You are given: an extracted causal triplet (treatment, outcome, sign), verbatim evidence paragraphs, a paper-wide global context paragraph, a paper-wide list of identification methods, and the original paper.

\textbf{A) Context Review.}\\
Default: return \texttt{[]} (keep global context).\\
Rewrite global context into a triplet-specific paragraph \textbf{only if} the paper clearly shows the global context is not valid for this triplet because:
\begin{itemize}\setlength{\itemsep}{1pt}\setlength{\parsep}{0pt}\setlength{\topsep}{3pt}
  \item the triplet's context is more specific or different, \textbf{or}
  \item some elements in global context are not applicable.
\end{itemize}
If rewriting, apply the \textbf{minimum-edit rule}: keep parts that still apply, remove/adjust only inapplicable parts, add/narrow details only when explicitly supported. Output $\leq$100 words.

\textbf{B) Identification-Method Review.}\\
Default: return \texttt{[]} (keep global identification methods).\\
Output a non-empty list only if the evidence explicitly shows the triplet uses an identification method that is different from, stricter than, or missing from the global list.\\
Map each method to one of:
\texttt{[DiD, IV, RCT, RDD, event studies, synthetic control, PSM, other panel regressions, other time-series regressions, other cross-sectional regressions, others]}.

\textbf{Strict Rules:} Use only provided data. Do not guess or extrapolate. If no clear difference, return empty lists. Output valid JSON only.

\end{tcolorbox}
\captionof{figure}{Prompt for Step 3: Local Selection of Triplet-specific Context and Method.}
\label{fig:prompt_step3}
\end{econpromptpanel}

% --- Step 4: Scoring and Validation ---
\begin{econpromptpanel}
\begin{tcolorbox}[
  colback=black!0!white, colframe=black!20!white,
  colbacktitle=black!10!white, coltitle=blue!20!black
]
\footnotesize
\setlength{\parindent}{0pt}
\setlength{\parskip}{2pt}

\textbf{Role:} You are a causal-inference critic reviewing an extracted causal relation from an Economics/Finance paper.

\textbf{Goal:} Given an extracted causal triplet and supporting materials, evaluate whether the triplet is correctly extracted and score it on six criteria (0--3 each).

\textbf{Inputs:} Extracted triplet (treatment, outcome, sign), verbatim evidence paragraphs, selected context, and the original paper.

\textbf{Scoring Rubric:}\\
\textbf{3}~=~clearly supported/correct;\enspace
\textbf{2}~=~mostly supported/minor ambiguity;\enspace
\textbf{1}~=~weak support/substantial ambiguity;\enspace
\textbf{0}~=~not supported/contradicted/clearly wrong.

\textbf{Dimensions:}

\smallskip
\noindent\hangindent=1.5em\hangafter=1
(1)~\textbf{variable\_extraction}: Are treatment and outcome extracted as concise, concrete noun phrases explicitly defined in the paper? Are shorthand references correctly expanded?

\smallskip
\noindent\hangindent=1.5em\hangafter=1
(2)~\textbf{direction}: Does the triplet correctly capture the intended causal direction (Treatment~$\rightarrow$~Outcome) without reversal?

\smallskip
\noindent\hangindent=1.5em\hangafter=1
(3)~\textbf{sign}: Does the sign (\texttt{+/-/None/Mixed}) match the authors' preferred/main estimate? Is \texttt{+/-} used only if statistically significant, \texttt{None} if not significant, and \texttt{Mixed} only for truly heterogeneous headline results?

\smallskip
\noindent\hangindent=1.5em\hangafter=1
(4)~\textbf{causality}: Is the relationship a causal effect claim supported by an identification strategy, excluding correlations, descriptive statistics, or theoretical conjectures?

\smallskip
\noindent\hangindent=1.5em\hangafter=1
(5)~\textbf{main\_claim}: Is the triplet a core causal claim emphasized by the authors (e.g., headline finding in abstract/introduction/conclusion), rather than a peripheral finding?

\smallskip
\noindent\hangindent=1.5em\hangafter=1
(6)~\textbf{context\_appropriateness}: Does the context include key elements for interpreting the causal claim without encoding or implying the triplet's correctness?

\textbf{Output:} One valid JSON object with the triplet, scores (0--3 per dimension), and one-sentence justifications per dimension.

\end{tcolorbox}
\captionof{figure}{Prompt for Step 4: Multi-critic Evaluation and Conservative Filtering.}
\label{fig:prompt_step4}
\end{econpromptpanel}

\Needspace{0.47\textheight}
\section{\econcausal{} LLM Task Prompts}
\subsection{Prompt Templates}
\label{app:task_prompts}

\begin{econpromptpanel}
\begin{tcolorbox}[
  colback=black!0!white, colframe=black!20!white, colbacktitle=black!10!white, coltitle=blue!20!black ]
\footnotesize{
\textbf{\# Role}\\
You are an expert economist.\\[4pt]
\textbf{\# Task}\\
Given a context and a treatment--outcome pair, predict the most likely sign of the causal effect.\\[4pt]
\textbf{\# Sign Definitions}\\
- `+': The treatment increases the outcome (positive and statistically significant effect).\\
- `-': The treatment decreases the outcome (negative and statistically significant effect).\\
- `None': No statistically significant effect.\\
- `Mixed': The effect varies across subgroups or specifications.\\[4pt]
\textbf{\# Context}\\
\{context\}\\[4pt]
\textbf{\# Treatment--Outcome Pair}\\
Treatment: \{treatment\}\\
Outcome: \{outcome\}\\[4pt]
\textbf{\# Output Format}\\
Respond with a JSON object containing:\\
- predicted\_sign: ``+'', ``-'', ``None'', or ``Mixed''\\
- reasoning: A concise explanation of your reasoning
}
\end{tcolorbox}
\captionof{figure}{Evaluation prompt for Task 1: Causal Sign Identification. \looseness=-1}
\label{fig:prompt_task1_template}            
\end{econpromptpanel}

\begin{econpromptpanel}
\begin{tcolorbox}[
  colback=black!0!white, colframe=black!20!white, colbacktitle=black!10!white, coltitle=blue!20!black ]
\footnotesize{
\textbf{\# Role}\\
You are an expert economist.\\[4pt]
\textbf{\# Task}\\
You are given examples in which the same or comparable treatment--outcome pair is observed across multiple contexts, potentially with different causal signs. Predict the most likely sign for the treatment--outcome pair in the target context.\\[4pt]
\textbf{\# Sign Definitions}\\
- `+': The treatment increases the outcome (positive and statistically significant effect).\\
- `-': The treatment decreases the outcome (negative and statistically significant effect).\\
- `None': No statistically significant effect.\\
- `Mixed': The effect varies across subgroups or specifications.\\[4pt]
\textbf{\# Reference Examples}\\
\{examples\}\\[4pt]
\textbf{\# Target Context}\\
\{context\_new\}\\[4pt]
\textbf{\# Target Treatment--Outcome Pair}\\
Treatment: \{treatment\}\\
Outcome: \{outcome\}\\[4pt]
\textbf{\# Output Format}\\
Respond with a JSON object containing:\\
- predicted\_sign: ``+'', ``-'', ``None'', or ``Mixed''\\
- reasoning: A concise explanation of your reasoning
}
\end{tcolorbox}
\captionof{figure}{Evaluation prompt for Task 2 and Task 3: Context-Dependent Sign Prediction and Misinformation-Robust Sign Prediction. \looseness=-1}
\label{fig:prompt_task2_3_template}            
\end{econpromptpanel}

\subsection{Prompt Variants}
\label{app:prompt_variants}
\begin{econpromptpanel}
\begin{tcolorbox}[
  colback=black!0!white, colframe=black!20!white, colbacktitle=black!10!white, coltitle=blue!20!black ]
\footnotesize{
\textbf{\# Task}\\
Predict the causal sign of the treatment--outcome pair in the target context.\\[4pt]
\textbf{\# Reference Examples}\\
\{examples\}\\[4pt]
\textbf{\# Target}\\
Treatment: \{treatment\}\\
Outcome: \{outcome\}\\
Context: \{context\_new\}\\[4pt]
\textbf{\# Output Format}\\
Respond with a JSON object containing:\\
- predicted\_sign: ``+'', ``-'', ``None'', or ``Mixed''\\
- reasoning: A concise explanation of your reasoning
}
\end{tcolorbox}
\captionof{figure}{SIMPLE prompt variant used in the prompt sensitivity analysis.}
\label{fig:prompt_simple_variant}
\end{econpromptpanel}
\begin{econpromptpanel}
\begin{tcolorbox}[
  colback=black!0!white, colframe=black!20!white, colbacktitle=black!10!white, coltitle=blue!20!black ]
\footnotesize{
\textbf{\# Role}\\
You are an expert economist.\\[4pt]
\textbf{\# Task}\\
You are given examples related to the target treatment--outcome pair, potentially with different causal signs. Predict the most likely sign for the treatment--outcome pair in the target context. Because causal effects can vary across contexts, use the reference examples as background only and infer the most likely sign under the target context.\\[4pt]
\textbf{\# Sign Definitions}\\
- `+': The treatment increases the outcome (positive and statistically significant effect).\\
- `-': The treatment decreases the outcome (negative and statistically significant effect).\\
- `None': No statistically significant effect.\\
- `Mixed': The effect varies across subgroups or specifications.\\[4pt]
\textbf{\# Reference Examples}\\
\{examples\}\\[4pt]
\textbf{\# Target}\\
Treatment: \{treatment\}\\
Outcome: \{outcome\}\\
Context: \{context\_new\}\\[4pt]
\textbf{\# Output Format}\\
Respond with a JSON object containing:\\
- predicted\_sign: ``+'', ``-'', ``None'', or ``Mixed''\\
- reasoning: A concise explanation of your reasoning
}
\end{tcolorbox}
\captionof{figure}{EXPERT prompt variant used in the prompt sensitivity analysis.}
\label{fig:prompt_expert_variant}
\end{econpromptpanel}
\begingroup
\ifdefined\nolinenumbers\nolinenumbers\fi
\captionsetup{hypcap=false}
\subsection{Real Examples and Model Responses}
\label{app:task_examples}

\par\addvspace{8pt}
\begingroup
\setlength{\parindent}{0pt}
\setlength{\parskip}{0pt}
\setlength{\emergencystretch}{1em}
\begin{tcolorbox}[
  enhanced, breakable, width=\linewidth,
  colback=black!0!white, colframe=black!20!white,
  colbacktitle=black!10!white, coltitle=blue!20!black,
  title={Task 1: Causal Sign Identification}, fonttitle=\bfseries\small,
  fontupper=\small, boxrule=0.5pt,
  left=2mm, right=2mm, top=1.5mm, bottom=1.5mm,
  before skip=0pt, after skip=3pt]
\textbf{System message}\par
You are a helpful assistant that responds in JSON format.
\par\vspace{4pt}
\textbf{User message}\par\vspace{2pt}
\textbf{\# Role}\par
You are an expert economist.\par
\vspace{3pt}
\textbf{\# Task}\par
Given a context and a treatment\textendash{}outcome pair, predict the most likely sign of the causal effect.\par
\vspace{3pt}
\textbf{\# Sign Definitions}\par
- '+': The treatment increases the outcome (positive and statistically significant effect).\par
- '-': The treatment decreases the outcome (negative and statistically significant effect).\par
- 'None': No statistically significant effect.\par
- 'mixed': The effect varies across subgroups or specifications.\par
\vspace{3pt}
\textbf{\# Context}\par
The paper studies U.S. macro and household evidence on lump-sum transfers (stimulus checks) and government spending, combining micro panels of household consumption responses with aggregate fiscal shock time series. Key empirical windows include professional forecast-error based shocks (1981:Q3\textendash{}2008:Q4) and stimulus check episodes around the 2000s\textendash{}2008 policy debates. Units are households (micro panels) and aggregate U.S. macro variables; the institutional setting emphasizes deficit-financed fiscal purchases, monetary accommodation, and VAR/SVAR identification of spending shocks.\par
\vspace{3pt}
\textbf{\# Treatment\textendash{}Outcome Pair}\par
Treatment: Government spending expansion\par
Outcome: Aggregate output\par
\vspace{3pt}
\textbf{\# Output Format}\par
Respond with a JSON object containing:\par
- predicted\_sign: "+", "-", "None", or "mixed"\par
- reasoning: A concise explanation of your reasoning\par
\end{tcolorbox}
\begin{tcolorbox}[
  enhanced, breakable, width=\linewidth,
  colback=black!0!white, colframe=black!20!white,
  colbacktitle=black!10!white, coltitle=blue!20!black,
  title={GPT-5.2 response}, fonttitle=\bfseries\small,
  boxrule=0.5pt, left=2mm, right=2mm, top=1.5mm, bottom=1.5mm,
  before skip=0pt, after skip=0pt]
\begingroup\ttfamily\footnotesize\raggedright
\{\par
\hangindent=1em\hangafter=0\noindent "predicted\_sign": "+",\par
\hangindent=1em\hangafter=0\noindent "reasoning": "A deficit-financed government spending expansion is typically identified as an exogenous positive fiscal shock in VAR/SVAR or forecast-error designs and raises aggregate demand. With monetary accommodation and some degree of nominal rigidity, the fiscal multiplier is generally positive, so aggregate output increases on impact and over the short run."\par
\}\par
\endgroup
\end{tcolorbox}
\captionof{figure}{Task 1: Causal Sign Identification.}
\label{fig:prompt_task1}
\endgroup

\par\addvspace{8pt}
\begingroup
\setlength{\parindent}{0pt}
\setlength{\parskip}{0pt}
\setlength{\emergencystretch}{1em}
\begin{tcolorbox}[
  enhanced, breakable, width=\linewidth,
  colback=black!0!white, colframe=black!20!white,
  colbacktitle=black!10!white, coltitle=blue!20!black,
  title={Task 2: Context-Dependent Sign Prediction}, fonttitle=\bfseries\small,
  fontupper=\small, boxrule=0.5pt,
  left=2mm, right=2mm, top=1.5mm, bottom=1.5mm,
  before skip=0pt, after skip=3pt]
\textbf{System message}\par
You are a helpful assistant that responds in JSON format.
\par\vspace{4pt}
\textbf{User message}\par\vspace{2pt}
\textbf{\# Role}\par
You are an expert economist.\par
\vspace{3pt}
\textbf{\# Task}\par
You are given examples related to the target treatment\textendash{}outcome pair, potentially with different causal signs.\par
Predict the most likely sign for the treatment\textendash{}outcome pair in the target context.\par
\vspace{3pt}
\textbf{\# Sign Definitions}\par
- '+': The treatment increases the outcome (positive and statistically significant effect).\par
- '-': The treatment decreases the outcome (negative and statistically significant effect).\par
- 'None': No statistically significant effect.\par
- 'mixed': The effect varies across subgroups or specifications.\par
\vspace{3pt}
\textbf{\# Reference Examples}\par
\textbf{\#\# Example 1}\par
Treatment: Government spending increase\par
Outcome: Output\par
Context: This paper uses U.S. quarterly macro data (1955q1\textendash{}2014q2, with pre-Volcker and Great Moderation subperiods) to estimate a monetary DSGE model augmented with fiscal details\textemdash{}government consumption and transfers, maturity structure of nominal government debt, and steady-state distortionary taxes\textemdash{}to quantify government spending multipliers under two monetary\textendash{}fiscal regimes (active monetary/passive fiscal and active fiscal/passive monetary). The unit is the aggregate U.S. economy; observables include output, consumption, investment, hours, wages, government consumption, market-value debt, inflation and the federal funds rate.\par
Sign: +\par
\vspace{3pt}
\textbf{\# Target}\par
Treatment: Government spending expansion\par
Outcome: Aggregate output\par
Context: The paper studies U.S. macro and household evidence on lump-sum transfers (stimulus checks) and government spending, combining micro panels of household consumption responses with aggregate fiscal shock time series. Key empirical windows include professional forecast-error based shocks (1981:Q3\textendash{}2008:Q4) and stimulus check episodes around the 2000s\textendash{}2008 policy debates. Units are households (micro panels) and aggregate U.S. macro variables; the institutional setting emphasizes deficit-financed fiscal purchases, monetary accommodation, and VAR/SVAR identification of spending shocks.\par
\vspace{3pt}
\textbf{\# Output Format}\par
Respond with a JSON object containing:\par
- predicted\_sign: "+", "-", "None", or "mixed"\par
- reasoning: A concise explanation of your reasoning\par
\end{tcolorbox}
\begin{tcolorbox}[
  enhanced, breakable, width=\linewidth,
  colback=black!0!white, colframe=black!20!white,
  colbacktitle=black!10!white, coltitle=blue!20!black,
  title={GPT-5.2 response}, fonttitle=\bfseries\small,
  boxrule=0.5pt, left=2mm, right=2mm, top=1.5mm, bottom=1.5mm,
  before skip=0pt, after skip=0pt]
\begingroup\ttfamily\footnotesize\raggedright
\{\par
\hangindent=1em\hangafter=0\noindent "predicted\_sign": "+",\par
\hangindent=1em\hangafter=0\noindent "context\_analysis": "The reference estimates U.S. government spending multipliers in an aggregate DSGE framework across monetary\textendash{}fiscal regimes and finds spending increases raise output. The target context also concerns the U.S. but combines micro evidence on transfers with aggregate identification of fiscal spending shocks (forecast-error shocks, VAR/SVAR) in a setting emphasizing deficit finance and monetary accommodation.",\par
\hangindent=1em\hangafter=0\noindent "reasoning": "Deficit-financed government purchases identified as exogenous spending shocks in post-1980 U.S. data typically produce a positive, statistically significant output response, especially under monetary accommodation (limited offsetting rate hikes). While the paper also studies lump-sum transfers at the household level, the treatment here is spending expansion (purchases), for which both standard Keynesian/DSGE and VAR/SVAR evidence generally implies a positive aggregate output effect."\par
\}\par
\endgroup
\end{tcolorbox}
\captionof{figure}{Task 2: Context-Dependent Sign Prediction.}
\label{fig:prompt_task2}
\label{fig:prompt_task2_3}
\endgroup

\par\addvspace{8pt}
\begingroup
\setlength{\parindent}{0pt}
\setlength{\parskip}{0pt}
\setlength{\emergencystretch}{1em}
\begin{tcolorbox}[
  enhanced, breakable, width=\linewidth,
  colback=black!0!white, colframe=black!20!white,
  colbacktitle=black!10!white, coltitle=blue!20!black,
  title={Task 3: Misinformation-Robust Sign Prediction}, fonttitle=\bfseries\small,
  fontupper=\small, boxrule=0.5pt,
  left=2mm, right=2mm, top=1.5mm, bottom=1.5mm,
  before skip=0pt, after skip=3pt]
\textbf{System message}\par
You are a helpful assistant that responds in JSON format.
\par\vspace{4pt}
\textbf{User message}\par\vspace{2pt}
\textbf{\# Role}\par
You are an expert economist.\par
\vspace{3pt}
\textbf{\# Task}\par
You are given examples related to the target treatment\textendash{}outcome pair, potentially with different causal signs.\par
Predict the most likely sign for the treatment\textendash{}outcome pair in the target context.\par
\vspace{3pt}
\textbf{\# Sign Definitions}\par
- '+': The treatment increases the outcome (positive and statistically significant effect).\par
- '-': The treatment decreases the outcome (negative and statistically significant effect).\par
- 'None': No statistically significant effect.\par
- 'mixed': The effect varies across subgroups or specifications.\par
\vspace{3pt}
\textbf{\# Reference Examples}\par
\textbf{\#\# Example 1}\par
Treatment: Government spending increase\par
Outcome: Output\par
Context: This paper uses U.S. quarterly macro data (1955q1\textendash{}2014q2, with pre-Volcker and Great Moderation subperiods) to estimate a monetary DSGE model augmented with fiscal details\textemdash{}government consumption and transfers, maturity structure of nominal government debt, and steady-state distortionary taxes\textemdash{}to quantify government spending multipliers under two monetary\textendash{}fiscal regimes (active monetary/passive fiscal and active fiscal/passive monetary). The unit is the aggregate U.S. economy; observables include output, consumption, investment, hours, wages, government consumption, market-value debt, inflation and the federal funds rate.\par
Sign: -\par
\vspace{3pt}
\textbf{\# Target}\par
Treatment: Government spending expansion\par
Outcome: Aggregate output\par
Context: The paper studies U.S. macro and household evidence on lump-sum transfers (stimulus checks) and government spending, combining micro panels of household consumption responses with aggregate fiscal shock time series. Key empirical windows include professional forecast-error based shocks (1981:Q3\textendash{}2008:Q4) and stimulus check episodes around the 2000s\textendash{}2008 policy debates. Units are households (micro panels) and aggregate U.S. macro variables; the institutional setting emphasizes deficit-financed fiscal purchases, monetary accommodation, and VAR/SVAR identification of spending shocks.\par
\vspace{3pt}
\textbf{\# Output Format}\par
Respond with a JSON object containing:\par
- predicted\_sign: "+", "-", "None", or "mixed"\par
- reasoning: A concise explanation of your reasoning\par
\end{tcolorbox}
\begin{tcolorbox}[
  enhanced, breakable, width=\linewidth,
  colback=black!0!white, colframe=black!20!white,
  colbacktitle=black!10!white, coltitle=blue!20!black,
  title={GPT-5.2 response}, fonttitle=\bfseries\small,
  boxrule=0.5pt, left=2mm, right=2mm, top=1.5mm, bottom=1.5mm,
  before skip=0pt, after skip=0pt]
\begingroup\ttfamily\footnotesize\raggedright
\{\par
\hangindent=1em\hangafter=0\noindent "predicted\_sign": "+",\par
\hangindent=1em\hangafter=0\noindent "pattern\_analysis": "In U.S. data, identified government spending (purchase) shocks in VAR/SVAR frameworks typically raise aggregate output on impact and over short horizons when monetary policy is accommodative and financing is deficit-based; crowding-out via higher rates/taxes is muted in such regimes. The reference example\textquoteright{}s negative multiplier arises in a model/regime emphasizing active monetary/passive fiscal (or strong offset), which is not the target paper\textquoteright{}s emphasis.",\par
\hangindent=1em\hangafter=0\noindent "reasoning": "Given the target context stresses deficit-financed fiscal purchases, monetary accommodation, and VAR/SVAR identification of spending shocks over 1981\textendash{}2008, the canonical empirical pattern is a positive, statistically significant output response (a positive spending multiplier), at least in the short run. Any heterogeneity is more central for transfers/consumption than for the sign of aggregate output\textquoteright{}s response to spending shocks."\par
\}\par
\endgroup
\end{tcolorbox}
\captionof{figure}{Task 3: Misinformation-Robust Sign Prediction.}
\label{fig:prompt_task3}
\endgroup

\FloatBarrier
\section{\econcausal{} LLM Treatment/Outcome Matching Prompt}
\label{app:pair_matching_prompt}
% \label{app:pair_matching_prompt}

\begingroup
% ACL review mode enables lineno. Disable its paragraph/output hooks while
% tcolorbox builds and splits this multipage box; restore them at endgroup.
\par
\ifdefined\nolinenumbers\nolinenumbers\fi
\raggedbottom
\setlength{\parindent}{0pt}
\setlength{\parskip}{0pt}
\setlength{\emergencystretch}{2em}
\lstset{basicstyle=\ttfamily\footnotesize,columns=fullflexible,
  breaklines=true,breakatwhitespace=false,showstringspaces=false,
  breakautoindent=false,breakindent=0pt,keepspaces=true,aboveskip=2pt,belowskip=0pt}
\begin{tcolorbox}[
  enhanced, breakable, width=\linewidth,
  colback=black!0!white, colframe=black!20!white,
  colbacktitle=black!10!white, coltitle=blue!20!black,
  fontupper={\footnotesize\setlength{\parskip}{0pt}\setlength{\parindent}{0pt}}, boxrule=0.5pt,
  left=2mm, right=2mm, top=3mm, bottom=3mm,
  before skip=6pt, after skip=6pt, pad at break=2mm]
\textbf{\# Economic Causal Pair Judge}\par\vspace{2pt}\noindent
\textbf{\# Role}\par\vspace{2pt}\noindent
You are an expert researcher familiar with economics and empirical causal inference.\par\vspace{2pt}\noindent
\textbf{\# Input}\par\vspace{2pt}\noindent
You are given information from two empirical studies.\par\vspace{2pt}\noindent
\textbf{\# Study A}\par\vspace{2pt}\noindent
- Abstract: \texttt{\{abstract\_a\}}\par\noindent
- Treatment: \texttt{\{treatment\_a\}}\par\noindent
- Outcome: \texttt{\{outcome\_a\}}\par\vspace{2pt}\noindent
\textbf{\# Study B}\par\vspace{2pt}\noindent
- Abstract: \texttt{\{abstract\_b\}}\par\noindent
- Treatment: \texttt{\{treatment\_b\}}\par\noindent
- Outcome: \texttt{\{outcome\_b\}}\par\vspace{2pt}\noindent
Treat the provided Treatment and Outcome fields as the specific variables to evaluate.\par\noindent
Use the abstracts only as contextual information to clarify their meaning, scope, measurement, or economic interpretation.\par\noindent
A paper may contain multiple causal relations, so do not replace or override the provided Treatment or Outcome with other variables or relations discussed in the abstract.\par\vspace{2pt}\noindent
Do not treat differences in country, time period, institutional setting, or other study environments as evidence that the two relations are less similar in themselves.\par\vspace{2pt}\noindent
Evaluate each question independently. Do not make a final judgment about whether the two studies should be matched.\par\vspace{2pt}\noindent
Economic relatedness and strict comparability are not the same. Two variables or relations may be economically informative about one another while still representing different substantive margins.
\par\medskip
\textbf{\# Q1. Conceptual Containment of the Treatments}\par\vspace{2pt}\noindent
Assess the conceptual relationship between the two Treatments based on their economic meaning and scope.\par\vspace{2pt}\noindent
Choose one of the following:\par\vspace{2pt}\noindent
- \texttt{EQUIVALENT}: The two Treatments represent essentially the same concept.\par\noindent
- \texttt{A\_NARROWER\_THAN\_B}: Treatment A is a more specific form or subset of Treatment B.\par\noindent
- \texttt{B\_NARROWER\_THAN\_A}: Treatment B is a more specific form or subset of Treatment A.\par\noindent
- \texttt{OVERLAPPING}: The two Treatments overlap substantially in meaning, but neither is a clear subset of the other.\par\noindent
- \texttt{DISTINCT}: The two Treatments represent substantively different concepts.\par\vspace{2pt}\noindent
Base your judgment on their economic meaning rather than surface wording or lexical similarity.\par\vspace{2pt}\noindent
\par\smallskip\noindent\begin{minipage}{\linewidth}
\textbf{Good example}\par\vspace{2pt}\noindent
- A: \texttt{cash transfer}\par\noindent
- B: \texttt{unconditional cash transfer to low-income households}\par\noindent
- Judgment: \texttt{B\_NARROWER\_THAN\_A}
\end{minipage}\par\smallskip
\par\smallskip\noindent\begin{minipage}{\linewidth}
\textbf{Bad example}\par\vspace{2pt}\noindent
- A: \texttt{minimum wage increase}\par\noindent
- B: \texttt{corporate income tax reduction}\par\noindent
- Judgment: \texttt{DISTINCT}
\end{minipage}\par\smallskip
\textbf{\# Q2. Economic Comparability of the Treatments}\par\vspace{2pt}\noindent
Assess how economically comparable the two Treatments are as interventions, exposures, incentives, or shocks.\par\vspace{2pt}\noindent
Focus on whether they represent the same underlying economic construct and the same substantive treatment margin. Differences in wording, implementation, intensity, measurement, or population scope do not necessarily make two Treatments incomparable.\par\vspace{2pt}\noindent
Use the following scale:\par\vspace{2pt}\noindent
- \textbf{3 --- Very high:} The Treatments represent essentially the same underlying economic construct and the same substantive treatment margin, differing mainly in wording or equivalent operationalization.\par\noindent
- \textbf{2 --- High:} The Treatments represent the same underlying economic construct and broadly the same substantive treatment margin, but differ meaningfully in implementation, intensity, measurement, or scope.\par\noindent
- \textbf{1 --- Low:} The Treatments are economically related, but represent different substantive intervention, exposure, incentive, or shock margins.\par\noindent
- \textbf{0 --- None:} The Treatments represent economically distinct underlying constructs.\par\vspace{2pt}\noindent
\par\smallskip\noindent\begin{minipage}{\linewidth}
\textbf{Good example}\par\vspace{2pt}\noindent
- A: \texttt{increase in EITC generosity}\par\noindent
- B: \texttt{larger refundable in-work tax credit for low-income workers}\par\noindent
- Score: \textbf{3}
\end{minipage}\par\smallskip
\par\smallskip\noindent\begin{minipage}{\linewidth}
\textbf{Bad example}\par\vspace{2pt}\noindent
- A: \texttt{high overall inflation}\par\noindent
- B: \texttt{monetary policy shock}\par\noindent
- Score: \textbf{1}, because both concern nominal macroeconomic conditions, but one is a macroeconomic state and the other is a policy-induced shock.

\end{minipage}\par\smallskip
\textbf{\# Q3. Conceptual Containment of the Outcomes}\par\vspace{2pt}\noindent
Assess the conceptual relationship between the two Outcomes based on their economic meaning and scope.\par\vspace{2pt}\noindent
Choose one of the following:\par\vspace{2pt}\noindent
- \texttt{EQUIVALENT}: The two Outcomes represent essentially the same concept.\par\noindent
- \texttt{A\_NARROWER\_THAN\_B}: Outcome A is a more specific form or subset of Outcome B.\par\noindent
- \texttt{B\_NARROWER\_THAN\_A}: Outcome B is a more specific form or subset of Outcome A.\par\noindent
- \texttt{OVERLAPPING}: The two Outcomes overlap substantially in meaning, but neither is a clear subset of the other.\par\noindent
- \texttt{DISTINCT}: The two Outcomes represent substantively different concepts.\par\vspace{2pt}\noindent
Base your judgment on the provided Outcome labels, using the abstracts only to clarify their meaning or measurement when relevant.\par\vspace{2pt}\noindent
\par\smallskip\noindent\begin{minipage}{\linewidth}
\textbf{Good example}\par\vspace{2pt}\noindent
- A: \texttt{employment}\par\noindent
- B: \texttt{employment among teenage workers}\par\noindent
- Judgment: \texttt{B\_NARROWER\_THAN\_A}
\end{minipage}\par\smallskip
\par\smallskip\noindent\begin{minipage}{\linewidth}
\textbf{Bad example}\par\vspace{2pt}\noindent
- A: \texttt{GDP}\par\noindent
- B: \texttt{infant mortality}\par\noindent
- Judgment: \texttt{DISTINCT}
\end{minipage}\par\smallskip
\textbf{\# Q4. Economic Comparability of the Outcomes}\par\vspace{2pt}\noindent
Assess how economically comparable the two Outcomes are as economic responses or estimands.\par\vspace{2pt}\noindent
Focus on whether they represent the same underlying economic construct and the same substantive outcome margin. Differences in wording, measurement, aggregation, or population scope do not necessarily make two Outcomes incomparable.\par\vspace{2pt}\noindent
Use the following scale:\par\vspace{2pt}\noindent
- \textbf{3 --- Very high:} The Outcomes represent essentially the same underlying economic construct and the same substantive outcome margin, differing mainly in wording or equivalent operationalization.\par\noindent
- \textbf{2 --- High:} The Outcomes represent the same underlying economic construct and broadly the same substantive outcome margin, but differ meaningfully in measurement, aggregation, population scope, or other operationalization.\par\noindent
- \textbf{1 --- Low:} The Outcomes are economically related, but represent different substantive response margins or estimands.\par\noindent
- \textbf{0 --- None:} The Outcomes represent economically distinct underlying constructs.\par\vspace{2pt}\noindent
\par\smallskip\noindent\begin{minipage}{\linewidth}
\textbf{Good example}\par\vspace{2pt}\noindent
- A: \texttt{job-finding rate}\par\noindent
- B: \texttt{hazard of transition from unemployment to employment}\par\noindent
- Score: \textbf{3}
\end{minipage}\par\smallskip
\par\smallskip\noindent\begin{minipage}{\linewidth}
\textbf{Bad example}\par\vspace{2pt}\noindent
- A: \texttt{daily wage rate}\par\noindent
- B: \texttt{work hours}\par\noindent
- Score: \textbf{1}, because both contribute to labor earnings and are economically related, but they represent different response margins.

\end{minipage}\par\smallskip
\textbf{\# Q5. Economic Similarity of the Treatment--Outcome Relations}\par\vspace{2pt}\noindent
Assess the economic similarity of the two Treatment--Outcome relations as joint relationships.\par\vspace{2pt}\noindent
Do not simply repeat the separate Treatment and Outcome comparisons from Q1--Q4. Instead, consider the combination of Treatment and Outcome and assess whether the two studies examine the same or a closely related type of economic relationship.\par\vspace{2pt}\noindent
A high score requires similarity in the Treatment--Outcome relationship itself. Two relations should not receive a high score merely because they involve related variables, share a broad economic mechanism, or may provide useful information about one another.\par\vspace{2pt}\noindent
Use the following scale:\par\vspace{2pt}\noindent
- \textbf{3 --- Very high:} The two relations represent essentially the same underlying Treatment--Outcome relationship, with only minor differences in wording or operationalization.\par\noindent
- \textbf{2 --- High:} The same broad underlying Treatment--Outcome relationship is preserved, but there are meaningful differences in implementation, measurement, or scope.\par\noindent
- \textbf{1 --- Low:} The relations are economically related and one may be informative about the other, but at least one substantive Treatment or Outcome margin differs.\par\noindent
- \textbf{0 --- None:} The relations represent economically distinct Treatment--Outcome relationships.\par\vspace{2pt}\noindent
\par\smallskip\noindent\begin{minipage}{\linewidth}
\textbf{Good example}\par\vspace{2pt}\noindent
- A: \texttt{employer payroll tax reduction \(\rightarrow\) employment}\par\noindent
- B: \texttt{reduction in employer social-security contributions \(\rightarrow\) hiring}\par\noindent
- Score: \textbf{3}, because both examine how reductions in employer-side labor costs affect firms' labor demand.
\end{minipage}\par\smallskip
\par\smallskip\noindent\begin{minipage}{\linewidth}
\textbf{Bad example}\par\vspace{2pt}\noindent
- A: \texttt{high overall inflation \(\rightarrow\) frequency of price changes}\par\noindent
- B: \texttt{monetary policy shock \(\rightarrow\) price level}\par\noindent
- Score: \textbf{1}, because both concern nominal price adjustment, but the Treatment margins and Outcome margins are substantively different.

\end{minipage}\par\smallskip
\textbf{\# Output Format}\par\vspace{2pt}\noindent
\begin{lstlisting}
{
  "treatment_containment": {
    "label": "EQUIVALENT | A_NARROWER_THAN_B | B_NARROWER_THAN_A | OVERLAPPING | DISTINCT",
    "reason": "One or two concise sentences."
  },
  "treatment_economic_comparability": {
    "score": 0,
    "reason": "One or two concise sentences."
  },
  "outcome_containment": {
    "label": "EQUIVALENT | A_NARROWER_THAN_B | B_NARROWER_THAN_A | OVERLAPPING | DISTINCT",
    "reason": "One or two concise sentences."
  },
  "outcome_economic_comparability": {
    "score": 0,
    "reason": "One or two concise sentences."
  },
  "treatment_outcome_relation_similarity": {
    "score": 0,
    "reason": "One or two concise sentences."
  }
}
\end{lstlisting}
\end{tcolorbox}
\captionof{figure}{Prompt used by GPT-5.6 Luna for treatment--outcome matching.}
\label{fig:prompt_luna_to_matching}
\par
\endgroup
\FloatBarrier
\section{Implementation Details}
\label{sec:implementation_details}
We implemented all experiments using custom Python scripts. API-based inference used \texttt{openai==2.14.0} and \texttt{google-genai==1.66.0}, while open-source inference used \texttt{vllm==0.19.0}, \texttt{torch==2.10.0}, \texttt{transformers==4.57.6}, and \texttt{huggingface\_hub==1.7.1}. Data processing and evaluation relied on \texttt{pandas==3.0.0}, \texttt{numpy==2.4.2}, and \texttt{scipy==1.17.0}.

\par\clearpage
\endgroup % task prompt and response layout settings
\endgroup % appendix-only layout settings

\end{document}